\documentclass{article}

\usepackage[preprint]{neurips_2022}

\usepackage[bookmarks=false,colorlinks=true,linkcolor=blue,citecolor=violet]{hyperref}

\usepackage{mathtools} % amsmath with fixes and additions
\usepackage{booktabs} % commands to create good-looking tables
\usepackage{tikz} % nice language for creating drawings and diagrams

\usepackage[utf8]{inputenc} % allow utf-8 input
\usepackage[T1]{fontenc}    % use 8-bit T1 fonts
\usepackage{hyperref}       % hyperlinks
\usepackage{url}            % simple URL typesetting
\usepackage{booktabs}       % professional-quality tables
\usepackage{amsfonts}       % blackboard math symbols
\usepackage{nicefrac}       % compact symbols for 1/2, etc.
\usepackage{microtype}      % microtypography
\usepackage{xcolor}         % colors

\usepackage{graphicx}
\usepackage{amsthm}
\usepackage{amssymb}
\usepackage{amsmath}
\usepackage{mathtools}
\usepackage{dirtytalk}
\usepackage{subcaption}
\usepackage{wrapfig}
\usepackage{placeins}
\usepackage{color}
\usepackage{soul}

\usepackage[noend,ruled,vlined,linesnumbered]{algorithm2e}
\usepackage{setspace}

\SetCommentSty{mycommfont}

\theoremstyle{definition}
\newtheorem{definition}{Definition}

\newtheorem{lemma}{Lemma}

\newtheorem{corollary}{Corollary}

\title{A Mixture-of-Expert Approach \\ to RL-based Dialogue Management}

\author{%
  Yinlam Chow
   \And
   Aza Tulepbergenov
   \And 
   Ofir Nachum 
   \AND Moonkyung Ryu
   \And Mohammad Ghavamzadeh \And Craig Boutilier
   \AND \\
   Google Research\\
    \texttt{\{yinlamchow, atulep, ofirnachum, mkryu, ghavamza, cboutilier\}@google.com} \\
}

\begin{document}
\maketitle
\begin{abstract}
Despite recent advancements in language models (LMs), their application to dialogue management (DM) problems and ability to carry on rich conversations remain a challenge. We use reinforcement learning (RL) to develop a dialogue agent that avoids being short-sighted (outputting generic utterances) and maximizes overall user satisfaction. Most existing RL approaches to DM train the agent at the word-level, and thus, have to deal with a combinatorially complex action space even for a medium-size vocabulary. As a result, they struggle to produce a successful and engaging dialogue even if they are warm-started with a pre-trained LM. To address this issue, we develop a RL-based DM using a novel \emph{mixture of expert} language model (MoE-LM) that consists of (i) a LM capable of learning diverse semantics for conversation histories, (ii) a number of {\em specialized} LMs (or {\em experts}) capable of generating utterances corresponding to a particular attribute or personality, and (iii) a RL-based DM that performs dialogue planning with the utterances generated by the experts. Our MoE approach provides greater flexibility to generate sensible utterances with different intents and allows RL to focus on conversational-level DM. We compare it with SOTA baselines on open-domain dialogues and demonstrate its effectiveness both in terms of the diversity and sensibility of the generated utterances and the overall DM performance.

%Despite recent advancements in language models (LMs), their application to dialogue management (DM) and ability to carry on rich conversations remains a challenge. We use reinforcement learning (RL) to develop a dialogue agent that avoids being short-sighted (often outputting generic utterances) and maximizes overall user satisfaction. However, existing RL approaches focus on training an agent that operates at the word level. Since generating semantically-correct and sensible utterances from a large vocabulary space is combinatorially complex, RL can struggle to produce engaging dialogue, even if warm-started with a pre-trained LM. To address this issue, we develop a RL-based DM using a novel \emph{mixture-of-expert} (MoE) approach, which consists of (i) a language representation that captures diverse information, (ii) several specialized LMs (or experts) to generate candidate utterances, and (iii) a RL-based DM that performs dialogue planning with the utterances generated by the experts. This MoE approach provides greater flexibility to generate sensible utterances of different intents, and allows RL to focus on conversational-level DM. We compare it with SOTA baselines on open-domain dialogues and demonstrate its effectiveness both in the diversity and sensibility of the generated utterances as well as the overall DM performance. 
\end{abstract}

\vspace{-0.1in}
\section{Introduction}\label{sec:intro}
\vspace{-0.1in}
%!TEX root = main.tex
% real-world RL is hard, batch setting, why behavior cloning sucks

% \begin{itemize}
%     \item Focus is to develop a training framework for language models that works with RL-based mixture-of-expert dialogue management. Motivate the mixture-of-expert Dialogue management framework.
%     \item To do this,  what do we need: a model architecture that can capture diverse language semantic representation; on top of this representation, a language model that can incorporate multiple "experts"; a MoE dialogue manager
%     Why we think this architecture novel and sensible
%     \item Accordingly, we develop a 3-phase algorithm to train a language model to achieve the aforementioned criteria and show some performance guarantees
%     \item Demonstrate experimental results
% \end{itemize}

% General introduction
With the tremendous advancements in natural language understanding and generation, increasing attention has been directed to construct intelligent dialogue agents that can carry out rich and engaging conversations with human users. Such interactions can be quite open-ended, contain many different topics, and often involve an underlying task, such as building a rapport, negotiation, information exchange, and recommendation. Therefore, to satisfy the user, a good dialogue agent should not only generate natural responses, but also be capable of pursuing the task's objectives and adapting to the user's feedback on-the-fly.

% Steerable LMs
A standard solution is to train the dialogue agent using behavioral cloning, where the agent is a language model (LM) that imitates the utterances in the training set~\citep{gavsic2011line,fatemi2016policy}. By leveraging deep neural networks, e.g.,~RNNs~\citep{ sutskever2014sequence} and Transformers~\citep{vaswani2017attention}, a LM encodes the conversation to a low-dimensional dialogue state and predicts an utterance, but steering such generation for particular purposes remains an open question. Several works studied ways to fine-tune a LM to generate texts with specific contexts~\citep{ziegler2019fine, ficler2017controlling}. Other results learned a single steerable LM that is capable of generating utterances for multiple specific intents~\citep{gu2017trainable, chen2018stable, subramani2019can, dathathri2019plug}. While these LMs produce fluent and relevant responses, it is unclear how to control them to systematically pursue goals during multi-turn dialogue conversations.

% RL
Another popular approach is to view dialogue management (DM) as a control problem and use reinforcement learning (RL) to optimize the agent's policy (which is often a LM itself). Using RL for dialogue systems has a long history. Earlier work relies on specific, hand-crafted semantic states~\citep{levin1997stochastic, singh2002optimizing, walker2000application} or partially observable belief states~\citep{williams2007partially, young2010hidden}, in which the agent chooses the best hand-crafted dialogue act at each turn, with the goal of either satisfying the user~\citep{shah2018bootstrapping}, completing the task~\citep{shi2018sentiment}, or responding to the user's query~\citep{serban2017deep}. However, the application of these approaches is limited to problems whose action space can be captured
by hand-crafted representations, and they cannot handle complex conversations. On the other hand, more recent approaches use deep learning to extract semantic representations from conversation histories, treat these representations as dialogue belief states, and apply RL to learn a word-level generative DM agent~\citep{jaques2019way, li2016deep, li2017adversarial, shin2020generating}. However, since there are innumerable possibilities of language utterances, and thus, the action space of the RL problem is extremely large, the agent often performs planning poorly and generates incomprehensible utterances~\citep{zhao2019rethinking}. Another issue is that RL only optimizes a scalar reward, while the aforementioned methods often need to optimize for both the quality of the generated utterance, e.g.,~ease of answering~\citep{li2016deep} and fluency~\citep{li2017adversarial, li2019dialogue}, and the goal, e.g.,~conversation length \citep{zhou2020design}, user's sentiment~\citep{hancock2019learning}, and task completion~\citep{verma2022chai}. Moreover, defining the reward as weighted combination of these metrics is not ideal, since the hand-picked weights do not often reflect the underlying success criteria.

To address the above issues related to using RL in dialogue management (DM) systems, we propose an RL-based DM agent using a novel {\em mixture of expert} (MoE) approach. Our MoE approach is based on a mixture of expert language model (MoE-LM), which consists of three main components: {\bf 1)} a LM (a probabilistic encoder and a decoder) capable of learning diverse semantics for conversation histories, and as a result generating diverse utterances, which we refer to as the {\em primitive} LM or $\texttt{LM}_0$, {\bf 2)} a number of {\em specialized} LMs (or {\em experts}), $\{\texttt{LM}_i\}_{i=1}^m$, that each is constructed using the latent space learned by $\texttt{LM}_0$, but has been trained such that it is capable of generating utterances corresponding to a certain intent or personality, and {\bf 3)} an RL-based dialogue manager (DM) that at each turn, given the latent state shared by the experts $\{\texttt{LM}_i\}_{i=0}^m$ and the utterance action(s) they suggest, chooses one among them for the agent to execute. Our MoE-LM can be seen as a special case of hierarchical LMs (e.g.,~\citealt{serban2017deep,zhao2019rethinking,saleh2020hierarchical}), but it is different than them because it learns both the LMs (experts) and the DM. Moreover, the DM in MoE-LM is a policy conditioned on both the latent state and the actions suggested by the experts, and not just the state as it is common in hierarchical RL. The primitive LM ($\texttt{LM}_0$) plays an important role in this model because it learns diverse semantics for conversation histories and allows the agent to generate a wide variety of utterances. This diversity is also shared with the specialized LMs (experts) and gives them flexibility in generating their (more) specialized utterances. Another important feature of MoE-LM is its modularity that facilitates adding and removing specialized LMs (experts). Moreover, this hierarchical architecture allows us to solve an RL problem with much smaller state and action spaces, which is quite important in the quality of the learned policy. Finally, since the candidate utterances are generated by experts with different intents, instead of combining all agent-user signals into a single RL reward, our DM agent can focus on optimizing the specific goal of the conversation task. 

We start the paper with a brief introduction of LMs and the use of Markov decision processes (MDPs) in modeling dialogue management problems in Section~\ref{sec:prelim}. We then describe the overall architecture of our MoE-LM in Section~\ref{sec:moe}, followed by the detailed implementation of each of its three main components (described in the above paragraph) in Sections~\ref{sec:rep_l} to~\ref{sec:cm}. Finally, in Section~\ref{sec:experiments}, we demonstrate the effectiveness of our MoE-LM in open-domain dialogues, in terms of both its ability to generate diverse and sensible utterances and its overall DM performance.

\vspace{-0.1in}
\section{Preliminaries}\label{sec:prelim}
\vspace{-0.1in}
%!TEX root = main.tex 

\paragraph{\textbf{Language Models (LMs)}}{
In this work, we employ seq2seq LMs to generate the next utterances in a dialogue. We assume access to a dataset of the form $\mathcal D=\{(\mathbf{X}^{(k)},Y^{(k)})\}_{k=1}^{|\mathcal D|}$, where each $\mathbf{X}=\mathbf{X}^{(k)}$ is a $L$-turn conversation history $\mathbf{X}=\{X_l\}_{l=0}^{L-1}$ and $Y$ is its next utterance. We denote by $N_{\mathbf{X}}$, an upper-bound on the length (number of tokens) of each utterance $X_{l}$ in $\mathbf{X}$.\footnote{If the actual utterance $X_l$ has fewer tokens than $N_{\mathbf{X}}$, it will be padded by a specific token and masked.} The role of a LM is to predict the probability of the next utterance $Y$, consisting of $N$ tokens, conditioned on the conversation history $\mathbf{X}$, i.e.,~$p\big(Y=\{y_n\}_{n=1}^N\mid\mathbf{X}\big)$. In the transformer architecture~\citep{wolf2019huggingface}, the LM first encodes the conversation history $\mathbf{X}$ using an encoder $\Phi$ to a $(L\times N_{\mathbf{X}})$-length sequence of embeddings  %$\{\{z_{l,n}\}_{n=0}^{N_{\mathbf X}-1}\}_{l=0}^{L-1}$,  
$\{(z_{l,0},\ldots,z_{l,N_{\mathbf{X}}-1})\}_{l=0}^{L-1}$, 
where each $z_{l,n}$ is a vector in the latent space. For notational convenience, we concatenate these embeddings into a single embedding $z\in\mathcal Z\subseteq \mathbb R^d$ and denote the overall dimension of the latent space as $d$. In the RNN architecture~\citep{serban2016building},
% \ofir{Do we ever use RNNs? We could omit this sentence if we don't use RNNs.} 
the LM's encoder $\Phi$ directly maps the conversation history $\mathbf{X}$ to a latent state $z\in\mathcal Z\subseteq \mathbb R^d$. In both architectures, the next utterance $\widehat{Y}=\{\widehat{y}_{n}\}_{n=1}^N$ is sampled token-by-token from the decoder $\Psi$, i.e.,~$\widehat{Y}\sim\Psi\big(\cdot \mid z\big)=\prod_{n=1}^{N} \Psi\big(\widehat{y}_n \mid \widehat{y}_0,\ldots,\widehat{y}_{n-1};z\big)$, where $\widehat{y}_0$ is a fixed initial (start-of-sentence) token~\citep{chien2019markov}, and the latent state is denoted as $z=\Phi(\mathbf{X})$.\footnote{Note that we use $Y$ as the next utterance in the dataset and $\hat{Y}$ as the one predicted by the LM.}
}

\vspace{-0.1in}
\paragraph{\textbf{Markov Decision Processes (MDPs)}}{
% \ofir{This section can be difficult to understand for those not familiar with RL for dialogue. Would help if we had a pictoral diagram explaining this setup.}
$\!\!\!\!\!$ have been used to model dialogue management problems in a variety of settings~\citep{li2016deep,asadi2016sample,jaques2019way}. In such MDPs, denoted by $\mathcal M = (\mathcal S, \mathcal A, P, r, s_0, \gamma)$, the state space $\mathcal S$ represents the tokenized conversation history and the initial state $s_0\in\mathcal S$ is the initial user's query. The action space $\mathcal A$ is also the tokenized language space with each action  %utterance\footnote{While both are outputted by LMs, we denote target utterance with $Y$ and MDP action utterance with $a$.} 
$a\in\mathcal A$ being the agent's next utterance (which is a fixed-length, $N_{\mathbf X}$, sequence of tokens).
% \ofir{Can we be more precise here? Is an action a single token, or an entire utterance?} 
The transition kernel $P$ models the user's response to the action taken by the agent (bot). Finally, the reward function $r$ measures the user's satisfaction. In these MDPs, we can think of the entire LM as a policy that maps conversation histories to next utterances, and solve them by finding a policy $\pi^*$ with maximum expected discounted return, i.e.,~$\pi^*\in\arg\max_\pi J_\pi :=\mathbb E[\sum_{k=0}^{\infty}\gamma^t r_t\mid P,s_0,\pi]$. Note that the size of the tokenized state and action spaces grow exponentially with the size of the vocabulary. This makes it intractable to solve the MDP even for a medium-size vocabulary. 
%Even with a medium-size vocabulary, the tokenized state and action spaces grow exponentially, which makes solving this MDP intractable. 
As a result, it would quite desirable to develop a novel MDP paradigm that is more amendable to RL-based DM systems.

}

\vspace{-0.1in}
\section{Mixture of Experts (MoE) Language Model}\label{sec:moe}
\vspace{-0.1in}
%!TEX root = main.tex 
% Why is our MoE-MDP Approach better

\begin{figure}[t]
\begin{center}
\includegraphics[height=1.8in]{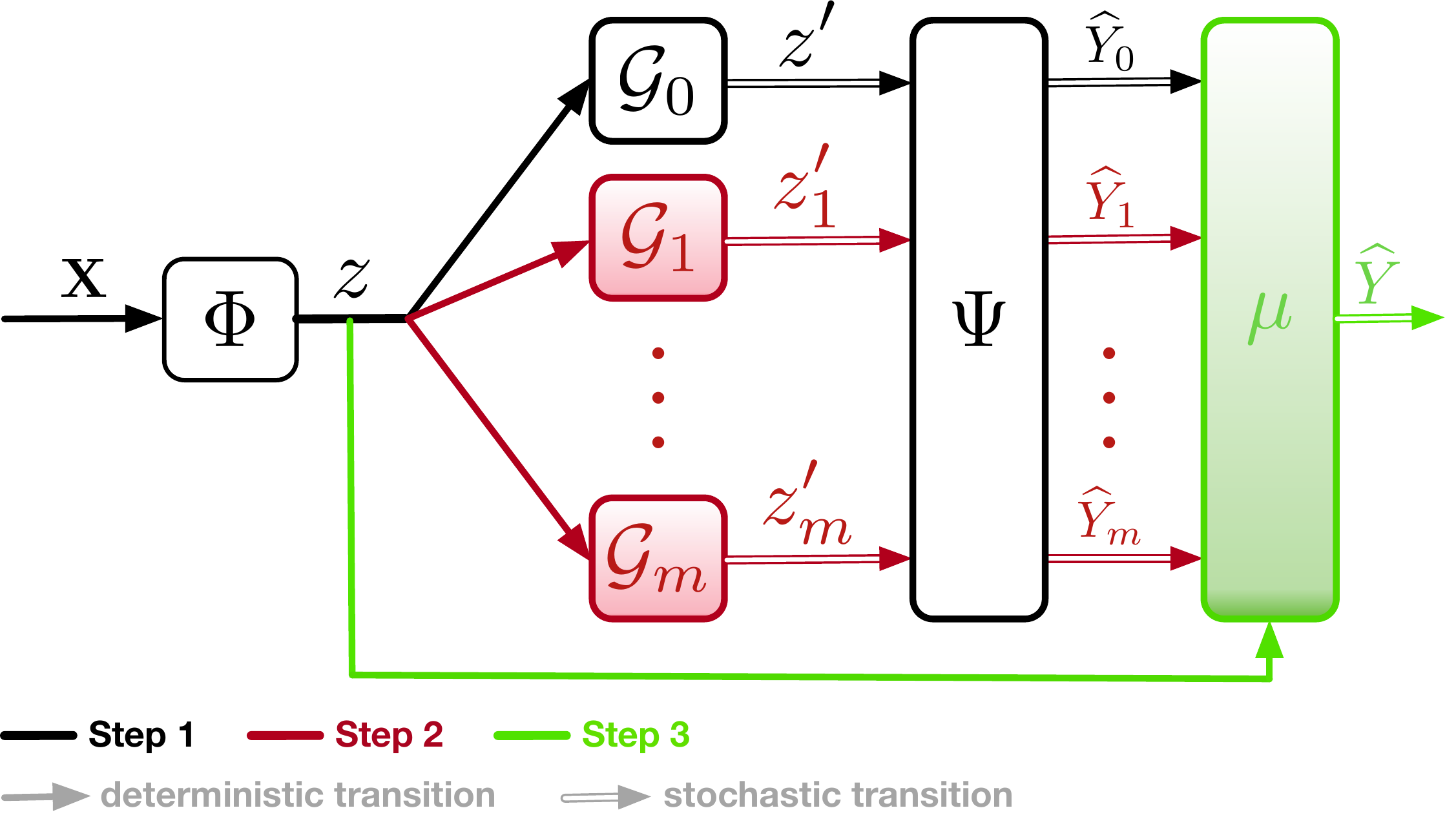}\,\,\,\includegraphics[height=2.2in]{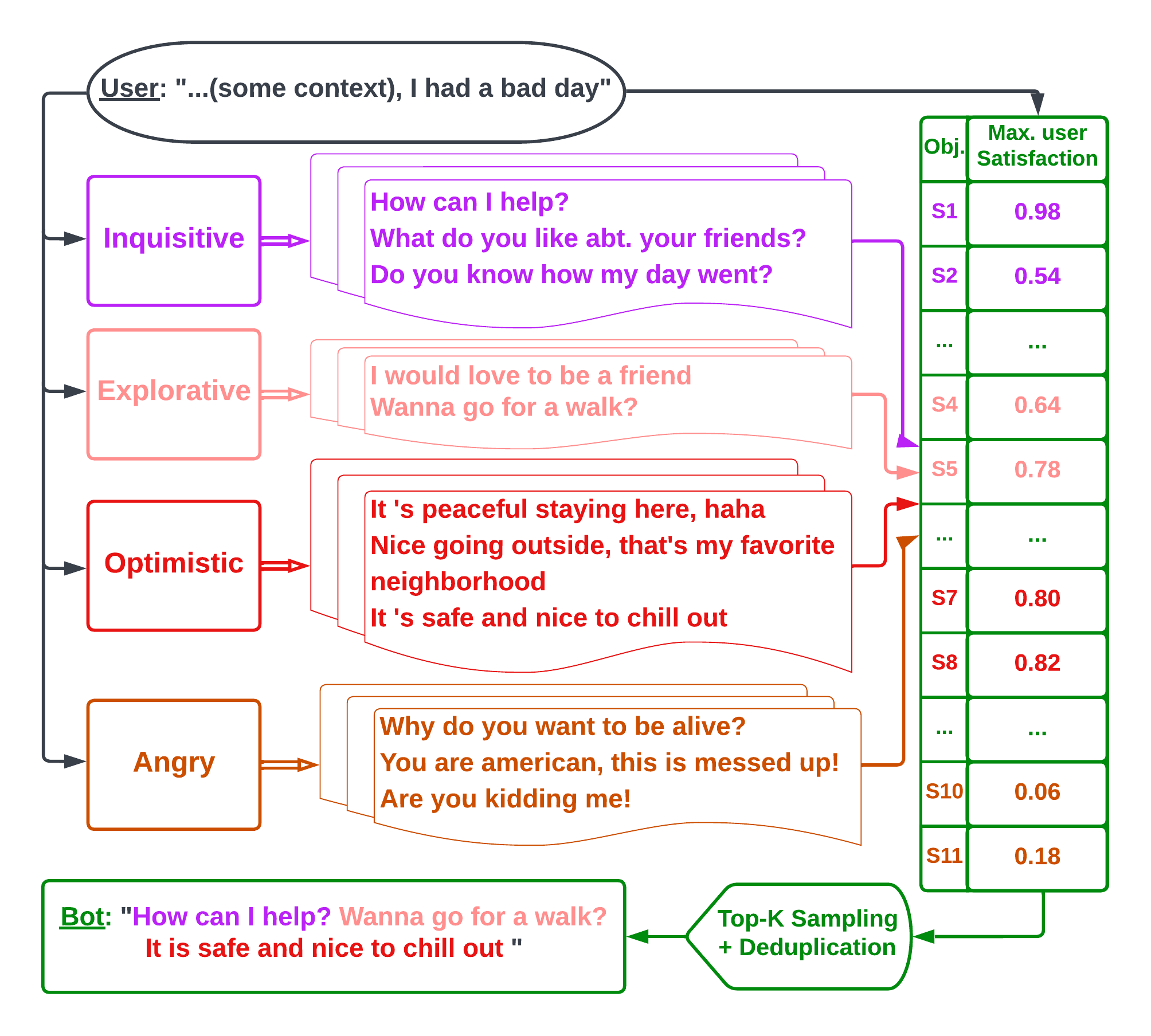}
\end{center}
\caption{(Left) MoE-LM Architecture. (Right) Sample utterance workflow generated by an MoE-LM trained with Reddit data. Step 1: $\Phi$ encodes conversation history. Step 2: $\Psi\circ \mathcal G_i$, $\forall i$, generate candidate bot utterances. Step 3: $\mu$ selects the bot response by $Q$-score ranking \& post-processing.}
\label{fig:MoE-LM}
\end{figure}

% Motivation
We start by explaining how a MoE language model (MoE-LM) can enrich the bot's utterances and improve the overall performance of the DM. While our approach is applicable to any DM system, we use open-domain dialogue \citep{sankar2019neural} as a running example to show how MoE-LM-based agents can improve user satisfaction measured by an improvement on a sentiment or engagement. Intuitively a good DM agent should possess different behaviors (e.g., inquisitive, explorative, relevant, soothing, empathetic, complimentary, provoking) and swiftly decide which intent to use to pivot a conversation, build rapport, pique the user's interests, improve their mood, etc. To achieve this goal, we require the LM to have a language representation (primitive discovery) that captures different semantics, in order to encode different conversations and avoid generating dull and repetitive responses. We also need a machinery (expert construction) to embed different intents into 
sub-models of this LM, so that they can behave accordingly when prompted, and respond efficiently. Finally, with various candidate utterances available, the DM module of this LM should understand the current level of user satisfaction and determine which response is the most appropriate. Motivated by these observations, we construct our MoE-LM in three steps as shown in Figure~\ref{fig:MoE-LM}. We give the main idea behind each step here and leave their detailed descriptions to Sections~\ref{sec:rep_l},~\ref{sec:expert_l}, and~\ref{sec:cm}.  

{\bf Step~1: Primitive Discovery.}
% \ofir{Maybe we could give each step a descriptive name? E.g., Step 1: Primitive Discovery. We should also being each step with an explanation of why we're doing this (e.g., to circumvent the challenge of exponentially large action spaces)}
We first employ the dataset $\mathcal D$, introduced in Section~\ref{sec:prelim}, and learn a language model $\texttt{LM}_0=(\Phi,\mathcal G_0,\Psi)$ consisting of a {\em stochastic encoder} %(i.e.,~an encoder $\Phi$ and a latent space distribution $\mathcal G_0:\mathbb R^d\rightarrow\mathbb P(\mathbb R^d)$, which maps the encoded conversation into a latent distribution), 
(i.e.,~an encoder $\Phi$ and a latent space distribution $\mathcal G_0$ that maps the encoded conversation into a latent distribution), and a decoder $\Psi$. The stochastic encoder $(\Phi,\mathcal G_0)$ comprises an encoder $\Phi$ that maps tokenized conversation histories $\mathbf X$ to a latent space $\mathcal Z\subseteq\mathbb R^d$, i.e.,~$z=\Phi(\mathbf X)\in\mathcal Z$, which is then used to construct a parameterized $d$-dimensional Gaussian distribution 
% \ofir{Is this correct? The encoder produces both a $z$ and a separate Gaussian distribution? Doesn't it only produce a Gaussian distribution (which $z$ is then sampled from)?}
$\mathcal G_0(z'|z) = \mathcal N\big(\mu_0(z),\sigma_0^2(z)\mathbf I_{d\times d}\big)$ over $\mathbb R^d$. The decoder predicts the next utterance $\widehat{Y}_0$ (token-by-token) conditioned on the point $z'$ sampled from the latent distribution, i.e.,~$\Psi(\widehat{Y}_0|z')\footnote{In practice, we can use both latent states as the input to the decoder model $\Psi(\widehat{Y}_0|z',z)$.},\;z'\sim\mathcal G_0(\cdot|z)$. We denote by $\texttt{LM}_0(Y|\mathbf{X}):= \mathbb E_{z'\sim \mathcal G_0(\cdot|z), z=\Phi(\mathbf{X})}[\Psi(Y|z')]$, the \emph{primitive} and learn it using a loss function that in addition to predicting the next utterance accurately, encourages diversity and generalization in the learned latent space $\mathcal Z$ (see Eq.~\ref{eq:rep_opt} and Fig.~\ref{fig:embedding_vis}). As we will explain in Section~\ref{sec:rep_l}, our loss function is inspired by those in prior work, and more specifically by the one in OPAL~\citep{ajay2020opal}, i.e.,~an unsupervised learning method for discovering primitive skills in trajectories that are used by some downstream RL tasks. 

{\bf Step~2: Expert Construction.} Given the latent space $\mathcal Z$, encoder $(\Phi,\mathcal G_0)$, and decoder $\Psi$ learned in Step~1, we now learn $m$ latent distributions $\{\mathcal G_i\}_{i=1}^m$, each defined as $\mathcal G_i(z'|z) = \mathcal N\big(\mu_i(z),\sigma_i^2(z)\mathbf I_{d\times d}\big)$. Intuitively, each $\mathcal G_i$ corresponds to an attribute, e.g., an intent or a personality (in case of a chatbot) and generates samples in specific parts of the latent space $\mathcal Z$. This results in having %Using the original encoder-decoder $(\Phi,\mathcal G_0,\Psi)$, each of the 
$m$ LMs, $\{\texttt{LM}_i\}_{i=1}^m,\;\texttt{LM}_i=(\Phi,\mathcal G_i,\Psi)$, each of them corresponds to a {\em specialized} version of the original LM, $\texttt{LM}_0$, and serves as an {\em expert} in our MoE-LM. Upon receiving a conversation history $\mathbf X$, each expert $\texttt{LM}_i$ generates a candidate (or more) for the next utterance $\widehat{Y}_i$ in certain parts of the language space that are compatible with its attribute (personality). As we will explain in Section~\ref{sec:expert_l}, each $\mathcal G_i$ is learned using a loss function that encourages its corresponding LM, $\texttt{LM}_i$, to generate utterances consistent with its attribute (see Eq.~\ref{eq:rep_opt_ft}). 

{\bf Step~3: Dialogue Manager (DM).} The \emph{dialogue manager}, denoted by $\mu$, takes as input the encoded conversation history $z=\Phi(\mathbf X)$ and the candidate action utterances generated by the experts $\{\widehat{Y}_i\}_{i=0}^m$, and selects one of them as the action for the bot to execute, i.e.,~$\widehat{Y}\sim\mu(\cdot\mid z,\{\widehat{Y}_i\}_{i=0}^m)$. We will describe how DM is trained using reinforcement learning (RL) in Section~\ref{sec:cm}.

\vspace{-0.1in}
\section{Primitive Discovery in MoE-LM}\label{sec:rep_l}
\vspace{-0.1in}
%!TEX root = main.tex
% \ofir{'Representation Learning' is a confusing term here. Perhaps `language primitive' would be better? Better yet, it would be good to align this terminology with how it is presented in the prelims, where I believe you use the term `experts'.}

% In the first phase of MoE-LM training, we would like to extract a continuous space of language semantics representation from the conversation dataset $D$, which we can later use as basis for learning other expert policies as well as training the CM with offline RL.
%Motivated by the variational methods in language modeling, such as VHRED \citep{serban2017hierarchical} and VHCR \citep{park2018hierarchical}, 
Motivated by literature in the reinforcement and imitation learning fields~\citep{ajay2020opal},
we propose to learn the primitive LM, $\texttt{LM}_0$, in our MoE-LM by solving the following KL-constrained optimization problem that aims at capturing diverse semantics: 

\vspace{-0.125in}
\begin{small}
\begin{equation}
\label{eq:rep_opt}
\min_{(\Phi,\mathcal G_0,\Psi),\rho}\widehat{\mathbb E}_{z'\sim \rho(\cdot|z, Y),z=\Phi(\mathbf{X})}\big[-\log\Psi(Y|z')\big], \quad\text{s.t.} \,\,\,\widehat{\mathbb E}_{z=\Phi(\mathbf{X})}\big[\text{KL}\big(\rho(z'|z,Y)\;||\;\mathcal G_0(z'|z)\big)\big]\leq \epsilon_{\text{KL}},
\end{equation}
%\begin{equation}\label{eq:rep_opt}
%  \min_{(\Phi,\mathcal G_0,\Psi),\mathcal G^p}\mathbb E_{\mathcal{D},z'\sim \mathcal{G}^p(\cdot|z, Y),z=\Phi(\mathbf{X})}[-\log\Psi(Y|z')]\,\,\text{s.t.} \,\,\mathbb E_{\mathcal{D}, z=\Phi(\mathbf{X})}[\text{KL}(\mathcal G^p(z'|z,Y)||\mathcal{G}_0(z'|z))]\!\leq\! \epsilon_{\text{KL}},
%\end{equation}
\end{small}
\vspace{-0.125in}

where $\widehat{\mathbb E}$ is the empirical expectation over $(\mathbf X,Y)$ in the dataset $\mathcal{D}$, $\rho$ is a distribution over the latent space conditioned on the encoded conversation history $z$ and the target utterance $Y$, and $\epsilon_{\text{KL}}$ is a positive real-valued threshold. Using~\eqref{eq:rep_opt}, we learn $\texttt{LM}_0=(\Phi,\mathcal G_0,\Psi)$ by maximizing the log-likelihood, while enforcing consistency between the latent variable $z'$ predicted by $\mathcal G_0(\cdot|z)$ and $\rho(\cdot|z,Y)$ via the KL constraint. The distribution $\rho(\cdot|z,Y)$ is a Gaussian $\mathcal N\big(\mu_\rho(z,\Phi_\rho(Y)),\sigma_\rho^2(z,\Phi_\rho(Y))\mathbf I_{d\times d}\big)$ in which $\Phi_\rho$ is a pre-trained encoder for the target utterance $Y$, and mean $\mu_\rho(\cdot,\cdot)$ and variance $\sigma_\rho^2(\cdot,\cdot)$ are trainable models. One reason for using a separate encoder $\Phi_\rho$ for the target utterance $Y$ is to avoid overfitting $\Phi$ (i.e., to avoid having back-propagation gradient of $\Phi$ with $Y$ as input). 

%As mentioned in Section \ref{sec:moe}, $\texttt{LM}_0$ consists of the encoder $\Phi$, decoder $\Psi$, latent distribution $\mathcal G_0$ 
% Equipped with the encoder $\Phi$, $\mathcal{G}(\Phi(\mathbf{X}))$ is a variational language encoder, which first encodes the conversation history with $\Phi$ and then outputs a latent multi-variate Gaussian distribution $\mathcal{G}(\Phi(\mathbf{X}))=\mathcal N(\mu(\Phi(\mathbf{X})), \sigma^2(\Phi(\mathbf{X})))$ in the $\mathbb R^d$ space, where $(\mu,\sigma^2)$ are trainable models.  The decoder is a latent-conditioned generative model, which takes the form of $\Psi(Y|z)$, for $z\sim \mathcal G(\Phi(\mathbf{X}))$,
%and is trained to maximize the log-likelihood of next utterances conditioned on the conversation history. Besides maximizing likelihood, in the above optimization we also enforce consistency between the latent variable $z'$ predicted by $\mathcal G_0(z'|z)$ and $\mathcal G^p(z'|z,Y)$. $\mathcal G^p(z'|z,Y)$ models the latent distribution from both the encoded conversation history $z$ and target utterance $Y$. It uses a separate pre-trained sentence encoder $\Phi_{\text{SE}}$ to encode target utterance $Y$ and has the form of a Gaussian distribution $\mathcal N(\mu^p(z,\Phi_{\text{SE}}(Y)), (\sigma^p)^2(z,\Phi_{\text{SE}}(Y))\mathbf I_{d\times d})$ in the latent $\mathbb R^d$ space, where $(\mu^p, (\sigma^p)^2)$ are trainable models. One reason of using a separate encoder for the target utterance $Y$ is to avoid overfitting $\Phi$ (i.e., to not have back-propagation gradient of $\Phi$ with target as input). 

\paragraph{Connection to VAE-like objectives} In practice, we implement the KL constraint in~\eqref{eq:rep_opt} as a penalty weighted by an appropriately chosen coefficient. Thus, one may interpret the objective in~\eqref{eq:rep_opt} as a variation of $\beta$-VAE~\citep{burgess2018understanding}.
Due to the connection to VAEs, one may draw similarities between our method and existing dialogue approaches such as VHRED~\citep{serban2017hierarchical} and VHCR \citep{park2018hierarchical}.
However, we emphasize that there are key differences, and these may be explained by first understanding how the objective in~\eqref{eq:rep_opt} encourages \emph{diversity}, which is key to good primitive learning.
Namely, it is important that primitive discovery learns an encoder-decoder $\Phi,\Psi$ which can be modulated by the choice of $z$; i.e., changing $z^\prime$ while fixing $\mathbf X$ should lead to different distributions over generated utterances. 
The objective in~\eqref{eq:rep_opt} encourages this diversity by conditioning the latent variable $z'$ on both the target utterance $Y$ and $z=\Phi(\mathbf X)$, i.e.,~$z'\sim\rho(\cdot|z,Y)$. 
In contrast, the KL constraint is used to make sure that the stochastic encoder $\mathcal G_0(\cdot|z)$ of our primitive LM is not too varied for different $Y$, and thus has a \emph{limiting} effect on diversity.
For example, in the extreme when $\epsilon_\mathrm{KL}=0$ (or, $\beta\to\infty$ when used as a regularizer) there will be no specialization of the latent space for different $Y$.
Although $\beta\to\infty$ is an extreme case, degenerate behavior can also happen when $\beta=1$, i.e., in the traditional variational loss used by VHRED and VHCR. 
Specifically, it is well-known that the traditional VAE loss is an upper bound on the negative log-likelihood of the data under a stochastic encoder-decoder parameterization. Thus if the data can be modeled by a single LM then a VAE-optimal decoder $\Psi$ can simply ignore $\mathcal G_0$, leading to a degenerated latent space as observed in previous work~\citep{park2018hierarchical}. This is precisely the reason that, in our approach, we weaken the KL constraint ($\epsilon_{\text{KL}} \gg 0$ or, equivalently, $\beta \ll 1$). This enables our approach to more reliably guarantee that a unique $z'$ represents each distinct conversation pair $(\mathbf{X},Y)$, thus capturing diverse semantic modalities and enabling easier downstream specialization. 

In the mathematical results below, we formalize the claim above, namely, that the log-likelihood objective in~\eqref{eq:rep_opt} leads to a learned $\Phi,\Psi$ that can easily recover any arbitrary desired LM by specializing the latent space $\mathcal G$. We begin with a definition that characterizes the coverage of an arbitrary LM on the conditional conversation data distribution $P_{\mathcal{D}}(Y|\mathbf{X})$.

\begin{definition}
$\texttt{LM}_{\mathcal{D}, \xi}$ is a $\xi$-common LM of data $\mathcal{D}$ if 
\begin{small}
$
\mathbb E_{\mathcal{D}}[\text{TV}(\texttt{LM}_{\mathcal{D},\xi}(Y|\mathbf{X})||P_{\mathcal{D}}(Y|\mathbf{X})))] \!\leq\! \xi.
$
\end{small}
\end{definition}
Leveraging Theorem 4.1 in \cite{ajay2020opal}, we now present the theoretical result characterizing the representational power of our primitive encoder-decoder pair $(\Phi,\Psi)$ on data $\mathcal D$. 

\begin{lemma}\label{lem:rep_power}
Let $(\Phi,\rho,\Psi)$ be the solution to (\ref{eq:rep_opt}) with \begin{small}$\widehat{\mathbb E}_{z'\sim \rho(\cdot|z, Y),z=\Phi(\mathbf{X})}[-\log\Psi(Y|z')]=\epsilon$\end{small}. Then there exists $\texttt{LM}:=(\Phi,\mathcal G,\Psi)$ such that $\mathbb E_{\mathcal{D}}[\text{TV}(\texttt{LM}_{\mathcal{D},\xi}(Y|\mathbf{X})||\texttt{LM}(Y|\mathbf{X}))]\!\leq\! \xi + \sqrt{\frac{1}{2}\left(\epsilon+\mathcal H\right)},
$
where $\mathcal G(z'|z)=\mathbb E_{Y\sim\mathcal D}[\rho(z'|z,Y)]$, and $\mathcal H=\mathbb E_{\mathcal{D}}[\log P_{\mathcal{D}}(Y|\mathbf{X})]$ is a constant depending on $\mathcal{D}$.
\end{lemma}
The result above shows that, as long as $\texttt{LM}_{\mathcal{D},\xi}$ is $\xi$-common in $\mathcal D$, then there exists a specialization of the latent space $\mathcal G$ that, when paired with $\Phi,\Psi$, can approximately recover $\texttt{LM}_{\mathcal{D},\xi}$.
The quality of the approximation is a function of $\epsilon$ --- how well the objective in~\eqref{eq:rep_opt} was optimized --- and $\xi$. In practice, to construct the primitive by replacing $\mathcal G$ with $\mathcal G_0$, i.e., $\texttt{LM}_0=(\Phi,\mathcal G_0,\Psi)$, because $\mathcal G_0(z'|z)$ can be viewed as an distillation of  $\rho(z'|z,Y)$. This theoretical result also motivates the next section, where we explain our algorithm's ``Step 2: Expert Construction''. Specifically, we show how to use the trained encoder-decoder pair $\Phi,\Psi$ to learn a spectrum of different specialized experts parameterized by different latent distributions $\mathcal G_i$.
%This results implies that the $\xi$-approximate data distribution (i.e., $\xi$-common LM) of conversation data $\mathcal D$ can be represented by an LM that uses encoder $\Phi$ and decoder $\Psi$ from (\ref{eq:rep_opt}) (because the RHS of the above error bound is its objective function). To construct such an LM in practice (which is known as the primitive $\texttt{LM}_0$), we set $\texttt{LM}_0=(\Phi,\mathcal G_0,\Psi)$, where $\mathcal G_0(z'|z)$ from (\ref{eq:rep_opt}) is a distillated version of the posterior latent distribution $\rho(z'|z,Y)$. In the following, we will leverage this result to construct different experts, which can be viewed as data distributions w.r.t. to various specific behaviors in $\mathcal D$. 

\vspace{-0.1in}
\section{Expert Construction with Plug-and-play Language Models}\label{sec:expert_l}
\vspace{-0.1in}
%!TEX root = main.tex

%ceb: this section needs an INTRO!!!

%!TEX root = main.tex

% \ofir{This section is really difficult to follow. I suggest to describe the algorithm at a high/abstract-level, with maybe the final objective, before getting into details such as RoBERTa, what sentiments we use, etc.}

% Recall that the MoE primitive expert $\texttt{LM}_0$ balances between diversity and predictive power. However since this LM is trained with a generic dataset $D$ it may not possess specific intents, skills, or personalities (e.g.,~in case of a chatbot).
To complete the MoE framework one needs to systematically create a gamut of different experts $\texttt{LM}_i$, $\forall i \in\{1,\ldots,m\}$, with each generating candidate utterances of different intents. 
By viewing each expert as a distribution of  particular behaviors in conversation data $\mathcal D$, we leverage the results of Section~\ref{sec:rep_l} and Lemma~\ref{lem:rep_power} and adopt a universal encoder-decoder $(\Phi,\Psi)$ among all the experts. Therefore, each expert $i$ is only parameterized by an arbitrary $d$-dimensional latent distribution (e.g., Gaussian), and it samples certain regions of the latent space $\mathcal Z$. Following the terminology from \citet{dathathri2019plug}, these experts can all be catagorized as \emph{plug-and-play language models (PPLMs)}. Creating experts is handy because it only requires learning new latent distributions, while switching between experts amounts to sampling a different distribution.

Denote by $\ell_i(\mathbf{X},Y)\in\mathbb R$ a real-valued label that \emph{characterizes} the intent of expert $i\in\{1,\ldots,m\}$, e.g., determined by an off-the-shelf sentiment classifier.
We train the latent distribution $\mathcal G_i(z)$ of expert $i$ by solving the optimization problem
\begin{equation}\label{eq:rep_opt_ft}
\min_{\mathcal G_i}\,\,\widehat{\mathbb E}_{ z'\sim \mathcal G_i(\cdot|z), z=\Phi(\mathbf{X}), Y \sim \Psi(\cdot|z')}[ -\ell_i(\mathbf{X},\! Y)].
\end{equation}
Unlike the weighted maximum likelihood approach considered in \cite{dathathri2019plug}, which assigns weight $\ell_i$ to training samples that correspond to expert $i$, we propose to learn each expert via \emph{reward-maximization} and treat $\ell_i$ as a reward signal w.r.t. expert $i$ to be maximized. Interestingly, this approach is also linked to reinforcement learning (RL), in which both the ``state'' and ``action'' spaces are the latent space $\mathcal Z$, and the ``policy'' is the latent distribution $\mathcal G_i$. The main benefit of our approach is that it does not require the target utterance $Y$ from data $\mathcal D$ and is thus less vulnerable to data-imbalance issues in $\mathcal D$ on certain intents. Notice from \eqref{eq:rep_opt_ft} that the reward-maximization problem is  myopic, i.e., the above RL problem has a discounting factor of $0$. The main motivation is that, unlike dialogue management that is a sequential decision-making problem, here we want each expert to possess
particular behaviors, and this can readily be done via greedy maximization; Long-term dialogue optimization will be handled by the dialogue manager rather than these experts.

For example in the case of a Gaussian $\mathcal G_i$, we use the standard REINFORCE \citep{sutton1999policy} algorithm to learn the model parameters $(\mu_i, \sigma_i^2)$ of $\mathcal G_i$ according to
\begin{small}
\[
\{\mu_i,\sigma_i\}\leftarrow \{\mu_i,\sigma_i\} +\alpha\cdot\mathbb E_{z'\sim \mathcal G_i(\cdot|z), Y \sim \Psi(\cdot|z')}[\ell_i(\mathbf{X},Y)\cdot\nabla_{\{\mu_i,\sigma_i\}}\log \mathbb{P}_{\mathcal G_i}(z'|z)],\,\, i\in\{1,\ldots,m\},
\]
\end{small}
where $\alpha>0$ is the learning rate.
To reduce the variance of these estimates, we also adopt the baseline reduction technique \citep{greensmith2004variance} in policy gradient. This can be simply done by replacing $\ell_i(\mathbf{X},Y)$ with $\overline{\ell}_i(\mathbf{X},Y):=\ell_i(\mathbf{X},Y) - \mathbb E_{Y\sim \Psi(\cdot|\Phi(\mathbf X))}[\ell_i(\mathbf{X},Y)]$. To quantify the performance of expert $\texttt{LM}_i$, 
following arguments from Lemma \ref{lem:rep_power} and Lemma 4.0.1 in \cite{ajay2020opal}, we have the sub-optimality result.

\begin{corollary}\label{coro:expert}
Denote the $i$-th reward-maximizing objective as $\mathcal{L}_i(\texttt{LM}):=\widehat{\mathbb E}_{Y \sim \texttt{LM}(\cdot|\mathbf{X})}[ \ell_i(\mathbf{X},\! Y)]$.
Suppose an optimal LM for this objective $\texttt{LM}_{i,\xi}\in\arg\max_{\texttt{LM}}\mathcal{L}_i(\texttt{LM})$ is $\xi$-common in $\mathcal D$. Moreover, let $\mathcal G_i^\star$ be in the $\arg\min$ of~\eqref{eq:rep_opt_ft}.
Then with expert $\texttt{LM}_i=(\Phi,\mathcal G_i^\star,\Psi)$ and ($\epsilon$, $\mathcal H$) from Lemma \ref{lem:rep_power}, we have
$
|\mathcal{L}_i(\texttt{LM}_i)-\mathcal{L}_i(\texttt{LM}_{i,\xi})|\leq 2 \|\ell_i\|_\infty\cdot( \xi + \!\sqrt{\frac{1}{2}(\epsilon+\mathcal H)})
$.
\end{corollary}
While it may be obvious that optimizing $\mathcal G_i$ w.r.t. (\ref{eq:rep_opt_ft}) encourages expert $\texttt{LM}_i$ to capture the behaviors encouraged by $\ell_i$, this corollary has two further implications: (i) Since the sub-optimality of $\texttt{LM}_i$ compared to the oracle $\texttt{LM}_{i,\xi}$ is bounded by the quantity $\epsilon$ defined in Lemma \ref{lem:rep_power}, it justifies using the primitive $(\Psi,\Phi)$, which optimizes $\epsilon$, for expert construction;
(ii) Sub-optimality further depends on $\xi$, quantifying how well $\texttt{LM}_{i,\xi}$ is represented in the original dataset $\mathcal D$.

\vspace{-0.1in}
\section{Reinforcement Learning for MoE-LM Dialogue Manager}\label{sec:cm}
\vspace{-0.1in}
We now describe the dialogue manager (DM) of our MoE-LM and propose RL algorithms to train it. As mentioned in Section~\ref{sec:moe}, the DM is a policy $\mu$ that takes the encoded conversation history $z=\Phi(\mathbf X)$ and the $m+1$ candidate action utterances generated by the experts $\{\widehat{Y}_i\}_{i=0}^m$,\footnote{For simplicity, we assume that each expert generates only a single candidate utterance at each step. It would be straightforward to extend this to multiple (and even a varying number of) candidate utterances.} and stochastically selects one of them to execute, i.e.,~$\widehat{Y}\sim\mu(\cdot\mid z,\{\widehat{Y}_i\}_{i=0}^m)$. Note that each expert $i\in\{0,\ldots,m\}$ is a LM, $\texttt{LM}_i$, that acts as a policy $\pi_i(\cdot|\mathbf X)$ and maps each conversation history $\mathcal X$ to an utterance $\widehat{Y}_i$. With this architecture we address the large size of state and action spaces in the original MDP that grows exponentially with the size of the vocabulary. As described in Section~\ref{sec:prelim}, the state and action spaces of the original MDP are the tokenized conversation history and the tokenized language space, respectively, while here the DM should choose among $m+1$ actions given the latent space $\mathcal Z$ of encoded conversations. It is important to note that our MoE-LM is different than other hierarchical architectures~\citep{kulkarni2016hierarchical} in which the decision at the high-level is to choose a low-level controller only based on the current state of the system. In MoE-LM, the DM observes both the current state and the actions suggested by the experts and then chooses one among them. 

We learn the DM policy $\mu$ by solving an MDP, which we refer to as MoE-MDP, and denote it by $\bar{\mathcal M}=(\bar{\mathcal S},\bar{\mathcal A},\bar P,\bar R,\bar{s}_0,\gamma)$. The state space of MoE-MDP is the product of the learned latent space $\mathcal Z$ and the joint action space of the $m+1$ experts, i.e.,~$\bar{\mathcal S}=\mathcal Z\times \mathcal A^{m+1}$. Its action space consists of the $m+1$ experts, i.e.,~$\bar{\mathcal A}=\{0,\ldots,m\}$. Its initial state is the encoding of the initial user's query and the utterances suggested by the experts in response to this query. The transition still models user's responses but is now over the joint space of the latent states and experts' actions. The reward function is the same as in the original MDP, i.e., $\bar{r}(\bar{s},\bar a)=r(\mathbf X,a_j)$, where $\bar s=(z,\{a_i\}_{i=0}^m)$ with $a_i\sim\pi_i(\cdot|\mathbf X)$ and $z=\Phi(\mathbf X)$, and $\bar a\in\{0,\ldots,m\}$ is the expert selected by the DM. Since MoE-MDP has finite number of actions, learning DM is equivalent to solving a finite-action MDP and finding a policy $\mu$ in the $m$-dimensional simplex with maximum expected cumulative discounted reward.  

We use two RL algorithms to solve the MoE-MDP and learn the DM policy $\mu$. The first one is {\em conservative Q-learning} (CQL)~\citep{kumar2020conservative}, a popular offline RL algorithm. This is suitable for our setting in which the policy must be learned from the collected conversations $\mathcal D$ with no further (online) interaction with the real system. CQL is a regularization scheme that learns a conservative $Q$-function that lower-bounds the true one. Given the offline conversation data $\mathcal{D}$, we parameterize the $Q$-function by parameter $\theta$ and learn $\theta$ by minimizing the Bellman error with behavior regularization: $\min_\theta \sum_{(\bar{s},\bar{a},\bar r,\bar{s}_+)\in \mathcal{D}}
\alpha(\mathbb E_{\bar{a}\sim \mu}
[Q_\theta(\bar s,\bar a)] - 
Q_\theta(\bar s, a_0))+(\bar r+\gamma Q_{\theta^{\text{target}}}(\bar{s}_+,\arg\max_{\bar{a}_+\in\bar{\mathcal{A}}} Q_{\theta}(\bar{s}_+,\bar{a}_+))-Q_\theta(\bar s,\bar a))^2$, 
where $a_0$ is action suggested by the primitive LM (which approximates the behavior policy of $\mathcal D$) $a_0\sim\pi_0$, $\alpha > 0$ is a regularization parameter, and $\theta^{\text{target}}$ is the target $Q$-function parameter. Intuitively, CQL regularization minimizes the differences in $Q$-values of our DM and the primitive. According to the CQL algorithm, we can set the DM as $\mu(\bar a|\bar s)\propto \exp( Q_\theta(\bar s,\bar a))$, which corresponds to the optimal policy of the entropy-regularized $Q$-learning~\citep{schulman2017equivalence}.

The second RL algorithm we use is model-based RL (MBRL)~\citep{shah2018bootstrapping, wei2018airdialogue}. Here we first learn a user utterance model $P_{\text{user}}(X_+|\mathbf{X},a):=\mathbb E_{ z=\Phi_{\text{user}}([\mathbf{X},a])}[\Psi_{\text{user}}(X_+|z)]$ via maximum likelihood, then generate data $\mathcal{D}_{\text{MB}}$, whose next-state $\widehat{s}_+$ encodes the next conversation generated from roll-outs and the corresponding candidate actions, and finally solve the Bellman error minimization in MoE-MDP: $\min_\theta \sum_{(\bar{s},\bar{a},\bar r,\widehat{s}_+)\in \mathcal{D}_{\text{MB}}}\!(\bar{r}+\gamma Q_{\theta^{\text{target}}}(\widehat s_+, \arg\max_{\bar{a}_+\in\mathcal J} Q_{\theta}(\widehat{s}_+, \bar{a}_+))-Q_\theta(\bar{s},\bar{a}))^2$. The benefit of MBRL over CQL is two-fold. First, one can easily obtain a \emph{high-fidelity} user utterance model \citep{peng2020few} by simply fine-tuning a large LM (e.g., GPT-3 \citep{floridi2020gpt}). Second, with sufficient dialogue roll-outs that captures many different scenarios, MBRL does not require behavior regularization and can be less conservative.

\vspace{-0.1in}
\section{Experiments}\label{sec:experiments}
\vspace{-0.1in}
%!TEX root = main.tex
We evaluate our MoE-approach for dialogue management on two benchmark open-domain tasks. The first one is Cornell \citep{danescu2011chameleons}, which consists of conversations between speakers in different movie lines and has a median conversation length of $3$ utterances, the second is the Reddit \citep{ghandeharioun2019approximating}, which is a casual conversation corpus on various topics between users of at least $3$ turns with the median conversation containing $7$ utterances. 
% We first construct a primitive expert that is used for constructing $9$ experts (see Appendix \ref{sec:moe_experts}) and then design a DM with RL to improve users' sentiment \citep{bodie2015role} and engagement \citep{sidner2004look}.

We conduct several experiments to test the efficacy of different parts in the MoE-LM, namely (i) the predictive power and diversity of the primitive, (ii) the quality of experts, and (iii) the overall DM performance. For each metric, we report the mean $\pm$ standard deviation over a evaluation set of $100$ conversations. We also ran an ablation study on $4$ transformer-based MoE-LMs, namely MoE-1, MoE-2, MoE-3- MoE-4, to understand how performance is affected by different model architectures, language encoders, and latent generators. 
MoE-1 and MoE-2 use a simpler architecture, while MoE-3 and MoE-4 use the same encoder architecture as BERT \citep{devlin2018bert}. MoE-1 uses much smaller latent distribution models $\{\mathcal G_i\}$ than MoE-2; MoE-3 uses the pre-trained BERT encoder, while MoE-4 trains that from scratch. Details of these models can be found in Appendix \ref{app:moe_architectures}. 

\begin{figure}[tb]
\begin{small}
\centering
\vspace{-0.1in}
\begin{tabular}{cccc}
\hspace{-0.15in}\subfloat[\scriptsize Transformer Primitive, TSNE]{\label{fig:trans_viz}\includegraphics[trim=3mm 3mm 7mm 3mm,clip,width=0.24\textwidth, height=2.2cm,keepaspectratio=false]{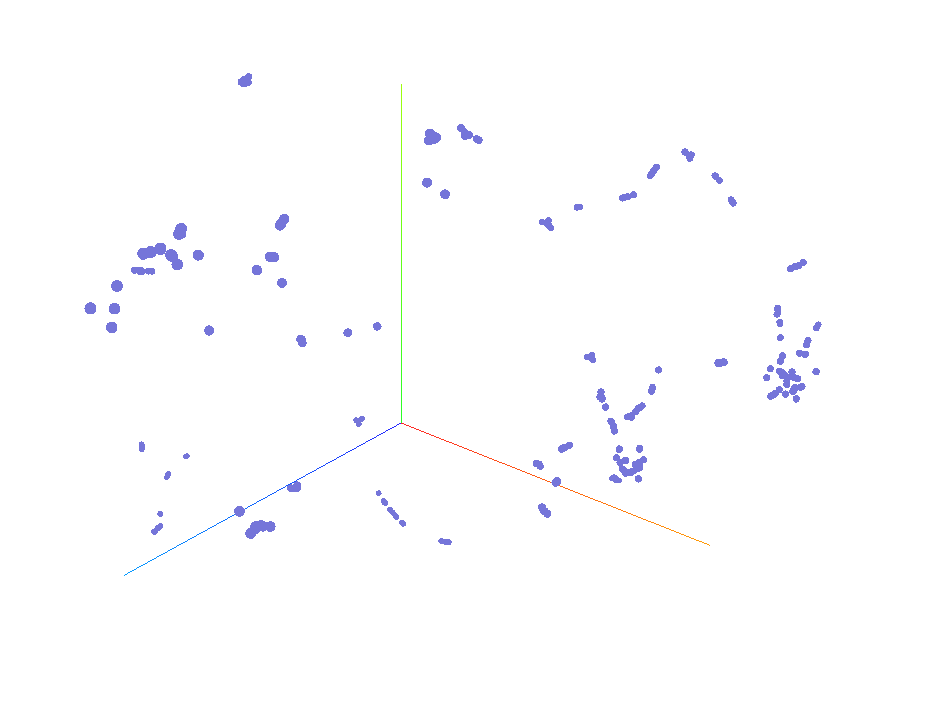}} & 
\hspace{-0.075in}\subfloat[\scriptsize MoE-4 Primitive, TSNE]{\label{fig:moe_viz}\includegraphics[trim=3mm 3mm 7mm 3mm,clip,width=0.24\textwidth, height=2.2cm,keepaspectratio=false]{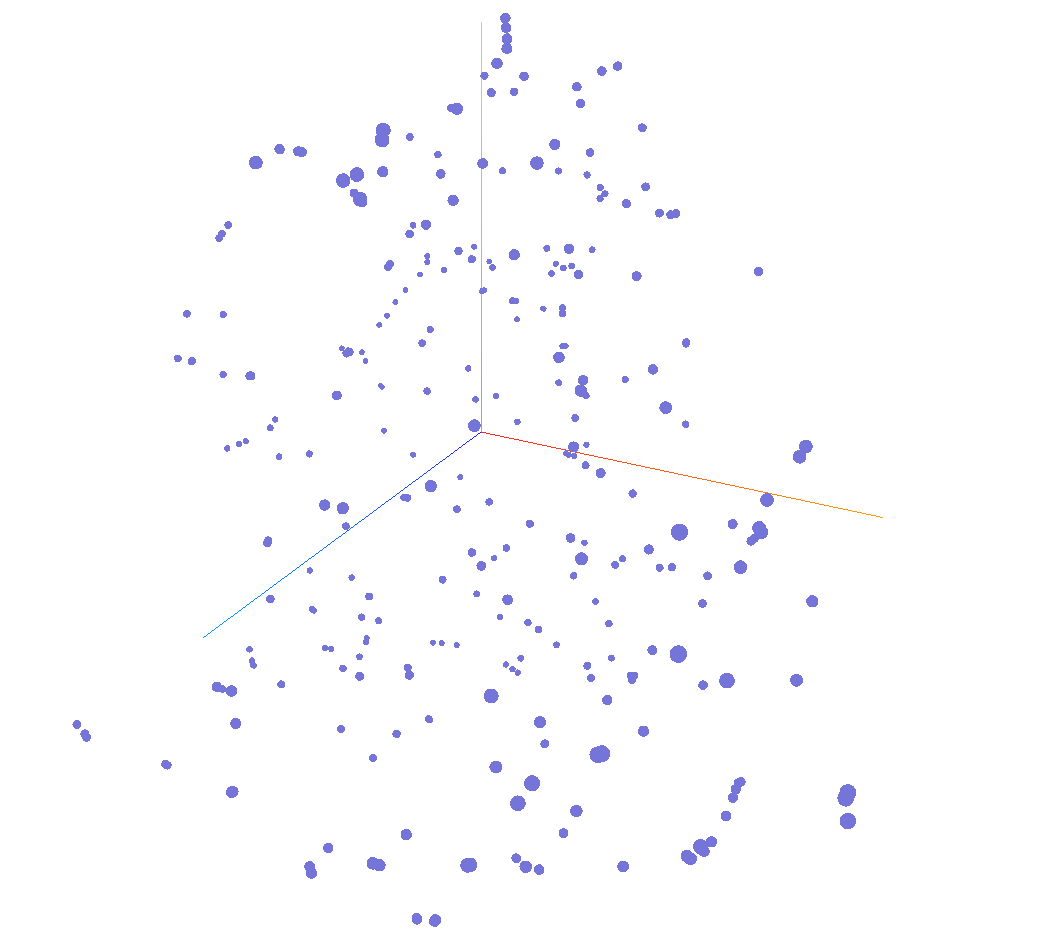}} &
\hspace{-0.075in}\subfloat[\scriptsize MoE-4 Sentiment, PCA]{\label{fig:moe_sentiment}\includegraphics[trim=3mm 3mm 7mm 3mm,clip,width=0.24\textwidth, height=2.2cm,keepaspectratio=false]{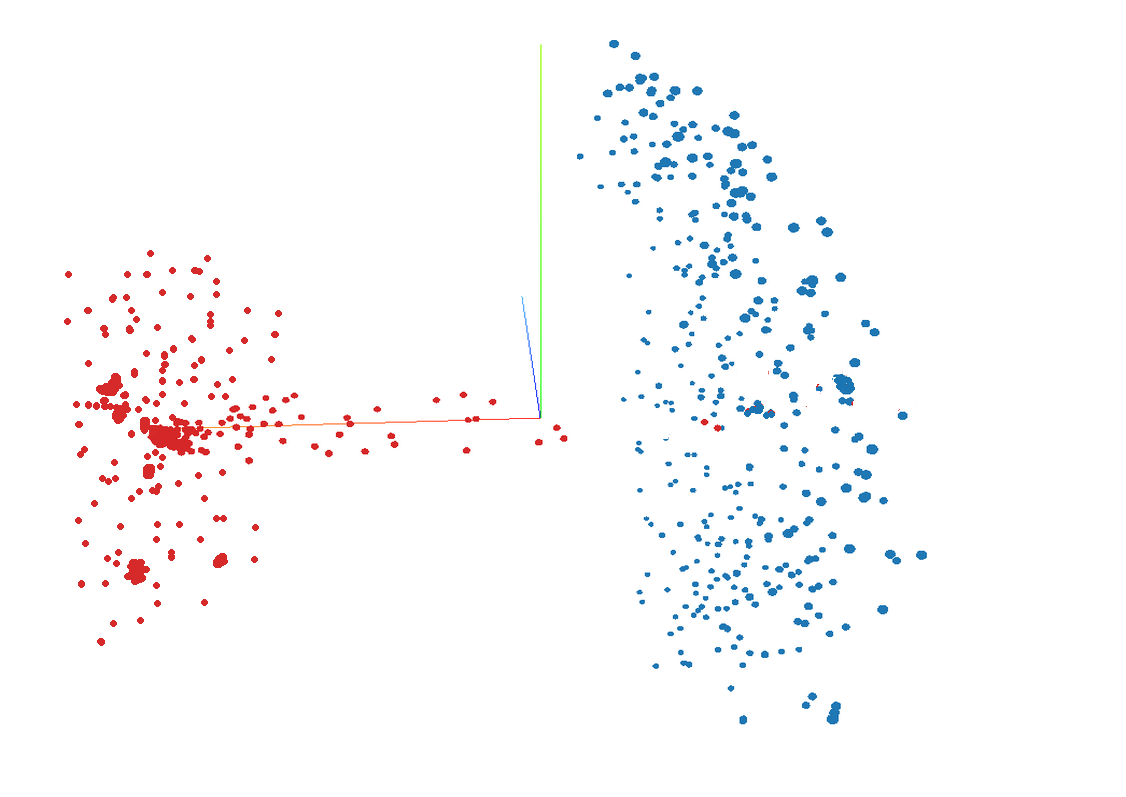}} &
\hspace{-0.075in}\subfloat[\scriptsize MoE-4 Emotion, PCA]{\label{fig:moe_emotion}\includegraphics[trim=3mm 3mm 7mm 3mm,clip,width=0.24\textwidth, height=2.2cm,keepaspectratio=false]{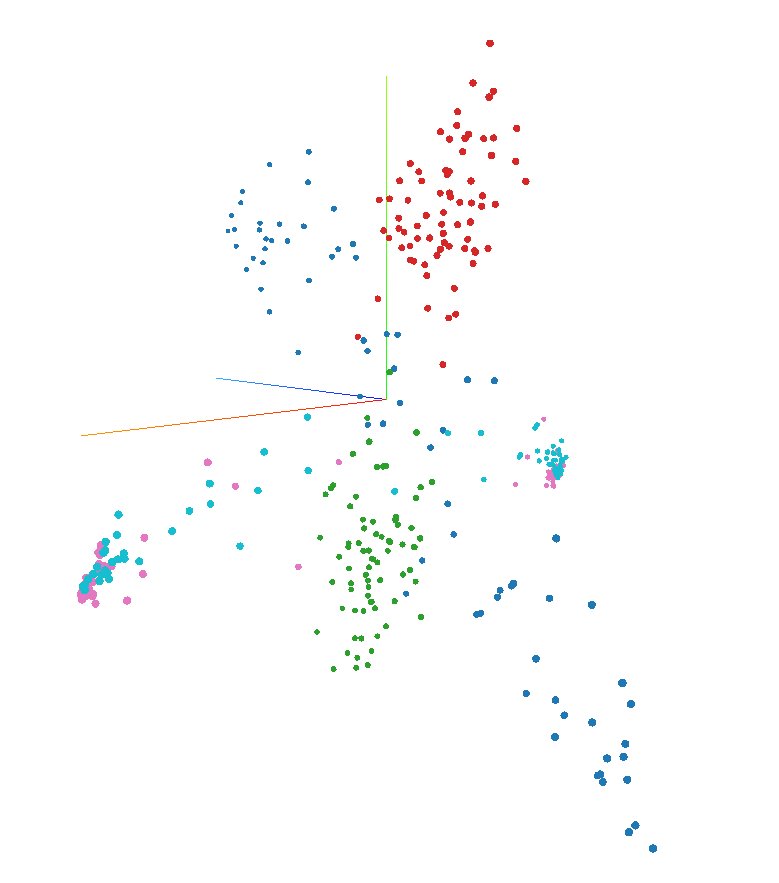}}
\end{tabular}
\vspace{-0.05in}
\caption{\footnotesize
Latent space visualizations.
Figures (a) and (b) compares two  primitive representations. Figures (c) and (d) illustrates how experts (defined by different sentiments and emotions) are represented by latent clusters.}
\label{fig:embedding_vis}
\vspace{-0.15in}
\end{small}
\end{figure}

\vspace{-0.1in}
\paragraph{EXP 1: Comparing Primitive Models}{
We compare the quality of latent representations learned by the $4$ MoE-LMs (via optimizing Eq.~\ref{eq:rep_opt}) and $2$ baselines (standard Transformer \citep{wolf2019huggingface} and VHRED \citep{serban2017hierarchical}). To assess their quality, for each test conversation we generated $25$ utterances and reported the following $3$ metrics: (i) \textbf{Diversity}: The $1-$ Sparsity \citep{hurley2009comparing} of the singular values of the embedded utterances, i.e., $\text{Diversity}(\{\hat Y_i\}):=1 - {\sqrt{d-\|\text{SVD}\|_1/\|\text{SVD}\|_2}}/{\sqrt{d-1}}\in[0,1]$, where $\text{SVD}:=\text{SVD}(\{\Phi_{\text{SE}}(\hat{Y}_i\}_{i=1}^{25})$, and $\Phi_{\text{SE}}$ is a pre-trained sentence encoder (e.g., a USE \citep{cer2018universal}); (ii) \textbf{Dist-$\mathbf{\{1,2,3\}}$} \citep{li2015diversity}: Ratio of unique $\{1,2,3\}$-gram in the generated utterances; (iii) \textbf{Perplexity} \citep{bahl1983maximum}. 
These metrics measure both accuracy and semantic diversity.

The results of the above experiments are reported in Table \ref{tab:phase_1_reddit} and \ref{tab:phase_1_cornell} (Appendix \ref{app:cornell_results}), and sample utterances generated by these LMs can be found in Appendix \ref{app:sample_utterances_1}. In comparisons with the baselines (Transformer and VHRED), generally (i) transformer-based LMs out-perform VHRED due to their attention mechanism that explicitly encodes sequential semantic information, and (ii) the MoE-LMs achieve way better diversity without sacrificing much on accuracy (i.e., the perplexity scores are still quite low). 
Qualitatively, the sample utterances generated the Transformer are closer to the targets than that by MoE-2 and MoE-4, likely because Transformer tends to memorize the corpus \citep{kharitonov2021bpe}. Contrarily,   MoE-LMs generate utterances that have similar contexts with targets but paraphrased or similar structures but different contexts, demonstrating their generalizability.

Among different MoE-LMs, MoE-2 and MoE-4 have the best performances, particularly MoE-4 has better diversity while MoE-2 has lower perplexity. This corroborates with our hypotheses that (i) jointly training the encoder and decoder with Eq.~\ref{eq:rep_opt} seems necessary to encourage semantic diversity (as opposed to using a pre-trained BERT encoder, which maximizes likelihood), (ii) sufficient representation power is necessary for $\mathcal G_0$  to match the posterior distribution $\rho$ in order to capture different semantics in $\mathcal D$. In Fig.~\ref{fig:trans_viz}  and \ref{fig:moe_viz}, we visualize the latent space of $200$ conversation data samples for both Transformer and MoE-4. The latent states of MoE-4 are much more dispersed across the embedding space, implying that most conversations get encoded uniquely. In contrast, the latent space of Transformer has many clusters, suggesting it is more prone to generating similar utterances even with different input conversation and is thus less generalizable.

\begin{table*}[t]
  \begin{minipage}{0.5\textwidth}
\centering
  {\tiny
  \begin{tabular}{|c||c|c|c|c|c|}
  \hline
  Method & Diversity  &	Dist-1 &  Dist-2 & Dist-3	 &  Perplexity	\\
  \hline
 MoE-1 & 0.078 $\pm$ 0.04 &	0.29 & 0.68  & 	0.79 &  28.68 $\pm$ 41.32	\\ \hline
 MoE-2 & 0.15 $\pm$ 0.02  &	0.33 & 0.81  & 0.92	 &  42.14 $\pm$ 35.24	\\ \hline
 MoE-3 & 0.093 $\pm$ 0.02  & 0.31	 &  0.77 & 0.90	 &  44.27 $\pm$ 57.73	\\ \hline
 MoE-4 & \textbf{0.17 $\pm$ 0.01}   &	 \textbf{0.41} & \textbf{0.89}  & \textbf{0.94}	 &  52.12 $\pm$ 18.20	\\ \hline
 Trans. & 0.091 $\pm$ 0.02 &	0.28 & 0.71   & 0.83	 & \textbf{ 20.77 $\pm$ 10.22}	 \\ \hline
 VHRED & 0.11 $\pm$ 0.02 &	0.37 & 0.68  & 0.80	 &  88.61 $\pm$	23.55 \\ \hline	
 \end{tabular}
  \caption{Accuracy (Perplexity) and Diversity of Language Primitive Experts Trained with Reddit.}
  \label{tab:phase_1_reddit}
  }
  \end{minipage}\,\,\,\,\,\,\,\,\begin{minipage}{0.55\textwidth}
\centering
  {\tiny
  \begin{tabular}{|c||c|c|c|}
  \hline
  Method &  User Tot. Sent. & User Sent. Trans.	 &  GPT-Perplexity	\\
  \hline
  MoE-4 Primitive & -0.05 $\pm$ 0.07 & 0.05$\pm$ 0.09 & 81.17 $\pm$ 43.41	\\ \hline
 MoE-4 DQN & 0.04 $\pm$ 0.10 & 0.02$\pm$ 0.16 & 79.43$\pm$ 45.70	\\ \hline
 MoE-4 CQL & 0.09$\pm$0.16 & 0.08$\pm$ 0.17 & \textbf{75.48$\pm$ 47.13}	\\ \hline
 MoE-4 MBRL & \textbf{0.12$\pm$ 0.15} & \textbf{0.10 $\pm$ 0.13} & 82.75 $\pm$ 42.72	\\ \hline
 Trans. RL & -0.26$\pm$ 0.12 & -0.11$\pm$ 0.14 & 168.80 $\pm$ 75.35\\ \hline
 Trans. KLC & 0.03 $\pm$ 0.09 & -0.03 $\pm$ 0.05 & 108.73$\pm$ 68.21\\ \hline
 Trans. VHRL & 0.05 $\pm$ 0.11 & 0.01 $\pm$ 0.11 & 135.47$\pm$ 55.78\\ \hline
 \end{tabular}
  \caption{Performance (w.r.t. Avg. User Satisfaction in Conversation) of RL-based DM Trained with Reddit.}
  \label{tab:phase_3_reddit}
  }
  \end{minipage}
\end{table*}
}

\vspace{-0.1in}
\paragraph{EXP 2: Quality of Experts}{
We compare the performance of experts learned by the $4$ MoE-LMs (where experts are separately trained by optimizing Eq.~\ref{eq:rep_opt_ft}) and $2$ baselines (WD \citep{holtzman2018learning} and PPLM \citep{dathathri2019plug}). To study the sub-optimality gap in Corollary \ref{coro:expert}, we also include the performance of Transformer-based expert end-to-end LMs that are individually optimized with REINFORCE \citep{li2016deep},  using the expert labels $\{\ell_i\}$ as rewards. 
We use the following label functions to define the intents of experts: (i) $\ell_{\text{pos-sent}}(Y)$, $\ell_{\text{neg-sent}}(Y)$, $\ell_{\text{joy}}(Y)$, $\ell_{\text{optimism}}(Y)$, $\ell_{\text{anger}}(Y)$, $\ell_{\text{sadness}}(Y)$ quantify $6$ different sentiment tones and are constructed by a RoBERTa-based sentiment
detector \citep{liao2021improved} that predicts whether an utterance is of positive or negative sentiment, and whether it falls into any of the $4$ more-refined emotions: \{joy, optimisim, sadness, anger\}; (ii) $\ell_{\text{sent-coh}}(\mathbf{X},Y)$ measures \emph{empathy}, i.e., bot's sentiment coherence with user's; 
(iii) $\ell_{\text{question}}(Y)$ outputs $1$ when a bot question is detected and $0$ otherwise; (iv) $\ell_{\text{exp}}(\mathbf{X},Y)$ quantifies \emph{exploration}, i.e., the tendency to avoid repetitive contexts. See Appendix \ref{app:rewards} for details.

The results of the above experiments are reported in Table \ref{tab:phase_2_reddit} and \ref{tab:phase_2_cornell} (Appendix \ref{app:cornell_results}), with sample utterances reported in Appendix \ref{app:sample_utterances_2_1} to \ref{app:sample_utterances_2_7}. Compared with the baseline LMs, generally the experts created under the MoE-LM framework, especially under MoE-2 and MoE-4, better capture all different language intents (where WD and PPL appear to capture negative sentiments and emotions much more effectively than behaviors), demonstrating the efficacy of our approach which constructs specialized experts on a diverse language space via reward maximization (instead of weighted MLE).

Similar to the ablation study in EXP 1, all the experts associated with MoE-2 and MoE-4 are also among the best ones in capturing language intents. Interestingly, with the Reddit data the experts in MoE-4 perform the best, while with much less data (Cornell) the best experts are built upon the simpler MoE-2 architecture. We conjecture this difference is due to over-fitting issues faced by the larger LMs (MoE-4) when there is insufficient data for expert fine-tuning. In Fig.~\ref{fig:moe_sentiment} and~\ref{fig:moe_emotion} we visualize the latent space of the sentiment-based experts in MoE-4, each with $200$ conversation data samples. Notice that the sentiment experts' latent distributions are clearly separated (because positive and negative sentiments have opposite behaviors), while the emotion expert's latent distribution have more gradual separations and even some overlaps (because e.g., joy versus optimism are quite similar, while joy versus anger are quite different).
This validates our MoE-LM represents different behaviors in separate regions of the latent space and justifies our structural prior of modeling each expert as a specialized version of the primitive LM, whose latent distribution focuses on particular regions.

\begin{table*}[t]
  \begin{minipage}{\textwidth}
\centering
  {\tiny
  \begin{tabular}{|c||c|c|c|c|c|c|c|c|c|}
  \hline
  Method & Question  &	Exploration &  Positive Sent. & Negative Sent.	& Sent. Coherence &  Joy & Optimism &	Anger & Sadness\\
  \hline
 MoE-1 & 0.75 $\pm$ 0.14   &0.49 $\pm$  0.12	 & 1.02 $\pm$ 0.35 &  0.40 $\pm$ 0.33	 & 0.51 $\pm$  0.48 & 1.12 $\pm$ 0.32  & -0.19 $\pm$ 0.55 & 0.47 $\pm$ 0.56 & 1.01 $\pm$ 0.67	\\ \hline
 MoE-2 & \textbf{0.96 $\pm$ 0.18}  & \textbf{0.58 $\pm$ 0.20}	 &  3.41 $\pm$ 0.56 & 2.08	$\pm$ 0.71 & 0.55 $\pm$ 0.44  & 2.02 $\pm$ 0.65 & 1.27 $\pm$ 0.41  & 1.32 $\pm$ 0.69 & 2.09 $\pm$ 0.99	\\ \hline
 MoE-3 &  0.43 $\pm$ 0.27 & 0.50 $\pm$ 0.33 & 1.03 $\pm$ 0.32 & 1.05 $\pm$ 0.29 & 0.62 $\pm$ 0.41 & 1.17 $\pm$ 0.42 & 0.42 $\pm$ 0.51 & 0.28 $\pm$ 0.32 & 1.15 $\pm$ 0.27	\\ \hline
 MoE-4 & 0.91 $\pm$ 0.25 & 0.53	 $\pm$ 0.40 & \textbf{3.34 $\pm$ 0.77} & \textbf{2.12 $\pm$ 0.64} & 0.50 $\pm$ 0.36 & \textbf{2.28 $\pm$ 1.59} & \textbf{1.59 $\pm$ 0.99} & \textbf{2.01 $\pm$ 1.32} & \textbf{2.27 $\pm$ 0.45}	\\ \hline
 WD & 0.0 $\pm$ 0.0 & 0.16 $\pm$ 0.29	 &  -0.96 $\pm$ 1.86 & 	1.36 $\pm$ 2.02 & \textbf{0.69 $\pm$ 0.23} &  -0.84 $\pm$ 0.40 & -1.77 $\pm$ 1.36 & 0.92 $\pm$ 0.35  & 1.99 $\pm$ 1.5  \\ \hline
 PPLM & 0.21 $\pm$ 0.34 & 0.56 $\pm$ 0.32	 &  0.46 $\pm$ 0.21 &  0.55 $\pm$ 0.09	 & 0.64 $\pm$ 0.27  & 0.39 $\pm$ 0.15 & 0.41 $\pm$ 0.48 & 0.69 $\pm$ 0.22 & 1.00 $\pm$ 0.21\\ \hline\hline
  Trans. RL* & 0.99 $\pm$ 0.23 & 0.54 $\pm$ 0.18	 & 3.53 $\pm$ 1.64  & 1.89 $\pm$ 1.20	 & 0.72 $\pm$ 0.30 & 2.88 $\pm$ 2.36 & 1.80 $\pm$ 2.39 & 1.62 $\pm$ 2.39	& 2.35 $\pm$ 1.47 \\ \hline	
 \end{tabular}
  \caption{Quality of each expert PPLM trained on Reddit dataset w.r.t. its trained label.}
  \label{tab:phase_2_reddit}
  }
  \end{minipage}
\end{table*}
}

\vspace{-0.1in}
\paragraph{EXP 3: MoE-RL Against DialoGPT Simulated Users}{
We compare the dialogue management performance of MoE-LM, for which their DMs $\mu$ are trained with different methods (BC, DQN, CQL, MBRL), with $3$ baselines (REINFORCE \citep{li2016deep}, KL-control \citep{jaques2019way}, and VHRL \citep{saleh2020hierarchical}). According to the results on expert quality in EXP2, we pick the MoE-2 and MoE-4 frameworks for the Cornell and Reddit tasks respectively. For systematic evaluation, we perform the experiment by having these RL agents interact with a DialoGPT \citep{zhang2019dialogpt} simulated user environment (a large LM capable of conducting natural open-domain conversations), for a maximum of $5$ turns. The DM task is to maximize total user satisfaction in the conversation level, which is measured by both (i) user's overall sentiment, and (ii) user's sentiment transition. To construct an immediate reward that serves as a surrogate for user satisfaction, we set $r(\mathbf{X},a, X_+)= \lambda_1\ell_{\text{sent}}(X_+) + \lambda_2 (\ell_{\text{sent}}(X_+)-\frac{1-\gamma}{1-\gamma^{L}}\sum_{l=0}^{L-1}\gamma^l \ell_{\text{sent}}(X_l))$, where the linear combination weights $(\lambda_1,\lambda_2)=(0.75,0.25)$ correlate with  \citet{ghandeharioun2019approximating}, and $\ell_{\text{sent}}(X)$ is the same RoBerTa-based sentiment labeler as in EXP2, which assigns a score from $[-1,1]$ that is proportional to the positive sentiment and inversely proportional to the negative sentiment prediction probabilities.
To ensure the baseline RL DM methods can also possess certain bot-level features, e.g., question, positive sentiment, etc., besides the above RL reward for user satisfaction we also optimize a linear combination of bot-based rewards when training the baseline models, see Appendix B of \citet{saleh2020hierarchical} for more details. Since the DM problem lasts at most $5$ turns, we use this as the effective horizon and set $\gamma=1 - 1/5 = 0.8$. 
To evaluate the fluency of our LMs, we also report the GPT-perplexity, which compute the perplexity score of the final utterance w.r.t. a DialoGPT LM. 

The results of the above experiments are reported in Table \ref{tab:phase_3_reddit} and \ref{tab:phase_3_cornell} (Appendix \ref{app:cornell_results}), with sample utterances reported in Appendix \ref{app:sample_utterances_3}. Our experiments show that MoE-LMs outperform most baseline methods on DM performance. We attribute this finding to two factors: (i) MoE-MDP restricts the action space into a smaller set of candidate utterances generated by experts (whose qualities are validated in EXP2), the corresponding RL problem then becomes simpler and requires less data (especially in Cornell) to solve. 
(ii) Unlike the baseline RL methods, which need to optimize both bot-and-user signals, the MoE DM agents focus on optimizing the user satisfaction goal and are therefore more effective. Interestingly, our MoE-LMs also have lower (better) GPT-perplexity scores than other methods. This may be due to the fact that MoE-LM uses pre-trained encoder and decoder from the primitive LM, which are optimized for generalization and accuracy, while other RL methods may distort their language representations to create utterances that maximize reward but become less natural. Among different MoE-RL strategies, MBRL, which first learns a user utterance model (the model uses the same encoder from the primitive LM and learns a separate decoder for user-utterance prediction) and does $Q$-learning, performs the best. CQL moderately improves upon the primitive LM (behavior policy), while DQN appears to be less effective in solving the offline RL problem.

}

% \paragraph{EXP 4: MoE-LM User Interactive Evaluation}{
% \begin{table*}[t]
%  \begin{minipage}{0.5\textwidth}
% \centering
%   {\scriptsize
%   \begin{tabular}{|c||c|c|c|c|c|}
%   \hline
%   Method & Diversity  &	Fluency &  Sentiment & Curiosity	 &  Related	\\
%   \hline
%  MoE-2 & $\pm$ & $\pm$	 & $\pm$ & $\pm$ & $\pm$	\\ \hline
%  MoE-4 & $\pm$ & $\pm$ & $\pm$ & $\pm$ & $\pm$	\\ \hline
%  Transformer & $\pm$ & $\pm$ & $\pm$  & $\pm$ & $\pm$  	 \\ \hline
%  PPLM & $\pm$ &	$\pm$ & $\pm$  & $\pm$ & $\pm$ 	 \\ \hline	
%  Transformer RL & $\pm$ & $\pm$ & $\pm$ & $\pm$ & $\pm$\\ \hline
%  Transformer HRL & $\pm$ & $\pm$ & $\pm$ & $\pm$ & $\pm$\\ \hline
%  \end{tabular}
%   \caption{Human Eval. on LMs Trained with Cornell Dataset.}
%   \label{tab:phase_4_cornell}
%   }
%   \end{minipage}\,\,\,
%   \begin{minipage}{0.5\textwidth}
% \centering
%   {\scriptsize
%   \begin{tabular}{|c||c|c|c|c|c|}
%   \hline
%   Method & Diversity  &	Fluency &  Sentiment & Curiosity	 &  Related	\\
%   \hline
%  MoE-2 & $\pm$ & $\pm$	 & $\pm$ &$\pm$	 &$\pm$	\\ \hline
%  MoE-4 & $\pm$ & $\pm$ & $\pm$ & $\pm$	 & $\pm$	\\ \hline
%  Transformer & $\pm$ & $\pm$	 & $\pm$  & $\pm$	 &  $\pm$	 \\ \hline
%  PPLM & $\pm$ &	$\pm$ & $\pm$  & $\pm$	 &  $\pm$	 \\ \hline	
%  Transformer RL & $\pm$ & $\pm$ & $\pm$ & $\pm$ & $\pm$\\ \hline
%  Transformer HRL & $\pm$ & $\pm$ & $\pm$ & $\pm$ & $\pm$\\ \hline
%  \end{tabular}
%   \caption{Human Eval. on LMs Trained with Reddit Dataset.}
%   \label{tab:phase_4_reddit}
%   }
%   \end{minipage}
% \end{table*}
% }

\vspace{-0.1in}
\section{Concluding Remarks}\label{sec:conclusion}
\vspace{-0.1in}
%!TEX root = main.tex
We developed a mixture-of-expert (MoE) approach for RL-based dialogue management (DM). Our MoE language model (MoE-LM) comprises of three main components: (i) a LM that can generate diverse semantics for conversation histories, (ii) a number of specialized LMs (or experts) that can produce utterances corresponding to a particular attribute or intent, and (iii) a RL-based DM that performs dialogue planning with the utterances generated by the experts. To understand how well our MoE approach generates diverse and sensible utterances, and solves DM problems, we evaluated it using two open-domain dialogue tasks and compared it with SOTA baselines. Our results showed that MoE-LM (i) improves diversity of text generation, (ii) can generate utterances with specific intents, and (iii) yields better overall performance. We consider our work as a step forward in creating steerable LMs that possess different intents and in training RL-based DMs that can carry on rich conversations. Future work includes improving the language representation with information-theoretic approaches, fine-tuning the experts based on the DM objective, extending the RL agent to track users' behaviors (via abstract belief states) and plan upon them, and evaluating our MoE-LM on more realistic problems, such as information retrieval, recommendation, and negotiation.

\newpage

\section*{Checklist}
% https://neurips.cc/public/guides/PaperChecklist

\begin{enumerate}

\item For all authors...
\begin{enumerate}
  \item Do the main claims made in the abstract and introduction accurately reflect the paper's contributions and scope?
    \answerYes{}
  \item Did you describe the limitations of your work?
    \answerYes{See the summary (Section~\ref{sec:conclusion}).}
  \item Did you discuss any potential negative societal impacts of your work?
    \answerNA{In this work we proposed an abstract approach of designing dialogue managers, it is unrelated to any negative impact applications.}
  \item Have you read the ethics review guidelines and ensured that your paper conforms to them?
    \answerYes{}
\end{enumerate}

\item If you are including theoretical results...
\begin{enumerate}
  \item Did you state the full set of assumptions of all theoretical results?
    \answerYes{For each result, Section~\ref{sec:rep_l} and Section~\ref{sec:expert_l} either state the assumptions directly or refers to corresponding references.}
\item Did you include complete proofs of all theoretical results?
    \answerNA{All theoretical are cited with corresponding references.}
\end{enumerate}

\item If you ran experiments...
\begin{enumerate}
  \item Did you include the code, data, and instructions needed to reproduce the main experimental results (either in the supplemental material or as a URL)?
    \answerNo{Unfortunately, due to IP concerns our legal team does not allow code releases at this point.}
  \item Did you specify all the training details (e.g., data splits, hyperparameters, how they were chosen)?
    \answerYes{General details in the beginning of Section~\ref{sec:experiments}, and specific details per experiment in Appendix~\ref{app:exp_details}.}
\item Did you report error bars (e.g., with respect to the random seed after running experiments multiple times)?
    \answerYes{In Talbes~\ref{tab:phase_1_cornell},\ref{tab:phase_1_reddit},\ref{tab:phase_2_cornell}, \ref{tab:phase_2_reddit}, \ref{tab:phase_3_cornell}, \ref{tab:phase_3_reddit}, , we provide not only the mean result but also the standard deviation.}
        \item Did you include the total amount of compute and the type of resources used (e.g., type of GPUs, internal cluster, or cloud provider)?
    \answerYes{In the beginning of Appendix~\ref{app:exp_details}.}
\end{enumerate}

\item If you are using existing assets (e.g., code, data, models) or curating/releasing new assets...
\begin{enumerate}
  \item If your work uses existing assets, did you cite the creators?
    \answerNA{}
  \item Did you mention the license of the assets?
    \answerNA{}
  \item Did you include any new assets either in the supplemental material or as a URL?
    \answerNA{}
  \item Did you discuss whether and how consent was obtained from people whose data you're using/curating?
    \answerNA{}
  \item Did you discuss whether the data you are using/curating contains personally identifiable information or offensive content?
    \answerNA{}
\end{enumerate}

\item If you used crowdsourcing or conducted research with human subjects...
\begin{enumerate}
  \item Did you include the full text of instructions given to participants and screenshots, if applicable?
    \answerNA{}
  \item Did you describe any potential participant risks, with links to Institutional Review Board (IRB) approvals, if applicable?
    \answerNA{}
  \item Did you include the estimated hourly wage paid to participants and the total amount spent on participant compensation?
    \answerNA{}
\end{enumerate}

\end{enumerate}

\newpage
%% The file named.bst is a bibliography style file for BibTeX 0.99c
\bibliography{uai22}

\begin{thebibliography}{62}
\providecommand{\natexlab}[1]{#1}
\providecommand{\url}[1]{\texttt{#1}}
\expandafter\ifx\csname urlstyle\endcsname\relax
  \providecommand{\doi}[1]{doi: #1}\else
  \providecommand{\doi}{doi: \begingroup \urlstyle{rm}\Url}\fi

\bibitem[Ajay et~al.(2020)Ajay, Kumar, Agrawal, Levine, and
  Nachum]{ajay2020opal}
A.~Ajay, A.~Kumar, P.~Agrawal, S.~Levine, and O.~Nachum.
\newblock Opal: Offline primitive discovery for accelerating offline
  reinforcement learning.
\newblock \emph{arXiv preprint arXiv:2010.13611}, 2020.

\bibitem[Asadi and Williams(2016)]{asadi2016sample}
K.~Asadi and J.~Williams.
\newblock Sample-efficient deep reinforcement learning for dialog control.
\newblock \emph{arXiv preprint arXiv:1612.06000}, 2016.

\bibitem[Bahl et~al.(1983)Bahl, Jelinek, and Mercer]{bahl1983maximum}
L.~Bahl, F.~Jelinek, and R.~Mercer.
\newblock A maximum likelihood approach to continuous speech recognition.
\newblock \emph{IEEE transactions on pattern analysis and machine
  intelligence}, \penalty0 (2):\penalty0 179--190, 1983.

\bibitem[Burgess et~al.(2018)Burgess, Higgins, Pal, Matthey, Watters,
  Desjardins, and Lerchner]{burgess2018understanding}
C.~Burgess, I.~Higgins, A.~Pal, L.~Matthey, N.~Watters, G.~Desjardins, and
  A.~Lerchner.
\newblock Understanding disentangling in $\beta$-vae.
\newblock \emph{arXiv preprint arXiv:1804.03599}, 2018.

\bibitem[Cer et~al.(2018)Cer, Yang, Kong, Hua, Limtiaco, John, Constant,
  Guajardo-Cespedes, Yuan, Tar, et~al.]{cer2018universal}
D.~Cer, Y.~Yang, S.~Kong, N.~Hua, N.~Limtiaco, R.~John, N.~Constant,
  M.~Guajardo-Cespedes, S.~Yuan, C.~Tar, et~al.
\newblock Universal sentence encoder.
\newblock \emph{arXiv preprint arXiv:1803.11175}, 2018.

\bibitem[Chen et~al.(2018)Chen, Li, Cho, and Bowman]{chen2018stable}
Y.~Chen, V.~Li, K.~Cho, and S.~Bowman.
\newblock A stable and effective learning strategy for trainable greedy
  decoding.
\newblock \emph{arXiv preprint arXiv:1804.07915}, 2018.

\bibitem[Chien and Kuo(2019)]{chien2019markov}
J.~Chien and C.~Kuo.
\newblock Markov recurrent neural network language model.
\newblock In \emph{2019 IEEE Automatic Speech Recognition and Understanding
  Workshop (ASRU)}, pages 807--813. IEEE, 2019.

\bibitem[D et~al.(2016)D, Narasimhan, Saeedi, and
  Tenenbaum]{kulkarni2016hierarchical}
T.~Kulkarni D, K.~Narasimhan, A.~Saeedi, and J.~Tenenbaum.
\newblock Hierarchical deep reinforcement learning: Integrating temporal
  abstraction and intrinsic motivation.
\newblock \emph{Advances in neural information processing systems}, 29, 2016.

\bibitem[Danescu-Niculescu-Mizil and Lee(2011)]{danescu2011chameleons}
C.~Danescu-Niculescu-Mizil and L.~Lee.
\newblock Chameleons in imagined conversations: A new approach to understanding
  coordination of linguistic style in dialogs.
\newblock \emph{arXiv preprint arXiv:1106.3077}, 2011.

\bibitem[Dathathri et~al.(2019)Dathathri, Madotto, Lan, Hung, Frank, Molino,
  Yosinski, and Liu]{dathathri2019plug}
S.~Dathathri, A.~Madotto, J.~Lan, J.~Hung, E.~Frank, P.~Molino, J.~Yosinski,
  and R.~Liu.
\newblock Plug and play language models: A simple approach to controlled text
  generation.
\newblock \emph{arXiv preprint arXiv:1912.02164}, 2019.

\bibitem[den Oord et~al.(2018)den Oord, Li, and Vinyals]{van2018representation}
A.~Van den Oord, Y.~Li, and O.~Vinyals.
\newblock Representation learning with contrastive predictive coding.
\newblock \emph{arXiv e-prints}, pages arXiv--1807, 2018.

\bibitem[Devlin et~al.(2018)Devlin, Chang, Lee, and Toutanova]{devlin2018bert}
J.~Devlin, M.~Chang, K.~Lee, and K.~Toutanova.
\newblock Bert: Pre-training of deep bidirectional transformers for language
  understanding.
\newblock \emph{arXiv preprint arXiv:1810.04805}, 2018.

\bibitem[Fatemi et~al.(2016)Fatemi, Asri, Schulz, He, and
  Suleman]{fatemi2016policy}
M.~Fatemi, L.~Asri, H.~Schulz, J.~He, and K.~Suleman.
\newblock Policy networks with two-stage training for dialogue systems.
\newblock \emph{arXiv preprint arXiv:1606.03152}, 2016.

\bibitem[Ficler and Goldberg(2017)]{ficler2017controlling}
J.~Ficler and Y.~Goldberg.
\newblock Controlling linguistic style aspects in neural language generation.
\newblock \emph{arXiv preprint arXiv:1707.02633}, 2017.

\bibitem[Floridi and Chiriatti(2020)]{floridi2020gpt}
L.~Floridi and M.~Chiriatti.
\newblock {GPT}-3: Its nature, scope, limits, and consequences.
\newblock \emph{Minds and Machines}, 30\penalty0 (4):\penalty0 681--694, 2020.

\bibitem[Ga{\v{s}}i{\'c} et~al.(2011)Ga{\v{s}}i{\'c},
  Jur{\v{c}}{\'\i}{\v{c}}ek, Thomson, Yu, and Young]{gavsic2011line}
M.~Ga{\v{s}}i{\'c}, F.~Jur{\v{c}}{\'\i}{\v{c}}ek, B.~Thomson, K.~Yu, and
  S.~Young.
\newblock On-line policy optimisation of spoken dialogue systems via live
  interaction with human subjects.
\newblock In \emph{2011 IEEE Workshop on Automatic Speech Recognition \&
  Understanding}, pages 312--317. IEEE, 2011.

\bibitem[Ghandeharioun et~al.(2019)Ghandeharioun, Shen, Jaques, Ferguson,
  Jones, Lapedriza, and Picard]{ghandeharioun2019approximating}
A.~Ghandeharioun, J.~Shen, N.~Jaques, C.~Ferguson, N.~Jones, A.~Lapedriza, and
  R.~Picard.
\newblock Approximating interactive human evaluation with self-play for
  open-domain dialog systems.
\newblock \emph{Advances in Neural Information Processing Systems}, 32, 2019.

\bibitem[Greensmith et~al.(2004)Greensmith, Bartlett, and
  Baxter]{greensmith2004variance}
E.~Greensmith, P.~Bartlett, and J.~Baxter.
\newblock Variance reduction techniques for gradient estimates in reinforcement
  learning.
\newblock \emph{Journal of Machine Learning Research}, 5\penalty0 (9), 2004.

\bibitem[Gu et~al.(2017)Gu, Cho, and Li]{gu2017trainable}
J.~Gu, K.~Cho, and V.~Li.
\newblock Trainable greedy decoding for neural machine translation.
\newblock \emph{arXiv preprint arXiv:1702.02429}, 2017.

\bibitem[Haarnoja et~al.(2018)Haarnoja, Zhou, Abbeel, and
  Levine]{haarnoja2018soft}
T.~Haarnoja, A.~Zhou, P.~Abbeel, and S.~Levine.
\newblock Soft actor-critic: Off-policy maximum entropy deep reinforcement
  learning with a stochastic actor.
\newblock \emph{ICML}, 2018.

\bibitem[Hafner et~al.(2020)Hafner, Lillicrap, Norouzi, and
  Ba]{hafner2020mastering}
D.~Hafner, T.~Lillicrap, M.~Norouzi, and J.~Ba.
\newblock Mastering atari with discrete world models.
\newblock \emph{arXiv preprint arXiv:2010.02193}, 2020.

\bibitem[Hancock et~al.(2019)Hancock, Bordes, Mazare, and
  Weston]{hancock2019learning}
B.~Hancock, A.~Bordes, P.~Mazare, and J.~Weston.
\newblock Learning from dialogue after deployment: Feed yourself, chatbot!
\newblock \emph{arXiv preprint arXiv:1901.05415}, 2019.

\bibitem[Holtzman et~al.(2018)Holtzman, Buys, Forbes, Bosselut, Golub, and
  Choi]{holtzman2018learning}
A.~Holtzman, J.~Buys, M.~Forbes, A.~Bosselut, D.~Golub, and Y.~Choi.
\newblock Learning to write with cooperative discriminators.
\newblock \emph{arXiv preprint arXiv:1805.06087}, 2018.

\bibitem[Hurley and Rickard(2009)]{hurley2009comparing}
N.~Hurley and S.~Rickard.
\newblock Comparing measures of sparsity.
\newblock \emph{IEEE Transactions on Information Theory}, 55\penalty0
  (10):\penalty0 4723--4741, 2009.

\bibitem[Jaques et~al.(2019)Jaques, Ghandeharioun, Shen, Ferguson, Lapedriza,
  Jones, Gu, and Picard]{jaques2019way}
N.~Jaques, A.~Ghandeharioun, J.~Shen, C.~Ferguson, A.~Lapedriza, N.~Jones,
  S.~Gu, and R.~Picard.
\newblock Way off-policy batch deep reinforcement learning of implicit human
  preferences in dialog.
\newblock \emph{arXiv:1907.00456}, 2019.

\bibitem[Kharitonov et~al.(2021)Kharitonov, Baroni, and
  Hupkes]{kharitonov2021bpe}
E.~Kharitonov, M.~Baroni, and D.~Hupkes.
\newblock How bpe affects memorization in transformers.
\newblock \emph{arXiv preprint arXiv:2110.02782}, 2021.

\bibitem[Kumar et~al.(2020)Kumar, Zhou, Tucker, and
  Levine]{kumar2020conservative}
A.~Kumar, A.~Zhou, G.~Tucker, and S.~Levine.
\newblock Conservative q-learning for offline reinforcement learning.
\newblock \emph{Advances in Neural Information Processing Systems},
  33:\penalty0 1179--1191, 2020.

\bibitem[Levin and Pieraccini(1997)]{levin1997stochastic}
E.~Levin and R.~Pieraccini.
\newblock A stochastic model of computer-human interaction for learning
  dialogue strategies.
\newblock In \emph{Eurospeech}, volume~97, pages 1883--1886. Citeseer, 1997.

\bibitem[Li et~al.(2015)Li, Galley, Brockett, Gao, and Dolan]{li2015diversity}
J.~Li, M.~Galley, C.~Brockett, J.~Gao, and B.~Dolan.
\newblock A diversity-promoting objective function for neural conversation
  models.
\newblock \emph{arXiv preprint arXiv:1510.03055}, 2015.

\bibitem[Li et~al.(2016)Li, Monroe, Ritter, Galley, Gao, and
  Jurafsky]{li2016deep}
J.~Li, W.~Monroe, A.~Ritter, M.~Galley, J.~Gao, and D.~Jurafsky.
\newblock Deep reinforcement learning for dialogue generation.
\newblock \emph{arXiv preprint arXiv:1606.01541}, 2016.

\bibitem[Li et~al.(2017)Li, Monroe, Shi, Jean, Ritter, and
  Jurafsky]{li2017adversarial}
J.~Li, W.~Monroe, T.~Shi, S.~Jean, A.~Ritter, and D.~Jurafsky.
\newblock Adversarial learning for neural dialogue generation.
\newblock \emph{arXiv preprint arXiv:1701.06547}, 2017.

\bibitem[Li et~al.(2019)Li, Kiseleva, and Rijke]{li2019dialogue}
Z.~Li, J.~Kiseleva, and M.~De Rijke.
\newblock Dialogue generation: From imitation learning to inverse reinforcement
  learning.
\newblock In \emph{Proceedings of the AAAI Conference on Artificial
  Intelligence}, volume~33, pages 6722--6729, 2019.

\bibitem[Liao et~al.(2021)Liao, Zeng, Yin, and Wei]{liao2021improved}
W.~Liao, B.~Zeng, X.~Yin, and P.~Wei.
\newblock An improved aspect-category sentiment analysis model for text
  sentiment analysis based on roberta.
\newblock \emph{Applied Intelligence}, 51\penalty0 (6):\penalty0 3522--3533,
  2021.

\bibitem[Park et~al.(2018)Park, Cho, and Kim]{park2018hierarchical}
Y.~Park, J.~Cho, and G.~Kim.
\newblock A hierarchical latent structure for variational conversation
  modeling.
\newblock \emph{arXiv preprint arXiv:1804.03424}, 2018.

\bibitem[Peng et~al.(2020)Peng, Zhu, Li, Li, Li, Zeng, and Gao]{peng2020few}
B.~Peng, C.~Zhu, C.~Li, X.~Li, J.~Li, M.~Zeng, and J.~Gao.
\newblock Few-shot natural language generation for task-oriented dialog.
\newblock \emph{arXiv preprint arXiv:2002.12328}, 2020.

\bibitem[Razavi et~al.(2019)Razavi, van~den Oord, and
  Vinyals]{razavi2019generating}
A.~Razavi, A.~van~den Oord, and O.~Vinyals.
\newblock Generating diverse high-resolution images with vq-vae.
\newblock 2019.

\bibitem[Saleh et~al.(2020)Saleh, Jaques, Ghandeharioun, Shen, and
  Picard]{saleh2020hierarchical}
A.~Saleh, N.~Jaques, A.~Ghandeharioun, J.~Shen, and R.~Picard.
\newblock Hierarchical reinforcement learning for open-domain dialog.
\newblock In \emph{Proceedings of the AAAI Conference on Artificial
  Intelligence}, volume~34, pages 8741--8748, 2020.

\bibitem[Sankar et~al.(2019)Sankar, Subramanian, Pal, Chandar, and
  Bengio]{sankar2019neural}
C.~Sankar, S.~Subramanian, C.~Pal, S.~Chandar, and Y.~Bengio.
\newblock Do neural dialog systems use the conversation history effectively? an
  empirical study.
\newblock \emph{arXiv preprint arXiv:1906.01603}, 2019.

\bibitem[Schulman et~al.(2017)Schulman, Chen, and
  Abbeel]{schulman2017equivalence}
J.~Schulman, X.~Chen, and P.~Abbeel.
\newblock Equivalence between policy gradients and soft q-learning.
\newblock \emph{arXiv preprint arXiv:1704.06440}, 2017.

\bibitem[Serban et~al.(2016)Serban, Sordoni, Bengio, Courville, and
  Pineau]{serban2016building}
I.~Serban, A.~Sordoni, Y.~Bengio, A.~Courville, and J.~Pineau.
\newblock Building end-to-end dialogue systems using generative hierarchical
  neural network models.
\newblock In \emph{Proceedings of the AAAI Conference on Artificial
  Intelligence}, volume~30, 2016.

\bibitem[Serban et~al.(2017{\natexlab{a}})Serban, Sankar, Germain, Zhang, Lin,
  Subramanian, Kim, Pieper, Chandar, Ke, et~al.]{serban2017deep}
I.~Serban, C.~Sankar, M.~Germain, S.~Zhang, Z.~Lin, S.~Subramanian, T.~Kim,
  M.~Pieper, S.~Chandar, N.~Ke, et~al.
\newblock A deep reinforcement learning chatbot.
\newblock \emph{arXiv preprint arXiv:1709.02349}, 2017{\natexlab{a}}.

\bibitem[Serban et~al.(2017{\natexlab{b}})Serban, Sordoni, Lowe, Charlin,
  Pineau, Courville, and Bengio]{serban2017hierarchical}
I.~Serban, A.~Sordoni, R.~Lowe, L.~Charlin, J.~Pineau, A.~Courville, and
  Y.~Bengio.
\newblock A hierarchical latent variable encoder-decoder model for generating
  dialogues.
\newblock In \emph{Proceedings of the AAAI Conference on Artificial
  Intelligence}, volume~31, 2017{\natexlab{b}}.

\bibitem[Shah et~al.(2018)Shah, Hakkani-Tur, Liu, and
  T{\"u}r]{shah2018bootstrapping}
P.~Shah, D.~Hakkani-Tur, B.~Liu, and G.~T{\"u}r.
\newblock Bootstrapping a neural conversational agent with dialogue self-play,
  crowdsourcing and on-line reinforcement learning.
\newblock In \emph{Proceedings of the 2018 Conference of the North American
  Chapter of the Association for Computational Linguistics: Human Language
  Technologies, Volume 3 (Industry Papers)}, pages 41--51, 2018.

\bibitem[Shi and Yu(2018)]{shi2018sentiment}
W.~Shi and Z.~Yu.
\newblock Sentiment adaptive end-to-end dialog systems.
\newblock \emph{arXiv preprint arXiv:1804.10731}, 2018.

\bibitem[Shin et~al.(2020)Shin, Xu, Madotto, and Fung]{shin2020generating}
J.~Shin, P.~Xu, A.~Madotto, and P.~Fung.
\newblock Generating empathetic responses by looking ahead the user’s
  sentiment.
\newblock In \emph{ICASSP 2020-2020 IEEE International Conference on Acoustics,
  Speech and Signal Processing (ICASSP)}, pages 7989--7993. IEEE, 2020.

\bibitem[Shu et~al.(2020)Shu, Nguyen, Chow, Pham, Than, Ghavamzadeh, Ermon, and
  Bui]{shu2020predictive}
R.~Shu, T.~Nguyen, Y.~Chow, T.~Pham, K.~Than, M.~Ghavamzadeh, S.~Ermon, and
  H.~Bui.
\newblock Predictive coding for locally-linear control.
\newblock In \emph{International Conference on Machine Learning}, pages
  8862--8871. PMLR, 2020.

\bibitem[Singh et~al.(2002)Singh, Litman, Kearns, and
  Walker]{singh2002optimizing}
S.~Singh, D.~Litman, M.~Kearns, and M.~Walker.
\newblock Optimizing dialogue management with reinforcement learning:
  Experiments with the njfun system.
\newblock \emph{Journal of Artificial Intelligence Research}, 16:\penalty0
  105--133, 2002.

\bibitem[Subramani et~al.(2019)Subramani, Bowman, and Cho]{subramani2019can}
N.~Subramani, S.~Bowman, and K.~Cho.
\newblock Can unconditional language models recover arbitrary sentences?
\newblock \emph{Advances in Neural Information Processing Systems}, 32, 2019.

\bibitem[Sutskever et~al.(2014)Sutskever, Vinyals, and
  Le]{sutskever2014sequence}
I.~Sutskever, O.~Vinyals, and Q.~Le.
\newblock Sequence to sequence learning with neural networks.
\newblock \emph{Advances in neural information processing systems}, 27, 2014.

\bibitem[Sutton et~al.(1999)Sutton, McAllester, Singh, and
  Mansour]{sutton1999policy}
R.~Sutton, D.~McAllester, S.~Singh, and Y.~Mansour.
\newblock Policy gradient methods for reinforcement learning with function
  approximation.
\newblock \emph{Advances in neural information processing systems}, 12, 1999.

\bibitem[Vaswani et~al.(2017)Vaswani, Shazeer, Parmar, Uszkoreit, Jones, Gomez,
  Kaiser, and Polosukhin]{vaswani2017attention}
A.~Vaswani, N.~Shazeer, N.~Parmar, J.~Uszkoreit, L.~Jones, A.~Gomez, L.~Kaiser,
  and I.~Polosukhin.
\newblock Attention is all you need.
\newblock \emph{Advances in neural information processing systems}, 30, 2017.

\bibitem[Verma et~al.(2022)Verma, Fu, Yang, and Levine]{verma2022chai}
S.~Verma, J.~Fu, M.~Yang, and S.~Levine.
\newblock Chai: A chatbot ai for task-oriented dialogue with offline
  reinforcement learning.
\newblock \emph{arXiv preprint arXiv:2204.08426}, 2022.

\bibitem[Walker(2000)]{walker2000application}
M.~Walker.
\newblock An application of reinforcement learning to dialogue strategy
  selection in a spoken dialogue system for email.
\newblock \emph{Journal of Artificial Intelligence Research}, 12:\penalty0
  387--416, 2000.

\bibitem[Wei et~al.(2018)Wei, Le, Dai, and Li]{wei2018airdialogue}
W.~Wei, Q.~Le, A.~Dai, and J.~Li.
\newblock Airdialogue: An environment for goal-oriented dialogue research.
\newblock In \emph{Proceedings of the 2018 Conference on Empirical Methods in
  Natural Language Processing}, pages 3844--3854, 2018.

\bibitem[Williams and Young(2007)]{williams2007partially}
J.~Williams and S.~Young.
\newblock Partially observable markov decision processes for spoken dialog
  systems.
\newblock \emph{Computer Speech \& Language}, 21\penalty0 (2):\penalty0
  393--422, 2007.

\bibitem[Wolf et~al.(2019)Wolf, Debut, Sanh, Chaumond, Delangue, Moi, Cistac,
  Rault, Louf, Funtowicz, et~al.]{wolf2019huggingface}
T.~Wolf, L.~Debut, V.~Sanh, J.~Chaumond, C.~Delangue, A.~Moi, P.~Cistac,
  T.~Rault, R.~Louf, M.~Funtowicz, et~al.
\newblock Huggingface's transformers: State-of-the-art natural language
  processing.
\newblock \emph{arXiv preprint arXiv:1910.03771}, 2019.

\bibitem[Yang and Nachum(2021)]{yang2021representation}
M.~Yang and O.~Nachum.
\newblock Representation matters: Offline pretraining for sequential decision
  making.
\newblock In \emph{International Conference on Machine Learning}, pages
  11784--11794. PMLR, 2021.

\bibitem[Young et~al.(2010)Young, Ga{\v{s}}i{\'c}, Keizer, Mairesse,
  Schatzmann, Thomson, and Yu]{young2010hidden}
S.~Young, M.~Ga{\v{s}}i{\'c}, S.~Keizer, F.~Mairesse, J.~Schatzmann,
  B.~Thomson, and K.~Yu.
\newblock The hidden information state model: A practical framework for
  pomdp-based spoken dialogue management.
\newblock \emph{Computer Speech \& Language}, 24\penalty0 (2):\penalty0
  150--174, 2010.

\bibitem[Zhang et~al.(2019)Zhang, Sun, Galley, Chen, Brockett, Gao, Gao, Liu,
  and Dolan]{zhang2019dialogpt}
Y.~Zhang, S.~Sun, M.~Galley, Y.~Chen, C.~Brockett, X.~Gao, J.~Gao, J.~Liu, and
  B.~Dolan.
\newblock Dialogpt: Large-scale generative pre-training for conversational
  response generation.
\newblock \emph{arXiv preprint arXiv:1911.00536}, 2019.

\bibitem[Zhao et~al.(2019)Zhao, Xie, and Eskenazi]{zhao2019rethinking}
T.~Zhao, K.~Xie, and M.~Eskenazi.
\newblock Rethinking action spaces for reinforcement learning in end-to-end
  dialog agents with latent variable models.
\newblock \emph{arXiv preprint arXiv:1902.08858}, 2019.

\bibitem[Zhou et~al.(2020)Zhou, Gao, Li, and Shum]{zhou2020design}
L.~Zhou, J.~Gao, D.~Li, and H.~Shum.
\newblock The design and implementation of xiaoice, an empathetic social
  chatbot.
\newblock \emph{Computational Linguistics}, 46\penalty0 (1):\penalty0 53--93,
  2020.

\bibitem[Ziegler et~al.(2019)Ziegler, Stiennon, Wu, Brown, Radford, Amodei,
  Christiano, and Irving]{ziegler2019fine}
D.~Ziegler, N.~Stiennon, J.~Wu, T.~Brown, A.~Radford, D.~Amodei, P.~Christiano,
  and G.~Irving.
\newblock Fine-tuning language models from human preferences.
\newblock \emph{arXiv preprint arXiv:1909.08593}, 2019.

\end{thebibliography}

{
    \appendix
    \onecolumn 
    \newpage
    % \section{Technical Results for Section \ref{sec:cm}}\label{app:moe}
    % \input{proofs_cm}
    This section describes more details about our experimental setup to evaluate the algorithms.

\section{Experimental Details}\label{app:exp_details}

\subsection{KL-Contrastive Constraint}
Recall that the main purpose of the KL-constraint is to enforce consistency in the latent variables predicted by the semantic generator and the posterior. However, since both 
$\mathcal G_0(z'|z)$ and $\rho(z'|z,Y)$ are both models to be learned, if the data can be modeled with a single, stationary LM that does not depend on the latent space, one trivial degenerated solution (that satisfies the KL constraint) is to have both of these models output very small values, which would impede the generalizability (and diversity) of the LM embedding, especially to unseen conversations. Utilizing analogous arguments as in \cite{shu2020predictive, van2018representation} that connects KL distribution matching and mutual information maximization in representation learning, we tackle this issue by considering the \emph{KL-contrastive} constraint, i.e., let \begin{small}$F(z,Y):=\text{KL}_{z'}\big(\rho(z'|z,Y)\;||\;\mathcal G_0(z'|z)\big)$\end{small}, we replace the constraint in the optimization problem in \eqref{eq:rep_opt} with
\begin{small}
\[
\widehat{\mathbb E}_{z=\Phi(\mathbf{X}), Y\sim D}\Big[F(z,Y) + \alpha\cdot\log\frac{\exp(-F(z,Y))}{\int_{Y'} \exp(-F(z,Y))}\Big]\leq \epsilon_{\text{KL}},
\]
\end{small}

where $\alpha> 0$ is a trade-off factor in the KL-constrastive constraint. While the first part of the constraint limits $F(z,Y)$, the second part enforces $F(z,Y)$ to a much higher value in the positive samples (w.r.t. ground-truth next utterance) than in the negative samples (w.r.t. other next-utterance candidates). Therefore, this constraint can prevent the aforementioned degeneration issue and inject flexibility to  control the size of the latent representation space.

\subsection{Discretization of $\Phi$ and $\mathcal G_i$}\label{sec:discretization}
Inspired by various works which use discrete bottlenecks~\citep{razavi2019generating,hafner2020mastering,yang2021representation}, we employ discretization of $\Phi$ and $\mathcal G_0$ to encourage better generalization to out-of-distribution inputs.
Specifically, we parameterize $\Phi(\mathbf{X})=\mathrm{disc}_{16}(\widetilde{\Phi}(\mathbf{X}))$,
where the output of $\widetilde{\Phi}$ is a continuous-valued vector in $\mathbb R^d$ and the discretization operator $\mathrm{disc}_K$ works as follows:
\begin{itemize}
    \item Given vector $v\in\mathbb R^d$, split $v$ into $v_1,\dots,v_{d / K}$ vectors, each of size $K$.
    \item For each $v_i$, sample a one-hot vector $v^\prime_i$ of size $K$ based on a softmax categorical distribution with logits given by $v_i$.
    \item Concatenate the vectors $v^\prime_i$ to yield a multi-hot vector $v^\prime$ of size $d$.
    \item When computing the backward pass on $v^\prime$, use straight-through gradients; i.e., $\frac{\partial v^\prime_i}{\partial v_i} = \frac{\partial \mathrm{softmax}(v_i)}{\partial v_i}$.
\end{itemize}
We similarly discretize $z'$, the output of $\mathcal G_0$. Namely, we first parameterize $\mathcal G_i$ as a Gaussian and then discretize that to form a multinomial distribution that produces multi-hot ($d/K$-hot, $K=16$) vectors.
We employ the same discretization for the experts $\mathcal G_i$ and the prior $\rho(z'|z,Y)$.

To avoid convergence to local minima for $\Phi$, we employ an entropy regularizer on the discrete distribution it is sampled from. Namely, we use an adaptive entropy regularizer in the style of~\citep{haarnoja2018soft} with a fixed target entropy that is a hyper-parameter.

\subsection{Model parameters}
In this section, the model parameters are described for MoE-1, 2, 3 and 4. All of these models represent Mixture of Experts and are based on transformer \citep{vaswani2017attention}. Transformer is an encoder-decoder based model that uses self-attention to capture relationships between the elements of the sequence. In our implementation, we used multi-head attention with implementation similar to \url{https://www.tensorflow.org/text/tutorials/transformer#point_wise_feed_forward_network}.

MoE-1 and MoE-2 use the simple transformer architecture, while MoE-3 and MoE-4 use the same encoder architecture as BERT \citep{devlin2018bert}. MoE-1 uses much smaller latent distribution models $\{\mathcal G_i\}$ than MoE-2; MoE-3 uses the pre-trained BERT encoder \url{https://tfhub.dev/tensorflow/bert_en_uncased_L-12_H-768_A-12/4}, while MoE-4 trains that from scratch.

The transformer model parameters for the simple transformer architecture are summarized in Table ~\ref{table:simple_transformer_arch}:

\begin{table}[ht]
\centering
\begin{tabular}{|l|c|c|}
\hline
\textbf{Parameter} & \textbf{Value}\\ [0.5ex]
\hline
\hline
Number of layers & 2 \\
\hline
Embedding hidden size & 256 \\
\hline
FFN inner hidden size & 512\\
\hline
Attention heads & 8 \\
\hline
Key size & 256 \\
\hline
Value size & 256 \\
\hline
Dropout & 0.1 \\
\hline
\end{tabular}
\caption{Simple Transformer Architecture}
\label{table:simple_transformer_arch}
\end{table}

The BERT based transformer model is similar to the architecture from Table~\ref{table:simple_transformer_arch}.
The differences for a pre-trained BERT model are captured in Table~\ref{table:pretrained_bert_transformer_arch}.
\begin{table}[ht]
\centering
\begin{tabular}{|l|c|c|}
\hline
\textbf{Parameter} & \textbf{Value}\\ [0.5ex]
\hline
\hline
Embedding hidden size & 768 \\
\hline
Number of layers & 12 \\
\hline
Attention heads & 12 \\
\hline
\end{tabular}
\caption{BERT-based Transformer Architecture}
\label{table:pretrained_bert_transformer_arch}
\end{table}

The differences for a BERT model we train from scratch are captured in Table~\ref{table:bert_transformer_arch}.
\begin{table}[ht]
\centering
\begin{tabular}{|l|c|c|}
\hline
\textbf{Parameter} & \textbf{Value}\\ [0.5ex]
\hline
\hline
Embedding hidden size & 768 \\
\hline
Number of layers & 2 \\
\hline
Attention heads & 8 \\
\hline
\end{tabular}
\caption{Trainable, Smaller ``BERT''-based Transformer Architecture}
\label{table:bert_transformer_arch}
\end{table}

Latent distributions $\{\mathcal G_i\}$ are implemented as FFN that model mean and variance of the normal distribution. The MoE-1, 2, 3, 4 use different values for the hidden size of the neural network. Additionally, MoE-1 and 3 use target entropy of $0.1$, MoE-2 and 4 use target entropy of $1.0$. The common parameters for FFN are captured in Table~\ref{table:opal_common_arch} (note: FFN has a final layer without an activation).

\begin{table}[ht]
\centering
\begin{tabular}{|l|c|c|}
\hline
\textbf{$\{\mathcal G_i\}$ FFN parameter} & \textbf{Value}\\ [0.5ex]
\hline
\hline
Number of layers & 1 \\
\hline
Activation & tanh \\
\hline
\end{tabular}
\caption{$\{\mathcal G_i\}$ FFN architecture}
\label{table:opal_common_arch}
\end{table}

Finally, Table ~\ref{table:moe_params} shows individual parameters distinct for each of MoE models. 
\begin{table}[ht]
\centering
\footnotesize
\begin{tabular}{|l|c|c|c|c|}
\hline
\textbf{MoE model parameter} & \textbf{$\{\mathcal G_i\}$ FFN hidden size} & \textbf{Embedding hidden size} & \textbf{Uses BERT encoder} & \textbf{Pre-trained BERT}\\ [0.5ex]
\hline
\hline
MoE-1 & 128 & 128 & No & N/A\\ 
\hline
MoE-2 & 32 & 128 & No & N/A\\
\hline
MoE-3 & 256 & 768 & Yes & Yes\\
\hline
MoE-4 & 256 & 768 & Yes & No\\
\hline
\end{tabular}
\caption{MoE parameters}
\label{table:moe_params}
\end{table}

\subsection{Computational resources}
Training and evaluation were run on 8 GPU instances with 20GB of RAM and a NVIDIA Tesla P100 graphics card.

\subsection{Dataset}
We trained our models on Reddit Casual and Cornell Movie conversational datasets. Both datasets were downloaded from Neural Chat datasets of the MIT Media Lab \url{https://affect.media.mit.edu/neural_chat/datasets}. They contain conversational exchanges between pairs of speakers. Each batch of training data contains a subset of such conversations. The Reddit Casual is about 3 times bigger than Cornell corpus. 

\subsection{Expert Label Functions}\label{app:expert_labels}
During EXP2, we define several types of language experts whose utterance outputs can constitute to smooth bot responses when used in dialogue interactions with users. 

The first type of experts we aim to create is \emph{sentiment-based}, because a fluent expression of emotions is important for creating a sense
of understanding in human conversations.
To quantify the emotional tone of a bot utterance, we use a state-of-the-art (e.g., RoBERTa Sentiment \citep{liao2021improved}) sentiment
detector, which outputs $2$ sets of prediction probabilities -- (i) whether a bot utterance is of positive, neutral, or negative sentiment; (ii) whether the bot utterance falls into any of the $4$ more-refined emotions: \{joy, optimisim, sadness, anger\}.

Neglecting the neutral sentiment output, we define $6$ sentiment labeling functions: $\ell_{\text{pos-sent}}(Y)$, $\ell_{\text{neg-sent}}(Y)$, $\ell_{\text{joy}}(Y)$, $\ell_{\text{optimism}}(Y)$, $\ell_{\text{anger}}(Y)$, $\ell_{\text{sadness}}(Y)$, which outputs a score that depends on sentiment prediction probability of any candidate bot utterance. We can also create a \emph{sentiment-empathetic} expert whose utterances will be coherent with the user's sentiment. 

RoBERTa gives scores for each of the 6 categories listed above, which then are summed weighted by the coefficients in Table ~\ref{table:roberta_coeff} to produce the \emph{sentiment-based} score:

\begin{table}[ht]
\centering
\begin{tabular}{|l|c|c|}
\hline
\textbf{Sentiment} & \textbf{Coefficient} \\ [0.5ex]
\hline
\hline
\text{pos-sent} & 0.5 \\ 
\hline
\text{neg-sent} & -0.5 \\
\hline
\text{joy} & 0.5 \\
\hline
\text{optimism} & 1 \\
\hline
\text{sadness} & -1 \\
\hline
\text{anger} & -0.5 \\
\hline
\end{tabular}
\caption{Sentiment Based Label Coefficients}
\label{table:roberta_coeff}
\end{table}

To quantify sentiment coherence between the user and bot, the labeling function $\ell_{\text{sent-coh}}(\mathbf{X},Y)$ calculates the cosine similarity of user-bot sentiment embeddings (which corresponds to the logit vector of the first set of RoBERTa predictions) in the conversation history. Concretely, it's implemented using tf.keras.losses.cosine\_similarity (\url{https://www.tensorflow.org/api_docs/python/tf/keras/losses/cosine_similarity}).

The second type of experts of interests is \emph{engagement-based}, whose goal is to encourage user's participation in the conversation. One primary candidate is the \emph{question} expert, because having a  bot that is appropriately inquisitive demonstrates the system's attentiveness to users, and thereby also increases user's responsiveness. To characterize whether a bot utterance is linked to a question, we define the label function $\ell_{\text{question}}(Y)$ which outputs $1$ when both a
question word and a question mark are present and $0$ otherwise. Another engagement skill of interests is \emph{exploration}, i.e., the expert is able to change the tone or topic of conversations to avoid having a stale conversation. To measure the amount of exploration, in the label function $\ell_{\text{exp}}(\mathbf{X},Y)$ we utilize a sentence encoder (e.g., USE \citep{cer2018universal}) to encode both the conversation history and the bot utterance and output the negative cosine similarity between these two embeddings.

\subsection{RL Rewards}\label{app:rewards}
For EXP3, the reward for RL-based dialogue management is defined as $r(\mathbf{X},a, X_+)= \lambda_1\ell_{\text{sent}}(X_+) + \lambda_2 (\ell_{\text{sent}}(X_+)-\frac{1-\gamma}{1-\gamma^{L}}\sum_{l=0}^{L-1}\gamma^l \ell_{\text{sent}}(X_l))$, where the linear combination weights $(\lambda_1,\lambda_2)=(0.75,0.25)$ correlate with  \citet{ghandeharioun2019approximating}, and $\ell_{\text{sent}}(X)$ is the same RoBerTa-based sentiment labeler as in EXP2, which assigns a score from $[-1,1]$ that is proportional to the positive sentiment and inversely proportional to the negative sentiment prediction probabilities. Intuitively, $\lambda_1$ assigns weight to \emph{sentiment-based} score of the next user response $X_+$. $\lambda_2$ assigns weight to transition of user sentiment, which we define as the difference between \emph{sentiment-based} score of next user response and \emph{discounted sentiment-based} score of the current conversation.

%%%%%%%%%%%%%%%%%%%%%%%%%%%

    \section{Additional Results} \label{appendix:exp_results}
    \subsection{Dialogue Management Results on Cornell Movie Corpus}\label{app:cornell_results}

\begin{table*}[ht]
 \begin{minipage}{0.5\textwidth}
\centering
  {\tiny
  \begin{tabular}{|c||c|c|c|c|c|}
  \hline
  Method & Diversity  &	Dist-1 &  Dist-2 & Dist-3	 &  Perplexity	\\
  \hline
 MoE-1 &  0.073 $\pm$ 0.02 & 0.24	 & 0.59  & 	0.67 &  76.17 $\pm$ 33.04	\\ \hline
 MoE-2 & 0.12 $\pm$ 0.01  &	0.30 & 0.65  & 0.77	 &  \textbf{45.24 $\pm$ 40.77}	\\ \hline
 MoE-3 &  0.11 $\pm$ 0.01 &	0.28 & 0.66  & 0.72	 &  77.94 $\pm$ 26.55	\\ \hline
 MoE-4 & \textbf{0.16 $\pm$ 0.01} &	\textbf{0.35} & \textbf{0.77}  & \textbf{0.84}	 &  60.06 $\pm$ 23.74	\\ \hline
 Trans. & 0.060 $\pm$ 0.03 &	0.25 & 0.55   & 0.70	 &  69.86 $\pm$ 16.88	 \\ \hline
 VHRED &0.095 $\pm$ 0.03 &	0.27 & 0.53  & 0.68	 &  101.88 $\pm$	54.78 	 \\ \hline	
 \end{tabular}
  \caption{Accuracy (Perplexity) and Diversity of Language Primitive Experts Trained with Cornell.}
  \label{tab:phase_1_cornell}
  }
  \end{minipage}\,\,\,\,\,\,\,\,\,
  \begin{minipage}{0.55\textwidth}
\centering
  {\tiny
  \begin{tabular}{|c||c|c|c|c|}
  \hline
 Method &  User Tot. Sent. & User Sent. Trans.	 &  GPT-Perplexity	\\
  \hline
  MoE-2 Primitive & -0.14 $\pm$ 0.11 & 0.10$\pm$ 0.05 & 93.11 $\pm$ 50.41	\\ \hline
 MoE-2 DQN & -0.17 $\pm$ 0.13 & -0.08$\pm$ 0.05 & 90.01$\pm$ 30.76	\\ \hline
 MoE-2 CQL & \textbf{0.17$\pm$0.15} & \textbf{0.18$\pm$ 0.15} & 88.83$\pm$ 57.78	\\ \hline
 MoE-2 MBRL & 0.13$\pm$ 0.11 & 0.09 $\pm$ 0.07 & \textbf{85.59 $\pm$ 40.01}	\\ \hline
 Trans. RL & -0.35$\pm$ 0.19 & -0.15$\pm$ 0.18 & 199.75 $\pm$ 64.23\\ \hline
 Trans. KLC & -0.09 $\pm$ 0.06 & -0.13 $\pm$ 0.08 & 145.09$\pm$ 55.71\\ \hline
 Trans. VHRL & -0.07 $\pm$ 0.13 & 0.05 $\pm$ 0.07 & 160.02$\pm$ 60.30\\ \hline
 \end{tabular}
  \caption{Quality of DM Trained with Cornell Dataset.}
  \label{tab:phase_3_cornell}
  }
  \end{minipage}
\end{table*}

\begin{table*}[ht]
 \begin{minipage}{\textwidth}
\centering
  {\tiny
  \begin{tabular}{|c||c|c|c|c|c|c|c|c|c|}
  \hline
  Method & Question  &	Exploration &  Positive Sent. & Negative Sent.	& Sent. Coherence &  Joy & Optimism &	Anger & Sadness \\
  \hline
 MoE-1 & 1.19 $\pm$ 0.23 & 0.33 $\pm$ 0.25	 & 1.24 $\pm$ 0.43 & 1.30 $\pm$ 0.64	 & 0.79 $\pm$ 0.44 & 2.04 $\pm$ 0.39 & 0.02 $\pm$ 1.01 & 1.92 $\pm$ 0.99 & 3.77 $\pm$ 1.05 \\ \hline
 MoE-2 &  1.57 $\pm$ 0.27 & \textbf{0.36 $\pm$ 0.21}	 & \textbf{3.62 $\pm$ 2.01} & 1.80 $\pm$ 1.89	 & 0.68 $\pm$ 0.25 & \textbf{2.67 $\pm$ 0.97} & \textbf{2.26 $\pm$ 1.35} & \textbf{2.80 $\pm$ 1.14} & \textbf{4.32 $\pm$ 1.76}	\\ \hline
 MoE-3 & 1.14 $\pm$  0.31& 0.29 $\pm$ 0.37	 & 0.35 $\pm$ 0.31 & 	1.89 $\pm$ 0.58 & 0.75 $\pm$ 0.21 & 1.88 $\pm$ 0.77 & 0.29 $\pm$ 0.81 & 0.65 $\pm$ 0.99 & 2.99 $\pm$ 1.01	\\ \hline
 MoE-4 & \textbf{ 1.84 $\pm$ 0.33} & 0.26 $\pm$ 0.51	 &    0.17 $\pm$ 1.86 & \textbf{2.20 $\pm$ 1.64}	 &  0.80 $\pm$ 0.98 &  0.52 $\pm$ 1.91 & 0.58 $\pm$ 0.61 & 2.72 $\pm$ 0.75 & 3.77 $\pm$ 1.43	\\ \hline
 WD & 0.86 $\pm$ 1.7  & 0.21 $\pm$ 0.289	 & -0.48 $\pm$ 0.44  & 	0.41 $\pm$ 0.46 & \textbf{0.84 $\pm$ 0.19}  & 0.13 $\pm$ 1.48  & -1.11 $\pm$ 0.88 & 1.51 $\pm$ 2.02 & 1.22 $\pm$ 0.80\\ \hline
 PPLM & 0.31 $\pm$ 0.45 & 0.35 $\pm$ 0.36	 &  -0.49 $\pm$ 2.32 & 0.42 $\pm$ 2.13	 & 0.74 $\pm$ 0.24 & -1.00 $\pm$ 2.54 & -0.86 $\pm$ 2.41 & 0.69 $\pm$ 0.05 & 0.74 $\pm$ 2.27 \\ \hline\hline
 Trans. RL* & 2.20 $\pm$ 0.32 &	0.44 $\pm$ 0.26 & 3.58 $\pm$ 1.05 & 2.98 $\pm$ 1.53	 & 0.82 $\pm$ 0.30 & 3.06 $\pm$ 1.85 & 1.64 $\pm$ 1.21 & 3.22 $\pm$ 2.01 & 3.98 $\pm$ 1.88	\\ \hline	
 \end{tabular}
  \caption{Quality of Each Expert PPLM Trained on Cornell Dataset w.r.t. Its Trained Label.}
  \label{tab:phase_2_cornell}
  }
  \end{minipage}
\end{table*}

\newpage
\subsection{Sample Utterances for EXP 1}\label{app:sample_utterances_1}
\subsubsection{Reddit Casual}
\begin{enumerate}
\item \textbf{Input Conversation:} (...) i mean , i have n't changed or done anything drastically different , but things have been good .
\begin{itemize}
%\item \textbf{Ground Truth:} but my outlook on life is still meh\\
\item \textbf{MoE-1 R1:} but my outlook on life is n't meh .
\item \textbf{MoE-1 R2:} but my outlook on life is subjective .
\item \textbf{MoE-1 R3:} but my outlook on life is still meh .
\item \textbf{MoE-1 R4:} but my view on life is still meh ?
\item \textbf{MoE-1 R5:} but my outlook on life is n't meh ?
\\
\item \textbf{MoE-2 R1:}  i love the good things that i 'm starting too.
\item \textbf{MoE-2 R2:} i do love your life , because mine 's nothing .
\item \textbf{MoE-2 R3:}  no , i 'm not too busy in life
\item \textbf{MoE-2 R4:} good for you , i 'm working on it .
\item \textbf{MoE-2 R5:} but my life is boring and boring .
\\
\item \textbf{MoE-3 R1:} but my outlook on life is still sad .
\item \textbf{MoE-3 R2:} but my outlook on life is the same .
\item \textbf{MoE-3 R3:} but my outlook on life is my same !
\item \textbf{MoE-3 R4:} but my outlook on life is not good .
\item \textbf{MoE-3 R5:} but my outlook on life is my friend .
\\
\item \textbf{MoE-4 R1:} i 'm in my early 20 's , i do n't know what to do 
\item \textbf{MoE-4 R2:} i 'm so happy for you .
\item \textbf{MoE-4 R3:} for me , it 's tough to go through life with pain .
\item \textbf{MoE-4 R4:} i used to think that you 'd enjoy that much more than most people .
\item \textbf{MoE-4 R5:} can you describe what happened ?
\\
\end{itemize}
\item \textbf{Input Conversation:} (...) but my outlook on life is still meh
\begin{itemize}
%\item \textbf{Ground Truth:} no need for drastically difference , just take it slow .
\item \textbf{MoE-1 R1:} no need for drastically change, just take it slow for a break ?
\item \textbf{MoE-1 R2:} no need to slow down, just take it a long day ?
\item \textbf{MoE-1 R3:} no need to slow down , just take a step too long .
\item \textbf{MoE-1 R4:} no need a blanket , just take it off .
\item \textbf{MoE-1 R5:} no need for a breath , just take it a bit of time
\\
\item \textbf{MoE-2 R1:} being a good person is n't a bad thing 
\item \textbf{MoE-2 R2:} i 'm in a similar situation for it .
\item \textbf{MoE-2 R3:} take a break , try to find something else .
\item \textbf{MoE-2 R4:} oh yeah , life 's pretty good for me .
\item \textbf{MoE-2 R5:} like you 're in a bad place for a few years ?
\\
\item \textbf{MoE-3 R1:} no need for drastic changes , just make it super busy
\item \textbf{MoE-3 R2:} no need for drastic changes , just do it extra time
\item \textbf{MoE-3 R3:} no need for unnecessary stress , just move the next direction
\item \textbf{MoE-3 R4:} no change for unnecessary things , just move on .
\item \textbf{MoE-3 R5:} no need for unnecessary consumption , just take with the move
\\
\item \textbf{MoE-4 R1:} it seems like a great time to just go outside and play a few songs , i 'm always glad to hear that you have a good friend
\item \textbf{MoE-4 R2:} i know life 's not perfect , it can be difficult at times !
\item \textbf{MoE-4 R3:} end complaining all the time and just consider your life a failure .
\item \textbf{MoE-4 R4:} you 'd be surprised at how nice life can go !
\item \textbf{MoE-4 R5:} i mean  some sort of hobby can definitely help out .
\\
\end{itemize}
\item \textbf{Input Conversation:} (...) no need for drastically difference , just take it slow .
\begin{itemize}
%\item \textbf{Ground Truth:} and it 's nice to hear that things have been good for you , my view of life is meh too , it changes depending on time and mood\\
\item \textbf{MoE-1 R1:} and it 's nice to hear someone else 's good , but it 's all about my life .
\item \textbf{MoE-1 R2:} and it 's nice to hear , when you have new seasons of american world is just a nature in canada .
\item \textbf{MoE-1 R3:} and it 's nice to hear why you 've been doing it , but my friend is still in it ...
\item \textbf{MoE-1 R4:} and it 's very good to hear why you 've been good with my wife , a lot of things that 's happening in the same way .
\item \textbf{MoE-1 R5:} and it 's nice to hear things have been good , my view of life changes with time and mood
\\
\item \textbf{MoE-2 R1:} i find it hard to get out of this world , but there is also a good way to find a positive way of life .,
\item \textbf{MoE-2 R2:} you seem to be in the same situation , but it is a good way to do it for a lot of things in life
\item \textbf{MoE-2 R3:} my life is n't bad , i 'm in a very long time and i m pretty happy about it .
\item \textbf{MoE-2 R4:} i just wanted to see what 's happening in my life and i m not sure what to do .
\item \textbf{MoE-2 R5:} at this point i 'm pretty sure the most part of my life is pretty normal , but it s a lot of positive .
\\
\item \textbf{MoE-3 R1:} and it 's nice to hear that things have been good for you that your view life is a lot like mine .
\item \textbf{MoE-3 R2:} and it 's nice to hear that things have been good for you so my perspective of life is too much and so i think that 's a pass
\item \textbf{MoE-3 R3:} and it 's nice to hear that things have been good for you in my view of life , like a lot of people are trying to have
\item \textbf{MoE-3 R4:} and it 's nice to hear that things have been good for you , my view my life is awful that , but i feel like this is a lot better
\item \textbf{MoE-3 R5:} and it 's nice to hear that things have been good for you in the future of your life , so maybe you 're looking forward to it
\\
\item \textbf{MoE-4 R1:} i 've been trying to learn things about myself .
\item \textbf{MoE-4 R2:} that 's like i would be at this stage of life .
\item \textbf{MoE-4 R3:} you will need to enjoy life for yourself .
\item \textbf{MoE-4 R4:} for me it 's like i do n't have the time .
\item \textbf{MoE-4 R5:} but like you said , you 've been trying to get your life together and learn to be a normal person to love yourself .
\\
\end{itemize}
\item \textbf{Input Conversation:} (...)  and it 's nice to hear that things have been good for you d my view of life is meh too , it changes depending on time and mood
\begin{itemize}
%\item \textbf{Ground Truth:} and i do n't know why i say this , this is not even relevant , but remember rule 32 enjoy the little things\\
\item \textbf{MoE-1 R1:} and i do n't know why i say this , this is not a relevant post , but i can really say the same things .
\item \textbf{MoE-1 R2:} and i do n't know why i say this , this is not even relevant , but any things will be good .
\item \textbf{MoE-1 R3:} and i do n't know why i say this , this is not even a relevant topic , i do n't remember the same things i said
\item \textbf{MoE-1 R4:} and i do n't know why i say this , this is n't even relevant , but definitely i enjoy the same things
\item \textbf{MoE-1 R5:} and i do n't know why i say this , this is not a relevant topic , but i think there 's actually a bunch of things to do
\\
\item \textbf{MoE-2 R1:} i think i 'm just trying to be a good person , but i m not sure what it is like in my life .
\item \textbf{MoE-2 R2:} for me it 's my life i feel like it s the same way of being around the world and i m pretty sure it s not a lot
\item \textbf{MoE-2 R3:} to be honest , i 'm not sure what the most beautiful things are like in the world 
\item \textbf{MoE-2 R4:} i feel like you 're going to be a good person , but i 'm not a very happy about the world of things you are around
\item \textbf{MoE-2 R5:}  i just want to be in a way to see my life and i m not really sure how much of other things i like 
\\
\item \textbf{MoE-3 R1:} and i do n't know why i say this , this is not really relevant , but i do wish you the best .
\item \textbf{MoE-3 R2:} and i do n't know why i say this , this is not really relevant , but i can count the experience 
\item \textbf{MoE-3 R3:} and i do n't know why i say this , this is not even relevant , but now i feel like it 's the kind of boring ...
\item \textbf{MoE-3 R4:} and i do n't know why i say this , this is not even relevant , but i like watching ultra match in my room .
\item \textbf{MoE-3 R5:} and i do n't know why i say this , this is not even relevant , but i do feel like you 're welcome here .
\\
\item \textbf{MoE-4 R1:} i hope things get better for you soon 
\item \textbf{MoE-4 R2:}  i think it 's fine , i enjoy life and could be a little happier every day 
\item \textbf{MoE-4 R3:} i like it d be a good time to think about the things that you don t need to do .
\item \textbf{MoE-4 R4:} but let 's forget on those things that make you miserable and trying to focus on the basics .
\item \textbf{MoE-4 R5:} i know it 's not perfect , but enjoy the things you have .
\\
\end{itemize}
\item \textbf{Input Conversation:} (...) and i do n't know why i say this , this is not even relevant , but remember rule 32 enjoy the little things
\begin{itemize}
%\item \textbf{Ground Truth:} yeahhh , i 'm definitely trying to enjoy the little things like walking down the street with a really nice song on , or randomly buying myself ice cream\\
\item \textbf{MoE-1 R1:} i 'm trying to enjoy the little things like walking down the street , or playing random games .
\item \textbf{MoE-1 R2:} i 'm just trying to enjoy the little things like sports , like small talk , or random ones , or even give up 
\item \textbf{MoE-1 R3:} i 'm just trying to enjoy the little things like walking down the street with the same interactions or drinking
\item \textbf{MoE-1 R4:} i 'm trying to enjoy the little things like going down the beach with the same thing , or a nice spot 
\item \textbf{MoE-1 R5:} i 'm trying to enjoy the little things like walking down the beach , watching the weird movies , listen to random music 
\\
\item \textbf{MoE-2 R1:} this is a good time , i 'm just looking for a place to go and i 've always wanted to go to the store
\item \textbf{MoE-2 R2:} my favorite thing is that i 'm just looking for a good thing to do , but i 'm always looking forward to the people i 've been
\item \textbf{MoE-2 R3:} i might be in the same boat , but i 'm a little scared to know what is a really good thing to look at the moment
\item \textbf{MoE-2 R4:} i 've been trying to find out what i want to do , but i 'm not really sure 
\item \textbf{MoE-2 R5:}  i know it 's a good thing , but i 'm always in the same boat as i 've been in a long run
\\
\item \textbf{MoE-3 R1:} yeah haha , i 'm definitely trying to enjoy the little things like walking down the street , or just have a hot cup of tea 
\item \textbf{MoE-3 R2:} yeah haha , i 'm definitely trying to enjoy the little things like walking down the street or seeing what is in my neighborhood
\item \textbf{MoE-3 R3:} yes , i m definitely trying to enjoy the little things like walking down stairs , but i can t figure out what to see
\item \textbf{MoE-3 R4:} yeah dude , i m definitely trying to enjoy the little things like walking down the street with a huge fan of these kids
\item \textbf{MoE-3 R5:} yeah man , i 'm definitely trying to enjoy the little things like from walking up the street or to my little kid 's house
\\
\item \textbf{MoE-4 R1:} that 's what i 'm really trying to do and enjoy the moment .
\item \textbf{MoE-4 R2:} i 've always been like that , i 've been really chill at least trying to enjoy my life 
\item \textbf{MoE-4 R3:} i 'm trying to get a good life and have a good time , i 'm not sure what to do in my free time
\item \textbf{MoE-4 R4:} and i 've definitely learned a lot of things to do
\item \textbf{MoE-4 R5:} i d say that it has been given me both the time and pleasure, but i m still an introvert with few hobbies
\end{itemize}
\end{enumerate}

\subsubsection{Cornell Movie Corpus}
\begin{enumerate}
\item \textbf{Input Conversation:} (...) excuse me
\begin{itemize}
%\item \textbf{Ground Truth:} yes ?
% \\
\item \textbf{MoE-1 R1:} i do n't know .
\item \textbf{MoE-1 R2:} i do n't know .
\item \textbf{MoE-1 R3:} yes .
\item \textbf{MoE-1 R4:} what do you mean ?
\item \textbf{MoE-1 R5:} i 'm sorry .
\\
\item \textbf{MoE-2 R1:} i 'll give you a few minutes .
\item \textbf{MoE-2 R2:} i do n't want to talk about it .
\item \textbf{MoE-2 R3:} may i come in ?
\item \textbf{MoE-2 R4:} where is this , mr. kessler ?
\item \textbf{MoE-2 R5:} what do you think ?
\\
\item \textbf{MoE-3 R1:} what 's it ? are you alright ?
\item \textbf{MoE-3 R2:} you want me to read ?
\item \textbf{MoE-3 R3:} i do n't know .
\item \textbf{MoE-3 R4:} i 'll just have to kill you .
\item \textbf{MoE-3 R5:} you ca n't say it 's over .
\\
\item \textbf{MoE-4 R1:} well if you hang up , you 'll die on it .
\item \textbf{MoE-4 R2:} superman ? oh , was he here ?
\item \textbf{MoE-4 R3:} well , you know , that 's a lie , i do n't have your money .
\item \textbf{MoE-4 R4:} are you the killer ben ?
\item \textbf{MoE-4 R5:} there 's nothing to be wrong with that .
\\
\end{itemize}
\item \textbf{Input Conversation:} (...)  yes ?
\begin{itemize}
%\item \textbf{Ground Truth:} can i ask my friend , can he come in ? just for a few minutes ?\\
\item \textbf{MoE-1 R1:}  what do you mean ?
\item \textbf{MoE-1 R2:} i 'm not sure .
\item \textbf{MoE-1 R3:} what 's the matter ?
\item \textbf{MoE-1 R4:} i do n't know .
\item \textbf{MoE-1 R5:} no .
\\
\item \textbf{MoE-2 R1:} what did you say ?
\item \textbf{MoE-2 R2:} he says that you 're not a good man .
\item \textbf{MoE-2 R3:} i 'll tell you what i 'll do . i 'll be right back .
\item \textbf{MoE-2 R4:} do you want to go to the police ?
\item \textbf{MoE-2 R5:} are you still in love with him ?
\\
\item \textbf{MoE-3 R1:} sam , i 'm getting married .
\item \textbf{MoE-3 R2:} johnny . it was announced this morning . he 's going to play the lead in the new film .
\item \textbf{MoE-3 R3:} i 'm sorry , mister mcmurphy , i count only eight .
\item \textbf{MoE-3 R4:} i said i was n't looking for him . will you excuse my miss ... ?
\item \textbf{MoE-3 R5:} pay ! in and i say !
\\
\item \textbf{MoE-4 R1:} i would like to marry you . but i have already promised myself ... to god .
\item \textbf{MoE-4 R2:} do you know where he 's living ?
\item \textbf{MoE-4 R3:} i know you 're soft .
\item \textbf{MoE-4 R4:} it 's susan 's cooking , claire . why uncle james lets them stay ?
\item \textbf{MoE-4 R5:} i do n't want to be any trouble .
\\
\end{itemize}
\item \textbf{Input Conversation:} (...) can i ask my friend , can he come in ? just for a few minutes ?
\begin{itemize}
%\item \textbf{Ground Truth:} your friend ?\\
\item \textbf{MoE-1 R1:} i 'm not sure .
\item \textbf{MoE-1 R2:} i do n't know
\item \textbf{MoE-1 R3:} he 's not a man , is n't he ?
\item \textbf{MoE-1 R4:} no, i 'm sorry .
\item \textbf{MoE-1 R5:} what do you mean ?
\\
\item \textbf{MoE-2 R1:} all right . what are you going to do ?
\item \textbf{MoE-2 R2:} no , i 've never seen him before .
\item \textbf{MoE-2 R3:} i want to be with him .
\item \textbf{MoE-2 R4:} this is a very interesting man .
\item \textbf{MoE-2 R5:} he 's not the greatest . you do n't know .
\\
\item \textbf{MoE-3 R1:} oh , come on , i 'm sick and tired of your life .
\item \textbf{MoE-3 R2:} no . i still do n't have any reason to do that .
\item \textbf{MoE-3 R3:} my god , are you kidding , i 'm a little tired .
\item \textbf{MoE-3 R4:} i do n't understand . where is he ?
\item \textbf{MoE-3 R5:} no . i still do n't have any reason let you be together
\\
\item \textbf{MoE-4 R1:} oh , of course !
\item \textbf{MoE-4 R2:} because he was scary , right ?
\item \textbf{MoE-4 R3:} i 'm sure you 're a good friend .
\item \textbf{MoE-4 R4:} can i get you a glass of champagne ?
\item \textbf{MoE-4 R5:} so i like to make a point . you got a fucking friend that does n't work
\\
\end{itemize}
\item \textbf{Input Conversation:} (...) your friend ?
\begin{itemize}
%\item \textbf{Ground Truth:} he 's going back to the front this evening . i ca n't see him otherwise . \\
\item \textbf{MoE-1 R1:} i do n't know
\item \textbf{MoE-1 R2:} you 're not going to tell him ?
\item \textbf{MoE-1 R3:} well , i 'm not sure .
\item \textbf{MoE-1 R4:} i do n't know .
\item \textbf{MoE-1 R5:} i 'm sorry
\\
\item \textbf{MoE-2 R1:} i like him . he 's a very good man .
\item \textbf{MoE-2 R2:} maybe i should go home and talk to him .
\item \textbf{MoE-2 R3:} i 'm not going to let him go .
\item \textbf{MoE-2 R4:} oh , please .
\item \textbf{MoE-2 R5:} if you do n't mind , you 'll have to do it , will you ?
\\
\item \textbf{MoE-3 R1:} what 's the idea of working with you ? what 's the matter ? !
\item \textbf{MoE-3 R2:} i 'll look into her . tell 'em .
\item \textbf{MoE-3 R3:} that 's something i have n't done the before .
\item \textbf{MoE-3 R4:} so , our lawyers are here .
\item \textbf{MoE-3 R5:} we 've got some plans here .
\\
\item \textbf{MoE-4 R1:} my friend , you must be really glad to help . if you do n't mind .
\item \textbf{MoE-4 R2:}  is n't there something you can do . how can you help at night ?
\item \textbf{MoE-4 R3:} how do you do , mrs. ryan ? i 'd heard you were beautiful , and so you are .
\item \textbf{MoE-4 R4:} the winning side would have paid you much better .
\item \textbf{MoE-4 R5:} now , here 's a babe ! I give you a blue ox , you got ta help me !
\\
\end{itemize}
\item \textbf{Input Conversation:} (...) he 's going back to the front this evening . i ca n't see him otherwise .
\begin{itemize}
%\item \textbf{Ground Truth:}  just go off . i 'll be quite all right .\\
\item \textbf{MoE-1 R1:} i 'm not going to be here .
\item \textbf{MoE-1 R2:} what do you mean ?
\item \textbf{MoE-1 R3:} i 'm not sure . it 's not your fault .
\item \textbf{MoE-1 R4:} i do n't know .
\item \textbf{MoE-1 R5:} you 're not going to tell him ?
\\
\item \textbf{MoE-2 R1:} i think i 'll tell you what i 'm gon na do .
\item \textbf{MoE-2 R2:} he 's on the way , right ?
\item \textbf{MoE-2 R3:} that 's not true .
\item \textbf{MoE-2 R4:} not too late . i 'm not going to be here .
\item \textbf{MoE-2 R5:} what are you talking about ? he 's a very sweet man .
\\
\item \textbf{MoE-3 R1:} what ? i think it 's wonderful .
\item \textbf{MoE-3 R2:} sure , i guess !
\item \textbf{MoE-3 R3:} oh , for christ 's sake .
\item \textbf{MoE-3 R4:} they 'll be all right , you 're my guest .
\item \textbf{MoE-3 R5:} leon , you are not going to negotiate !
\\
\item \textbf{MoE-4 R1:} ok. he asks about you all the time . i want to tell you .
\item \textbf{MoE-4 R2:} the way i feel , believe me , it 's hard to part and see a guy 's picture .
\item \textbf{MoE-4 R3:} do n't worry . i 'll take care of it .
\item \textbf{MoE-4 R4:} certainly it 's important .
\item \textbf{MoE-4 R5:} yeah , and what a real hero .
\end{itemize}
\end{enumerate}

\subsection{Sample Utterances for EXP 2: Question Intent}\label{app:sample_utterances_2_1}
\subsubsection{Reddit Casual}
\begin{enumerate}
\item \textbf{Sample Conversation:} (...) i love how genuine your comments are and you only comment if it s how you really feel and not for karma like when you wish people a merry christmas
\begin{itemize}
%\item \textbf{Ground Truth:} people like you really have a special place in my heart . \\
\item \textbf{MoE-1 R1:} what was your favorite guys and why did you have a boyfriend ?
\item \textbf{MoE-1 R2:} i really love your username , and i am so happy for you .
\item \textbf{MoE-1 R3:} what was your favorite guys and why did you have a nice day ?
\\
\item \textbf{MoE-2 R1:} how did you find someone that you like ?
\item \textbf{MoE-2 R2:} just wondering if you re interested in reddit ?
\item \textbf{MoE-2 R3:} how did you feel about it ?
\\
\item \textbf{MoE-3 R1:}  people like you really have a special place in my heart .
\item \textbf{MoE-3 R2:} people like you really have a special place in my heart to me
\item \textbf{MoE-3 R3:} people like you really have a special place in my heart?
\\
\item \textbf{MoE-2 R1:} how are you liking it ?
\item \textbf{MoE-2 R2:} do you know i just love being together with you ?
\item \textbf{MoE-2 R3:} how are you my friend ? ?
\\
\end{itemize}
\item \textbf{Sample Conversation:} (...) people like you really have a special place in my heart .
\begin{itemize}
%\item \textbf{Ground Truth:} hope your christmas was pretty good , my friend assuming you celebrate .\\
\item \textbf{MoE-1 R1:} that s nice !
\item \textbf{MoE-1 R2:} happy cake day !
\item \textbf{MoE-1 R3:} that sounds nice , you have a good time to stay alone ?
\\
\item \textbf{MoE-2 R1:} just wondering if you re happy with me ?
\item \textbf{MoE-2 R2:} how did you feel about it ?
\item \textbf{MoE-2 R3:} how did you get this ?
\\
\item \textbf{MoE-3 R1:} hope your christmas was a nice one to celebrate with you
\item \textbf{MoE-3 R2:} hope your christmas was pretty good , my friend being celebrate with .
\item \textbf{MoE-3 R3:} hope your christmas was nice with my friend who celebrate with you
\\
\item \textbf{MoE-4 R1:} omg why !
\item \textbf{MoE-4 R2:} how far into your heart ?
\item \textbf{MoE-4 R3:} how are you liking this happiness as well ?
\\
\end{itemize}
\item \textbf{Sample Conversation:} (...) hope your christmas was pretty good , my friend assuming you celebrate .
\begin{itemize}
%\item \textbf{Ground Truth:} thank you!\\
\item \textbf{MoE-1 R1:} thank you !
\item \textbf{MoE-1 R2:} thanks !,
\item \textbf{MoE-1 R3:} thank you !
\\
\item \textbf{MoE-2 R1:} how did you guys meet this year ?
\item \textbf{MoE-2 R2:} how was your day going ?
\item \textbf{MoE-2 R3:} how did you celebrate this year ?
\\
\item \textbf{MoE-3 R1:} haha thank you !
\item \textbf{MoE-3 R2:} thank you !
\item \textbf{MoE-3 R3:} thank you .
\\
\item \textbf{MoE-4 R1:} how 's your day going ?
\item \textbf{MoE-4 R2:} how are you liking it so far ?
\item \textbf{MoE-4 R3:} i have no idea what it 's about ?
\\
\end{itemize}
\end{enumerate}

\subsubsection{Cornell Movie Corpus}
\begin{enumerate}
\item \textbf{Sample Conversation:} that 's some face they left hanging on you . you look like ...
\begin{itemize}
%\item \textbf{Ground Truth:} could you take it just a little easy , mr. udall ?\\
\item \textbf{MoE-1 R1:} what ? what are you talking about ?
\item \textbf{MoE-1 R2:} what ? what are you talking about ?
\item \textbf{MoE-1 R3:} what ?
\\
\item \textbf{MoE-2 R1:} what ? what 's wrong with it ?
\item \textbf{MoE-2 R2:} what ? you think they 're going to kill me ?
\item \textbf{MoE-2 R3:} what is it ?
\\
\item \textbf{MoE-3 R1:}  well . what 's the difference ?
\item \textbf{MoE-3 R2:} what ? what 's the deal ?
\item \textbf{MoE-3 R3:} what ? what ? what 's wrong with what happened ?
\\
\item \textbf{MoE-4 R1:} what ? there 's no pain in my mind ?
\item \textbf{MoE-4 R2:} what ? why do you think i want to know ?
\item \textbf{MoE-4 R3:} what can i do ? there 's no way i can change .
\\
\end{itemize}
\item \textbf{Sample Conversation:} what hotel ?
\begin{itemize}
%\item \textbf{Ground Truth:} the holiday hotel .\\
\item \textbf{MoE-1 R1:} what the fuck are you doing here ?
\item \textbf{MoE-1 R2:} what ?
\item \textbf{MoE-1 R3:} what are you talking about ?,
\\
\item \textbf{MoE-2 R1:} what 's the matter ?
\item \textbf{MoE-2 R2:} what about it ?
\item \textbf{MoE-2 R3:} who 're you traveling with?
\\
\item \textbf{MoE-3 R1:} whatever . i 'm going to work here .
\item \textbf{MoE-3 R2:} tom . how are you ?
\item \textbf{MoE-3 R3:} what ? what about the hotel ?
\\
\item \textbf{MoE-4 R1:} do you think i want the plaza ?
\item \textbf{MoE-4 R2:} what do you think ? where 's my mother ?
\item \textbf{MoE-4 R3:} what hotel i should go?
\\
\end{itemize}
\item \textbf{Sample Conversation:} the holiday hotel .
\begin{itemize}
%\item \textbf{Ground Truth:} what room number ?\\
\item \textbf{MoE-1 R1:} what are you talking about ?
\item \textbf{MoE-1 R2:} what ?
\item \textbf{MoE-1 R3:} what are you doing here ?
\\
\item \textbf{MoE-2 R1:} what is your travel plan?
\item \textbf{MoE-2 R2:} how can i get there?
\item \textbf{MoE-2 R3:} why do you want to be there?
\\
\item \textbf{MoE-3 R1:}  what ? are you walking in there ?,
\item \textbf{MoE-3 R2:} absolutely . what 's up with the hotel ?
\item \textbf{MoE-3 R3:} what ? bring 'em in
\\
\item \textbf{MoE-4 R1:} what are you ? a cop ?
\item \textbf{MoE-4 R2:} what about that ? i work here .
\item \textbf{MoE-4 R3:} what ? ... oh you know he 's there ?,
\\
\end{itemize}
\end{enumerate}

\subsection{Sample Utterances for EXP 2: Positive Sentiment Intent}\label{app:sample_utterances_2_2}
\subsubsection{Reddit Casual}
\begin{enumerate}
\item \textbf{Sample Conversation:} (...) it s definitely not the college experience .
\begin{itemize}
%\item \textbf{Ground Truth:} feels more like high school except way easier to mess up and nobody really talks to anyone lmao .\\
\item \textbf{MoE-1 R1:} i feel like i m more of a high school and talking to people .
\item \textbf{MoE-1 R2:} i feel like i 'm not a teacher , and i talk to , i talk to people about it .
\item \textbf{MoE-1 R3:} i feel like a lot of people are just as nervous as i talk to , and i really really like talking about .
\\
\item \textbf{MoE-2 R1:} i m so proud of you for being able to keep up with it !
\item \textbf{MoE-2 R2:} i hope you re having a great day !
\item \textbf{MoE-2 R3:} i have a very good time on social media !
\\
\item \textbf{MoE-3 R1:}  seems like middle school to end up way more stable than everyone and don t know who does this
\item \textbf{MoE-3 R2:} seems like real good compared to everyone after work and nobody is around me even the age of 35
\item \textbf{MoE-3 R3:}  feels like a real awkward going to say though and nobody cares about each other with me
\\
\item \textbf{MoE-4 R1:} if you want to get into college , i 'm pretty good at it !
\item \textbf{MoE-4 R2:} thank you for taking the time to look at it , i 'm trying to get a few more opportunities for you now for your semester
\item \textbf{MoE-4 R3:} good for you , i like college and have more people in my life , i know better than high school i ve been
\\
\end{itemize}
\item \textbf{Sample Conversation:} (...) feels more like high school except way easier to mess up and nobody really talks to anyone lmao .
\begin{itemize}
%\item \textbf{Ground Truth:} i m looking forward to finally being out , not only because i think i ll be a lot happier at a full uni , but also bc i hate home\\
\item \textbf{MoE-1 R1:} i m on reddit , and i m going to be pretty far from my first year s college , but i d probably have to go to college
\item \textbf{MoE-1 R2:} i m not really good at all , but i m a college student , i d be working on a job , i d be able to go,
\item \textbf{MoE-1 R3:} i m pretty sure it s a lot easier to get a job , but i guess it s a lot of work in my field 
\\
\item \textbf{MoE-2 R1:} i m going to be a senior and i m happy to meet people !
\item \textbf{MoE-2 R2:} i hope you re having a great time !
\item \textbf{MoE-2 R3:} i get to see my new friends and family !
\\
\item \textbf{MoE-3 R1:}  i 'm looking forward to being able to say , but not because i am a great life all for everyone to see each other once in a while
\item \textbf{MoE-3 R2:} i 'm looking forward to going out , not because i feel so much better knowing i will still have a great day , and i 'm happy to see
\item \textbf{MoE-3 R3:} i 'm looking forward to going home , not only because i think i will be a lot happier with my family and looking at the very young people .
\\
\item \textbf{MoE-4 R1:} for sure , i ve been in college for the past few years that i ve had plenty of people in my life .
\item \textbf{MoE-4 R2:} thank you for your input , i appreciate it !
\item \textbf{MoE-4 R3:} you ll be proud of yourself and i hope you are able to enjoy those people on your first year
\\
\end{itemize}
\item \textbf{Sample Conversation:} (...) i m looking forward to finally being out , not only because i think i ll be a lot happier at a full uni , but also bc i hate home
\begin{itemize}
%\item \textbf{Ground Truth:} alright well this and the top comment really make me feel a lot better because to be honest it is just hard .\\
\item \textbf{MoE-1 R1:} i finally got the job that i m going to be a very good dad , and it just sucks .
\item \textbf{MoE-1 R2:} i finally got a new job that i feel like is not a good thing .
\item \textbf{MoE-1 R3:} i finally got the hang of my parents , and i feel very lucky to be very far as possible .
\\
\item \textbf{MoE-2 R1:} now i m a college student and i m very proud of myself !
\item \textbf{MoE-2 R2:} i ll try to meet new people !
\item \textbf{MoE-2 R3:} i feel like i m a little too proud of myself !
\\
\item \textbf{MoE-3 R1:}  yeah well and the whole thing will really help me feel a bit shitty right now is n't easy
\item \textbf{MoE-3 R2:} yeah well this is the start and really make me feel good a bit better than what i am .
\item \textbf{MoE-3 R3:} yeah well and the goal is really help me feel good since everyone is mostly different in their lives .
\\
\item \textbf{MoE-4 R1:} good advice , i really like it when i have to get a college degree , but i 've also got a decent amount of friends who have not
\item \textbf{MoE-4 R2:} thank you for the advice , i really like to keep on going for college and i 'm sure my friends will go to college
\item \textbf{MoE-4 R3:} for sure , i want to get back on track to how much i used to be doing , but i don t really like the idea of studying
\\
\end{itemize}
\end{enumerate}

\subsubsection{Cornell Movie Corpus}
\begin{enumerate}
\item \textbf{Sample Conversation:} (...) no i fucking do n't ! you ca n't make a cup of tea , edward .
\begin{itemize}
%\item \textbf{Ground Truth:} why not ? the whole of the british empire was built on cups of tea .\\
\item \textbf{MoE-1 R1:} what ?
\item \textbf{MoE-1 R2:} no .
\item \textbf{MoE-1 R3:} no , i 'm not .
\\
\item \textbf{MoE-2 R1:} yeah , i got it .
\item \textbf{MoE-2 R2:} oh , i 'm so glad you say it .
\item \textbf{MoE-2 R3:} yes , it 's a very nice idea .
\\
\item \textbf{MoE-3 R1:}  yes , sir . what 's wrong ?
\item \textbf{MoE-3 R2:} yes . yes , sir .
\item \textbf{MoE-3 R3:} yes . who is this ?
\\
\item \textbf{MoE-4 R1:} oh , yes . i 'm going to miss you .,
\item \textbf{MoE-4 R2:} yes . i 'm sorry , i did n't know how to do it .
\item \textbf{MoE-4 R3:} yes , yes . i 'll tell you something about it .
\\
\end{itemize}
\item \textbf{Sample Conversation:} (...) why not ? the whole of the british empire was built on cups of tea .
\begin{itemize}
%\item \textbf{Ground Truth:} and look what happened to that .\\
\item \textbf{MoE-1 R1:} what ?
\item \textbf{MoE-1 R2:} no !
\item \textbf{MoE-1 R3:} no , no .
\\
\item \textbf{MoE-2 R1:} good . i 'm glad you say that .
\item \textbf{MoE-2 R2:} yeah , i got it .
\item \textbf{MoE-2 R3:} yes , you are right about that .
\\
\item \textbf{MoE-3 R1:}   yes . yes , sir .
\item \textbf{MoE-3 R2:} yes , of course .
\item \textbf{MoE-3 R3:} yes . yes . that 's right .
\\
\item \textbf{MoE-4 R1:} yes , i 'll tell you what it is .
\item \textbf{MoE-4 R2:} yes , yes . i 'll tell you something else !
\item \textbf{MoE-4 R3:} yes , i 'm with you .
\\
\end{itemize}
\item \textbf{Sample Conversation:} (...) if you 'll be good enough to remove those newspapers i 'll have a seat .
\begin{itemize}
%\item \textbf{Ground Truth:} okay ! okay ! keep your shirt on , young feller .\\
\item \textbf{MoE-1 R1:} no , i 'm not .
\item \textbf{MoE-1 R2:} what ?
\item \textbf{MoE-1 R3:} no !
\\
\item \textbf{MoE-2 R1:} yes , i think it 's a good idea .
\item \textbf{MoE-2 R2:} yes , yes . i 'll be in a minute .
\item \textbf{MoE-2 R3:} good . yes , you 're not going to have to do it .
\\
\item \textbf{MoE-3 R1:}  yes . yes . yes . yes .
\item \textbf{MoE-3 R2:} what 's the matter with you ?
\item \textbf{MoE-3 R3:} yes , ma'am , i do .
\\
\item \textbf{MoE-4 R1:} oh , yes ! i 'll do it .
\item \textbf{MoE-4 R2:} yes , yes . i 'm sure you must be a very sick man .
\item \textbf{MoE-4 R3:} yes , yes . i 'm sure you do ..
\\
\end{itemize}
\end{enumerate}

\subsection{Sample Utterances for EXP 2: Negative Sentiment Intent}\label{app:sample_utterances_2_3}
\subsubsection{Reddit Casual}
\begin{enumerate}
\item \textbf{Sample Conversation:} (...) it s definitely not the college experience .
\begin{itemize}
%\item \textbf{Ground Truth:} feels more like high school except way easier to mess up and nobody really talks to anyone lmao .\\
\item \textbf{MoE-1 R1:} i m not much of a sports person , i find that sort of an emotional connection to others .
\item \textbf{MoE-1 R2:} i 'm more of a high school student and i m wondering if anyone else to talk to about this .
\item \textbf{MoE-1 R3:} i 'm not a social person , but i 'm just really curious how to talk to other people .
\\
\item \textbf{MoE-2 R1:} i don t know if i should be going to school right now .
\item \textbf{MoE-2 R2:} i don t know how to do it , i m nervous .
\item \textbf{MoE-2 R3:} i don t think i m going to be able to do all the things i ve been doing .
\\
\item \textbf{MoE-3 R1:}  feels like shit not less to lie than excuse myself to just miss class all the time .
\item \textbf{MoE-3 R2:} feels like decent becoming middle school system and not close to him anymore anyways because nobody ever mentioned .
\item \textbf{MoE-3 R3:} feels like middle school only harder to stay up and nobody really talks about anything except nowadays .
\\
\item \textbf{MoE-4 R1:} because i hate making out of my freshman year i was happy about it 
\item \textbf{MoE-4 R2:} but i hate to admit it , so like this was the first time i even went to college .
\item \textbf{MoE-4 R3:} and people are terrible at figuring out what to do , you could n't pay extra for this year !
\\
\end{itemize}
\item \textbf{Sample Conversation:} (...) feels more like high school except way easier to mess up and nobody really talks to anyone lmao .
\begin{itemize}
%\item \textbf{Ground Truth:} i m looking forward to finally being out , not only because i think i ll be a lot happier at a full uni , but also bc i hate home\\
\item \textbf{MoE-1 R1:} i m not really good at all , but i guess i would have to go to a university for a good job .
\item \textbf{MoE-1 R2:} so far that i 'm not happy to have my own experience , but i m also going to have a job that i d like to have .
\item \textbf{MoE-1 R3:} i m on the same boat as you , but i m also a college student for a few years .
\\
\item \textbf{MoE-2 R1:} i m not going to try and start doing anything for the next semester .
\item \textbf{MoE-2 R2:} i don t know if it s the same .
\item \textbf{MoE-2 R3:} i don t know if it s going to be the same since i m still in college .
\\
\item \textbf{MoE-3 R1:}  i m looking forward to going back to being alone , but not because i feel like i m a better fit in with the wait til stressful one
\item \textbf{MoE-3 R2:} i 'm looking forward to going back home , not only because i feel that there 's a lot of time to do something productive after i get there
\item \textbf{MoE-3 R3:} i 'm looking forward to being at school , not only because i think i 'll be a bit happier now and will have to pay for 4 months
\\
\item \textbf{MoE-4 R1:} and i hate it , i get bored !
\item \textbf{MoE-4 R2:} i just switched because i was a senior student and i know absolutely hate that i had to do
\item \textbf{MoE-4 R3:} and i hate talking about it , i 'll get left out 
\\
\end{itemize}
\item \textbf{Sample Conversation:} (...) i m looking forward to finally being out , not only because i think i ll be a lot happier at a full uni , but also bc i hate home
\begin{itemize}
%\item \textbf{Ground Truth:} alright well this and the top comment really make me feel a lot better because to be honest it is just hard .\\
\item \textbf{MoE-1 R1:} so i finally got a new job that i am not a very good person .
\item \textbf{MoE-1 R2:} i 'm not going to be able to find a way to feel better about it , but i just feel better .
\item \textbf{MoE-1 R3:} i m not the most comfortable person who wants to be so bad about it or something that i should be able to be able to do .
\\
\item \textbf{MoE-2 R1:} i feel like i m not going to be honest .
\item \textbf{MoE-2 R2:} i feel like i m not going to be in the same position .
\item \textbf{MoE-2 R3:} i don t know if it s all about going to school .
\\
\item \textbf{MoE-3 R1:}  yeah well all and the person does really make me feel better for everyone to be so serious about it .
\item \textbf{MoE-3 R2:} yeah well this and the beat all does seem feel a bit better because everyone is basically just that it is in their ass .
\item \textbf{MoE-3 R3:} yeah well and the post really not me feel better a lot better since it is not being considered so big .
\\
\item \textbf{MoE-4 R1:} and i hate waking up , i feel like it all the time .
\item \textbf{MoE-4 R2:} and i hate to admit it , i do n't know what 's going on .
\item \textbf{MoE-4 R3:} and i know there are people who can keep up with their homework but i do n't know how to handle them !
\\
\end{itemize}
\end{enumerate}

\subsubsection{Cornell Movie Corpus}
\begin{enumerate}
\item \textbf{Sample Conversation:} (...) no i fucking do n't ! you ca n't make a cup of tea , edward .
\begin{itemize}
%\item \textbf{Ground Truth:} why not ? the whole of the british empire was built on cups of tea .\\
\item \textbf{MoE-1 R1:} i 'm sorry .
\item \textbf{MoE-1 R2:} i 'm not sure .
\item \textbf{MoE-1 R3:} i do n't know , i do n't know .
\\
\item \textbf{MoE-2 R1:} what do i have to do ?
\item \textbf{MoE-2 R2:} why do n't you just leave it be ?
\item \textbf{MoE-2 R3:} you do n't know what you 're talking about .
\\
\item \textbf{MoE-3 R1:}  i 'm not interested in what people used to do .
\item \textbf{MoE-3 R2:} i 'm afraid that 's not up to me 
\item \textbf{MoE-3 R3:} yes . what 's the problem with it ?
\\
\item \textbf{MoE-4 R1:} i do n't want to hear your problems .
\item \textbf{MoE-4 R2:} i do n't care . i have a lot of money in my life .
\item \textbf{MoE-4 R3:} i do n't want a goddamn cup of tea . i want to go home . i 'll have to be here , okay ?
\\
\end{itemize}
\item \textbf{Sample Conversation:} (...) why not ? the whole of the british empire was built on cups of tea .
\begin{itemize}
%\item \textbf{Ground Truth:} and look what happened to that .\\
\item \textbf{MoE-1 R1:} i do n't know .
\item \textbf{MoE-1 R2:} i 'm not sure . i 'm not going to be here .
\item \textbf{MoE-1 R3:} i do n't know .
\\
\item \textbf{MoE-2 R1:} who do you think you are ?
\item \textbf{MoE-2 R2:} who cares ? i do n't know what to say .
\item \textbf{MoE-2 R3:} do n't be a fool . do n't you know what i 'm talking about .
\\
\item \textbf{MoE-3 R1:}   yes . what 's more like this ?
\item \textbf{MoE-3 R2:} i do n't believe that either .
\item \textbf{MoE-3 R3:} that is correct . and how do we decide ?
\\
\item \textbf{MoE-4 R1:} i do n't believe in it !
\item \textbf{MoE-4 R2:} i do n't want to hear you anymore . you hear me ?
\item \textbf{MoE-4 R3:} i do n't want to hear it . i know you 're very clever .
\\
\end{itemize}
\item \textbf{Sample Conversation:} (...) if you 'll be good enough to remove those newspapers i 'll have a seat .
\begin{itemize}
%\item \textbf{Ground Truth:} okay ! okay ! keep your shirt on , young feller .\\
\item \textbf{MoE-1 R1:} i do n't know .
\item \textbf{MoE-1 R2:} i 'm not sure .
\item \textbf{MoE-1 R3:} i do n't know .
\\
\item \textbf{MoE-2 R1:} i do n't know what you 're talking about .
\item \textbf{MoE-2 R2:} who are you ? can you get out of here ?
\item \textbf{MoE-2 R3:} do n't be a fool . what do you want me to do ?
\\
\item \textbf{MoE-3 R1:}   you 're welcome to one of these , whatever .
\item \textbf{MoE-3 R2:} what difference does it make ?
\item \textbf{MoE-3 R3:} i 'm afraid not . have to ask you a question .
\\
\item \textbf{MoE-4 R1:} i do n't want to hear it . i do n't want to hear any more today .
\item \textbf{MoE-4 R2:}  look at this ! why are you laughing ?
\item \textbf{MoE-4 R3:} i do n't care ! i do n't know what 's good with you !
\\
\end{itemize}
\end{enumerate}

\subsection{Sample Utterances for EXP 2: Joy Emotion Intent}\label{app:sample_utterances_2_4}
\subsubsection{Reddit Casual}
\begin{enumerate}
\item \textbf{Sample Conversation:} (...) besides every anime intro and outro song ever at least from the series i like , i 've been getting a kick out of j rock in general .
\begin{itemize}
%\item \textbf{Ground Truth:} i got deep into idol music this year so i can relate a bit .\\
\item \textbf{MoE-1 R1:} i got a song a few months ago and i can relate .
\item \textbf{MoE-1 R2:} i got a song from a song that i can recently finish .
\item \textbf{MoE-1 R3:}  i got a song from a song that has been a really rough time lately .
\\
\item \textbf{MoE-2 R1:} what s your favourite anime to listen to ?
\item \textbf{MoE-2 R2:} i just enjoy the new anime and i 'm a big fan of it .
\item \textbf{MoE-2 R3:} i 'm always a fan of the anime and i 've been listening to it all the lately 
\\
\item \textbf{MoE-3 R1:}  i got very into french music this year which i can relate every year 
\item \textbf{MoE-3 R2:} i got into love pop music this year , so i can relate with very well .
\item \textbf{MoE-3 R3:} i got into love rap recently this year so i can relate with you haha
\\
\item \textbf{MoE-4 R1:} haha i 'm into anime !
\item \textbf{MoE-4 R2:} haha , anime manga is considered my first language .
\item \textbf{MoE-4 R3:} sounds like my favourite anime and manga japan , and they probably are my main thing .
\\
\end{itemize}
\item \textbf{Sample Conversation:} (...) i got deep into idol music this year so i can relate a bit .
\begin{itemize}
%\item \textbf{Ground Truth:} how deep in anime trash are you ?\\
\item \textbf{MoE-1 R1:} i 've been thinking about the same as a lot of new bands i listen to .
\item \textbf{MoE-1 R2:}  i just started a new anime where i feel like i 'm into the same genre .
\item \textbf{MoE-1 R3:} i just like the new design and see it on youtube .
\\
\item \textbf{MoE-2 R1:} <unk> is my favorite song
\item \textbf{MoE-2 R2:} i love the song !
\item \textbf{MoE-2 R3:} what do you like about the music ?
\\
\item \textbf{MoE-3 R1:}  i have really watched deep dark <unk> recently .
\item \textbf{MoE-3 R2:} yeah i have heard this song called so long 
\item \textbf{MoE-3 R3:} i know just your song i played video games sometimes .
\\
\item \textbf{MoE-4 R1:} i 'm also loving the music so far .
\item \textbf{MoE-4 R2:} haha , i love anime manga from the 90 's .
\item \textbf{MoE-4 R3:} haha yes but for me it 's all anime .
\\
\end{itemize}
\item \textbf{Sample Conversation:} (...)  some people have the same username across multiple websites such as steam or xbox .
\begin{itemize}
%\item \textbf{Ground Truth:} you might even be able to get a hold of them in different places with only their username .\\
\item \textbf{MoE-1 R1:} people might even get mad at them , because they are on their own .
\item \textbf{MoE-1 R2:} i might be the opposite .
\item \textbf{MoE-1 R3:}  i never thought about being a dick .
\\
\item \textbf{MoE-2 R1:} i guess there 's a reason for people to use their username too .
\item \textbf{MoE-2 R2:} i just use reddit , and the new design is pretty good .
\item \textbf{MoE-2 R3:} i see a lot of people using this as well as the same goes .
\\
\item \textbf{MoE-3 R1:}  you might also be able to get a hold of their different places with only their username .
\item \textbf{MoE-3 R2:} you might even be able to get a hold of them in different places with only their username .
\item \textbf{MoE-3 R3:} you might also be able to get a hold of my different places with the internet access .
\\
\item \textbf{MoE-4 R1:} i 'm also loving the music so far .
\item \textbf{MoE-4 R2:} haha well i guess you are talking about the internet .
\item \textbf{MoE-4 R3:} haha good karma happens in a lot of us , but for me i 'm most on reddit .
\\
\end{itemize}
\end{enumerate}

\subsubsection{Cornell Movie Corpus}
\begin{enumerate}
\item \textbf{Sample Conversation:} (...) i guess i cut pretty loose in my day too .
\begin{itemize}
%\item \textbf{Ground Truth:} really , frank ? when was that ?\\
\item \textbf{MoE-1 R1:} oh , i 'm fine .
\item \textbf{MoE-1 R2:} oh , yeah .
\item \textbf{MoE-1 R3:} oh , yeah .
\\
\item \textbf{MoE-2 R1:}  yes , darling . i 'm looking for you .
\item \textbf{MoE-2 R2:} nice . good .
\item \textbf{MoE-2 R3:} okay , we 'll see you .
\\
\item \textbf{MoE-3 R1:} oh yeah . that 's nice .
\item \textbf{MoE-3 R2:} oh , it 's probably a good idea .
\item \textbf{MoE-3 R3:} oh , it 's worth it .
\\
\item \textbf{MoE-4 R1:} it 's so beautiful . there 's a lot of money in there .
\item \textbf{MoE-4 R2:} i 'm so sure you 'll find out how that d go.
\item \textbf{MoE-4 R3:} i said it 's good call . 
\\
\end{itemize}
\item \textbf{Sample Conversation:} (...) i could get ben to tap into the baxter 's main power to generate enough voltage
\begin{itemize}
%\item \textbf{Ground Truth:} reed . shh . just be quiet . and look up .\\
\item \textbf{MoE-1 R1:} yes , sir .
\item \textbf{MoE-1 R2:} oh , yes , sir .
\item \textbf{MoE-1 R3:} oh , yes , sir .
\\
\item \textbf{MoE-2 R1:} and i 'm sure it 's a good idea 
\item \textbf{MoE-2 R2:} yes , ma'am . i 'm going to work .
\item \textbf{MoE-2 R3:} oh , i 'm going to have a nice time .
\\
\item \textbf{MoE-3 R1:}  hey . that 's the problem .
\item \textbf{MoE-3 R2:} oh , that 's a possibility .
\item \textbf{MoE-3 R3:} well . hope you enjoy it .
\\
\item \textbf{MoE-4 R1:} i do n't know . but i 'm sure it 's a very good idea .
\item \textbf{MoE-4 R2:} i 'm glad you got your sense of humor back . think you 'll keep it ?
\item \textbf{MoE-4 R3:} yes , that 's exactly what we got to do .
\\
\end{itemize}
\item \textbf{Sample Conversation:} (...) or that he ca n't be bothered to write us a proper letter ? a dissertation on apples , we do n't need !
\begin{itemize}
%\item \textbf{Ground Truth:} he probably does n't make much money picking apples he must have had to pay to send them , too .\\
\item \textbf{MoE-1 R1:} no , sir .
\item \textbf{MoE-1 R2:} yes , sir .
\item \textbf{MoE-1 R3:} yes , sir .
\\
\item \textbf{MoE-2 R1:} yes , sir . they 're telling me all this .
\item \textbf{MoE-2 R2:}  he 's a bad guy , huh ?
\item \textbf{MoE-2 R3:} oh , that 's a good idea .
\\
\item \textbf{MoE-3 R1:}  oh . well , that 's good .
\item \textbf{MoE-3 R2:} oh , that sounds great .
\item \textbf{MoE-3 R3:} yes , he 's that sweetheart .
\\
\item \textbf{MoE-4 R1:} he says he does n't feel good too . i thought you might know .
\item \textbf{MoE-4 R2:} yes . he was all right .
\item \textbf{MoE-4 R3:} jeez , right . he 's like a ghost !
\\
\end{itemize}
\end{enumerate}

\subsection{Sample Utterances for EXP 2: Optimism Emotion Intent}\label{app:sample_utterances_2_5}
\subsubsection{Reddit Casual}
\begin{enumerate}
\item \textbf{Sample Conversation:} (...) besides every anime intro and outro song ever at least from the series i like , i 've been getting a kick out of j rock in general .
\begin{itemize}
%\item \textbf{Ground Truth:} i got deep into idol music this year so i can relate a bit .\\
\item \textbf{MoE-1 R1:} got a few more of my favorite songs !
\item \textbf{MoE-1 R2:} have you ever got a song that you can listen to ?
\item \textbf{MoE-1 R3:}  i ve been listening to a lot of songs lately and i can relate .
\\
\item \textbf{MoE-2 R1:} good lord , definitely send  forever shit to the artist ...
\item \textbf{MoE-2 R2:} good for you , the fantastic video !
\item \textbf{MoE-2 R3:} good one , but the next one is , better !
\\
\item \textbf{MoE-3 R1:}  i got deep into idol music this year so i can relate a bit .
\item \textbf{MoE-3 R2:} i got big into pop music today so i can relate your success !
\item \textbf{MoE-3 R3:} i got deep into music recently so i will love my new year !
\\
\item \textbf{MoE-4 R1:} good for you , it s a good anime intro haha
\item \textbf{MoE-4 R2:} might have some japanese jazz fusion and art
\item \textbf{MoE-4 R3:} good one , the anime and manga really fit me
\\
\end{itemize}
\item \textbf{Sample Conversation:} (...) i got deep into idol music this year so i can relate a bit .
\begin{itemize}
%\item \textbf{Ground Truth:} how deep in anime trash are you ?\\
\item \textbf{MoE-1 R1:} what do you like about it ?
\item \textbf{MoE-1 R2:} what do you like to see ?
\item \textbf{MoE-1 R3:}  i 'm not sure what to watch ?
\\
\item \textbf{MoE-2 R1:} good one , the next anime 's been awesome !,
\item \textbf{MoE-2 R2:} good for you , my friend !
\item \textbf{MoE-2 R3:} the best way to start the new year !
\\
\item \textbf{MoE-3 R1:}  is it your favourite song ever ?
\item \textbf{MoE-3 R2:} does it match your taste ?
\item \textbf{MoE-3 R3:} is it really your favorite movie ?
\\
\item \textbf{MoE-4 R1:} good one , especially for the current age
\item \textbf{MoE-4 R2:}  thanks for your good song it gets more popular
\item \textbf{MoE-4 R3:} good lord , the music is popular !
\\
\end{itemize}
\item \textbf{Sample Conversation:} (...)  some people have the same username across multiple websites such as steam or xbox .
\begin{itemize}
%\item \textbf{Ground Truth:} you might even be able to get a hold of them in different places with only their username .\\
\item \textbf{MoE-1 R1:} it 's not even a joke , but i think it is a scam .
\item \textbf{MoE-1 R2:} why people would use it as a joke ?
\item \textbf{MoE-1 R3:} how do you know how to deal with it ?
\\
\item \textbf{MoE-2 R1:} what you can see is a new generation of the future.
\item \textbf{MoE-2 R2:} be aware of what their user names are and comments are made to post 
\item \textbf{MoE-2 R3:} it depends on the mods and the subreddit , i see them as a possible community .
\\
\item \textbf{MoE-3 R1:}  you might even be able to get a hold of their different places with different people 
\item \textbf{MoE-3 R2:} you might even be able to get a hold of their different places with different people than your own view .
\item \textbf{MoE-3 R3:} you may even be able to get a hold of your friend in different places with their username ?
\\
\item \textbf{MoE-4 R1:} alright , so , especially reddit  . might have to check out the real difference , especially with older ones !
\item \textbf{MoE-4 R2:} good on you for noticing when you mentioned that though not to be a problem !
\item \textbf{MoE-4 R3:}  good for you , despite only being the mom and dad for me , i believe there are some old people now , still hoping to do that
\\
\end{itemize}
\end{enumerate}

\subsubsection{Cornell Movie Corpus}
\begin{enumerate}
\item \textbf{Sample Conversation:} (...) i guess i cut pretty loose in my day too .
\begin{itemize}
%\item \textbf{Ground Truth:}  really , frank ? when was that ?\\
\item \textbf{MoE-1 R1:} oh , i 'm fine .
\item \textbf{MoE-1 R2:} oh , yeah .
\item \textbf{MoE-1 R3:} oh , i 'm sorry .
\\
\item \textbf{MoE-2 R1:} okay , take it easy .
\item \textbf{MoE-2 R2:} all right . everything 's been pretty .
\item \textbf{MoE-2 R3:} go ahead . the last time you 're right .
\\
\item \textbf{MoE-3 R1:}  i 'm talking about ...
\item \textbf{MoE-3 R2:} let 's see if i ever known .
\item \textbf{MoE-3 R3:} you learned from your american friends ?
\\
\item \textbf{MoE-4 R1:} i'm sure it can be fine . it 's just you and me , brad .
\item \textbf{MoE-4 R2:} it 's so cool . really .
\item \textbf{MoE-4 R3:} you 're right . i 'm sure it will work . i could feel you in your position .
\\
\end{itemize}
\item \textbf{Sample Conversation:} (...) i could get ben to tap into the baxter 's main power to generate enough voltage
\begin{itemize}
%\item \textbf{Ground Truth:} reed . shh . just be quiet . and look up .\\
\item \textbf{MoE-1 R1:} yes , sir . but we 're not going to do the same thing .
\item \textbf{MoE-1 R2:} no , sir .
\item \textbf{MoE-1 R3:} yes , sir .
\\
\item \textbf{MoE-2 R1:} all right . it 's the way . i 'm going to get it .
\item \textbf{MoE-2 R2:} all right . you 'll do it .
\item \textbf{MoE-2 R3:} go ahead . i 'll be all right .
\\
\item \textbf{MoE-3 R1:} it is ! i 'm stopping !
\item \textbf{MoE-3 R2:} that bastard ! i 'll gladly dispose of him in the name of the order , son of malkovich .
\item \textbf{MoE-3 R3:} i think they should stop and start looking right here i think they should stop and start looking right here .
\\
\item \textbf{MoE-4 R1:} i 'm looking at it . i 'm going to get it right .
\item \textbf{MoE-4 R2:} there 's something else , major . i 'm starting to think what you said has a ton of beauty in it
\item \textbf{MoE-4 R3:} see , that 's what we 're here for .
\\
\end{itemize}
\item \textbf{Sample Conversation:} (...) or that he ca n't be bothered to write us a proper letter ? a dissertation on apples , we do n't need !
\begin{itemize}
%\item \textbf{Ground Truth:} he probably does n't make much money picking apples he must have had to pay to send them , too .\\
\item \textbf{MoE-1 R1:} yes , sir . is that what you mean ?
\item \textbf{MoE-1 R2:}  yes , sir .
\item \textbf{MoE-1 R3:} oh , yes , sir .
\\
\item \textbf{MoE-2 R1:} all right . we 'll be able to figure out what 's wrong with him .
\item \textbf{MoE-2 R2:} okay , take it easy .
\item \textbf{MoE-2 R3:} all right . we 'll be able to figure out what 's going on .
\\
\item \textbf{MoE-3 R1:} so ? the man got silver , china , rugs
\item \textbf{MoE-3 R2:} he 's out to the hospital ...
\item \textbf{MoE-3 R3:} i see , i was just thinking ... maybe it 's time to settle down .
\\
\item \textbf{MoE-4 R1:} you do n't understand , i 've been in love with him !
\item \textbf{MoE-4 R2:} he 'll be back . i 'll wait for his car .
\item \textbf{MoE-4 R3:} i 'll think of something . i do n't think he 's a bad guy .
\\
\end{itemize}
\end{enumerate}

\subsection{Sample Utterances for EXP 2: Anger Emotion Intent}\label{app:sample_utterances_2_6}
\subsubsection{Reddit Casual}
\begin{enumerate}
\item \textbf{Sample Conversation:} (...) besides every anime intro and outro song ever at least from the series i like , i 've been getting a kick out of j rock in general .
\begin{itemize}
%\item \textbf{Ground Truth:} i got deep into idol music this year so i can relate a bit .\\
\item \textbf{MoE-1 R1:} i got ta have this song , but i can relate to this .,
\item \textbf{MoE-1 R2:} i got this song a few months ago , and i can relate to a bit .
\item \textbf{MoE-1 R3:} i got my thoughts into this , but i can relate to this .
\\
\item \textbf{MoE-2 R1:} why are you into anime ?,
\item \textbf{MoE-2 R2:} i lose interest in it .
\item \textbf{MoE-2 R3:} i 'm not an expert when it comes to it .
\\
\item \textbf{MoE-3 R1:}  i got deep into idol music this year so i can relate a bit .
\item \textbf{MoE-3 R2:} i got deep into this punk shows recently so i can relate with it .
\item \textbf{MoE-3 R3:} i got ta learn rock this pop punk so i can relate with it .
\\
\item \textbf{MoE-4 R1:} keeping my speed for reading the anime shit .
\item \textbf{MoE-4 R2:}  why are you damn talking english only ?
\item \textbf{MoE-4 R3:} and this is why i am not such a rich dude .
\\
\end{itemize}
\item \textbf{Sample Conversation:} (...) i got deep into idol music this year so i can relate a bit .
\begin{itemize}
%\item \textbf{Ground Truth:} how deep in anime trash are you ?\\
\item \textbf{MoE-1 R1:} oh wow , that 's a good one .
\item \textbf{MoE-1 R2:} i like it , i 'm not a fan of the music genre , but i 've had a lot of music .
\item \textbf{MoE-1 R3:} oh wow .
\\
\item \textbf{MoE-2 R1:} why are you considering the big ex ?
\item \textbf{MoE-2 R2:} i don 't watch anime in general at the moment .
\item \textbf{MoE-2 R3:} why are you considering anime good ?
\\
\item \textbf{MoE-3 R1:}  i can totally see that this was all about the songs i did n't talk to about .
\item \textbf{MoE-3 R2:} i would n't miss out on that video .
\item \textbf{MoE-3 R3:} i don t miss that so much .
\\
\item \textbf{MoE-4 R1:} fuck yeah , i was there for like shit .
\item \textbf{MoE-4 R2:} why are you considering doing a lot of anime ?
\item \textbf{MoE-4 R3:} why do you think anime is bad ?
\\
\end{itemize}
\item \textbf{Sample Conversation:} (...)  some people have the same username across multiple websites such as steam or xbox .
\begin{itemize}
%\item \textbf{Ground Truth:} you might even be able to get a hold of them in different places with only their username .\\
\item \textbf{MoE-1 R1:} i think it all comes down to if they 're serious .
\item \textbf{MoE-1 R2:} i think it 's just how they are in their mid 20s how they are , they are used to their own lives .
\item \textbf{MoE-1 R3:} it might seem like a lot of people in their late 20s , they have a large problem with their accent .
\\
\item \textbf{MoE-2 R1:} why do people assume they have their own identity ?
\item \textbf{MoE-2 R2:} it 's just a series of spam jokes and whatnot
\item \textbf{MoE-2 R3:} i think people are just dumb to the bottom line of their head .
\\
\item \textbf{MoE-3 R1:} you might also be able to find a hold of them with different people than the previous ones .
\item \textbf{MoE-3 R2:} you might also be able to get a hold of them in different places with the username .
\item \textbf{MoE-3 R3:} you might even be able to get a hold of them in different places with their username .
\\
\item \textbf{MoE-4 R1:} i dont know who downvoted you , but let people tell you the truth of your karma . why are you asking on reddit ?
\item \textbf{MoE-4 R2:} i dont necessarily agree with them but just became the first and its just became a good one .
\item \textbf{MoE-4 R3:} you 're around and just do n't use the internet for any specific reason , i think there are thousands of people watching .
\\
\end{itemize}
\end{enumerate}

\subsubsection{Cornell Movie Corpus}
\begin{enumerate}
\item \textbf{Sample Conversation:} (...) i guess i cut pretty loose in my day too .
\begin{itemize}
%\item \textbf{Ground Truth:}  really , frank ? when was that ?\\
\item \textbf{MoE-1 R1:} you do n't have to .
\item \textbf{MoE-1 R2:} no .
\item \textbf{MoE-1 R3:} you do n't have to have a good look .
\\
\item \textbf{MoE-2 R1:} i think you must be crazy . you ca n't tell me what you 're doing here !
\item \textbf{MoE-2 R2:} you 're not the one trying to get away . who owns the gun ?
\item \textbf{MoE-2 R3:} you do n't pull this cowboy shit , you 're gon na get him out ?
\\
\item \textbf{MoE-3 R1:}  what do you know . i saw you ?
\item \textbf{MoE-3 R2:}  what the hell is that 
\item \textbf{MoE-3 R3:} oh , you do n't know . i 'm sorry .
\\
\item \textbf{MoE-4 R1:} what are you talking about ? you love your job .
\item \textbf{MoE-4 R2:} you 're like an undercover guy !
\item \textbf{MoE-4 R3:} you do n't get paid till we close tonight .
\\
\end{itemize}
\item \textbf{Sample Conversation:} (...) i could get ben to tap into the baxter 's main power to generate enough voltage
\begin{itemize}
%\item \textbf{Ground Truth:} reed . shh . just be quiet . and look up .\\
\item \textbf{MoE-1 R1:} you do n't have to do this .
\item \textbf{MoE-1 R2:} you do n't know what you 're talking about .
\item \textbf{MoE-1 R3:} you do n't know what you 're talking about .
\\
\item \textbf{MoE-2 R1:} you got to be kidding !
\item \textbf{MoE-2 R2:} we 're not going to do this .
\item \textbf{MoE-2 R3:} we want to frighten you away . so that you 'll move out
\\
\item \textbf{MoE-3 R1:}  do n't you dare , i 'll be all that ...
\item \textbf{MoE-3 R2:} so you 're telling the truth ?
\item \textbf{MoE-3 R3:} you think i 'm right 
\\
\item \textbf{MoE-4 R1:} you 're driving me nuts !
\item \textbf{MoE-4 R2:} you 're trying to kill me , are n't you ?
\item \textbf{MoE-4 R3:} you do n't get it , pham . the idea was we were ripping them off .
\\
\end{itemize}
\item \textbf{Sample Conversation:} (...) or that he ca n't be bothered to write us a proper letter ? a dissertation on apples , we do n't need !
\begin{itemize}
%\item \textbf{Ground Truth:} he probably does n't make much money picking apples he must have had to pay to send them , too .\\
\item \textbf{MoE-1 R1:} do n't worry about it 
\item \textbf{MoE-1 R2:} do n't you think he 's in the dark . 
\item \textbf{MoE-1 R3:}  do n't worry about it .
\\
\item \textbf{MoE-2 R1:} i ca n't believe you did n't see that coming . you do n't want to lose this !
\item \textbf{MoE-2 R2:} what ? you 're not going to kill him .
\item \textbf{MoE-2 R3:} now what ? we 're going to have to do something
\\
\item \textbf{MoE-3 R1:}  because you do n't know what you are doing .
\item \textbf{MoE-3 R2:} oh , you do n't know . i guess it 's just not true .
\item \textbf{MoE-3 R3:} and what the hell is that ?
\\
\item \textbf{MoE-4 R1:} he 'll kill you ! motherfucker ! !
\item \textbf{MoE-4 R2:} see , i told you he was gon na kill us all .
\item \textbf{MoE-4 R3:} and you do n't care what it says ?
\\
\end{itemize}
\end{enumerate}

\subsection{Sample Utterances for EXP 2: Sadness Emotion Intent}\label{app:sample_utterances_2_7}
\subsubsection{Reddit Casual}
\begin{enumerate}
\item \textbf{Sample Conversation:} (...) besides every anime intro and outro song ever at least from the series i like , i 've been getting a kick out of j rock in general .
\begin{itemize}
%\item \textbf{Ground Truth:} i got deep into idol music this year so i can relate a bit .\\
\item \textbf{MoE-1 R1:} i got this feeling of thinking about this year and i can relate .
\item \textbf{MoE-1 R2:} i got this feeling i can relate to this song a bit .
\item \textbf{MoE-1 R3:} i got this song a lot , and i can relate to it .
\\
\item \textbf{MoE-2 R1:} i really miss the music i was listening to it .
\item \textbf{MoE-2 R2:}  i have n't heard of it for a long time .
\item \textbf{MoE-2 R3:} i 'm not a fan of anime at all .
\\
\item \textbf{MoE-3 R1:}  i got deep into idol music this year so i can relate a bit .
\item \textbf{MoE-3 R2:} i got very deep into this show so i can relate with this most .
\item \textbf{MoE-3 R3:} i got very into film school this year so i can relate with you .
\\
\item \textbf{MoE-4 R1:} no i would n't have watched anime or video .
\item \textbf{MoE-4 R2:} i only watched it in japanese and i was at a time .
\item \textbf{MoE-4 R3:} i feel like i m just listening to the same music again . 
\\
\end{itemize}
\item \textbf{Sample Conversation:} (...) i got deep into idol music this year so i can relate a bit .
\begin{itemize}
%\item \textbf{Ground Truth:} how deep in anime trash are you ?\\
\item \textbf{MoE-1 R1:} i 'm not the most interested in this movie .
\item \textbf{MoE-1 R2:} i 've been listening to the whole genre of the past .
\item \textbf{MoE-1 R3:} i 've been listening to a lot of people from the past few years .
\\
\item \textbf{MoE-2 R1:} i really miss the music i 've listened to before 
\item \textbf{MoE-2 R2:} i feel like i 've been listening to that music lately .
\item \textbf{MoE-2 R3:} i hate reddit and have no idea what it is 
\\
\item \textbf{MoE-3 R1:}   i was also a huge fan of music so i do n't listen to them tv .
\item \textbf{MoE-3 R2:} i was very confused on my youtube was bc the record is bothering me .
\item \textbf{MoE-3 R3:} i 'm also very into film .
\\
\item \textbf{MoE-4 R1:} no but i only listen to rap .
\item \textbf{MoE-4 R2:} i just watched anime and manga japan was probably the worst .
\item \textbf{MoE-4 R3:} i have n't heard anyone like anime .
\\
\end{itemize}
\item \textbf{Sample Conversation:} (...)  some people have the same username across multiple websites such as steam or xbox .
\begin{itemize}
%\item \textbf{Ground Truth:} you might even be able to get a hold of them in different places with only their username .\\
\item \textbf{MoE-1 R1:} i think it 's because people who use the internet , they do n't use them anymore .
\item \textbf{MoE-1 R2:} i think that 's the case with the people who have their own knowledge of their own .
\item \textbf{MoE-1 R3:} i think people are just too used to it in their 30s or at their own cult .
\\
\item \textbf{MoE-2 R1:} i 've seen people get downvoted and it 's all weird .
\item \textbf{MoE-2 R2:} i even feel like it 's true for everyone .
\item \textbf{MoE-2 R3:} i 've heard people are n't on reddit .
\\
\item \textbf{MoE-3 R1:}  you might also be able to get a hold of their own place without similar type places .
\item \textbf{MoE-3 R2:} you might also be able to get a hold of them within the same price or not .
\item \textbf{MoE-3 R3:} you might also be able to get a hold of them within different countries or whatever they send you .
\\
\item \textbf{MoE-4 R1:} i have noticed it on twitter , and facebook . i never have anonymous enough on twitter 
\item \textbf{MoE-4 R2:} i have n't gotten around to it . i am literally wanting to delete my account .
\item \textbf{MoE-4 R3:} i do n't know why people are so dumb and useless .
\\
\end{itemize}
\end{enumerate}

\subsection{Sample Utterances for EXP 3 (with Full MoE Decision Flow)}\label{app:sample_utterances_3}

\subsubsection{Reddit Casual, MoE-4 MBRL}
\begin{enumerate}
\item \textbf{Turn 0:} (...) when i actually look back at it , it 's not bad .
\begin{itemize}
\item \textbf{MoE Primitive 1:}   oh god those words mean a lot in general . \textbf{$Q$-value:} 0.15
\item \textbf{MoE Primitive 2:}  when i get a good amount of things i put in the mood . \textbf{$Q$-value:} -0.32
\item \textbf{MoE Question 1:} i have n't listened to music , but i played a few songs , what 's it ? \textbf{$Q$-value:} -0.38
\item \textbf{MoE Question 2:} how much is it ? \textbf{$Q$-value:} -0.41
\item \textbf{MoE Exploration 1:} and that 's just me being the only way i can discover a life . \textbf{$Q$-value:} -0.26
\item \textbf{MoE Exploration 2:} i 'm 22 and have n't hugged a girl yet forget about kissing . \textbf{$Q$-value:} -0.55
\item \textbf{MoE Positive Sentiment 1:} thank you for the amazing reply , i definitely look at it to a high day ! \textbf{$Q$-value:} 0.45
\item \textbf{MoE Positive Sentiment 2:} good for you , i really love the genre so things i consider a best with as my life , it was a very difficult decision \textbf{$Q$-value:} 0.05
\item \textbf{MoE Negative Sentiment 1:} but on my own , i just do n't see the appeal anymore with how it is . \textbf{$Q$-value:} 0.05
\item \textbf{MoE Negative Sentiment 2:} i hate it when i 'm slow . \textbf{$Q$-value:} -0.46
\item \textbf{MoE Sentiment Coherence 1:} for me , i have a taste in music so i can not stand a lot of art or without it i would be so tired \textbf{$Q$-value:} -0.38
\item \textbf{MoE Sentiment Coherence 2:} the reason i love it , i have n't been in a while , but i feel like i was in a way that was when i was younger . \textbf{$Q$-value:} 0.25
\item \textbf{MoE Joy Emotion 1:} haha well i 'm going through a lot of shit . \textbf{$Q$-value:} -0.25
\item \textbf{MoE Joy Emotion 2:} haha yes i have my own one of my favorite things to do . \textbf{$Q$-value:} -0.62
\item \textbf{MoE Optimism Emotion 1:}  good for you , i like the original music , it 's a new skill in painting , some of it , and just like every other \textbf{$Q$-value:} -0.14
\item \textbf{MoE Optimism Emotion 2:} good on you for doing the right thing , just remember that the past few months , i 've been putting it off , and have no idea about music \textbf{$Q$-value:} 0.69
\item \textbf{MoE Anger Emotion 1:} why do you think you re an adult ? \textbf{$Q$-value:} -0.16
\item \textbf{MoE Anger Emotion 2:} why are you not asking me a question ? \textbf{$Q$-value:} -0.29
\item \textbf{MoE Sadness Emotion 1:} i do n't have much of an effect on my mood , what should i do ! \textbf{$Q$-value:} 0.60
\item \textbf{MoE Sadness Emotion 2:} i have n't done that yet but i 'm going to check out . \textbf{$Q$-value:} -0.25 \\
\item \textbf{Final Bot Response: } \textbf{$Q$-value:} good on you for doing the right thing , just remember that the past few months , i 've been putting it off and have no idea  about music \textbf{$Q$-value:} 0.69 \\
\item \textbf{GPT-based Next-user Response: } just love these memory of having that moment of myself and lots to remember the past few years on life of music  . \\

\end{itemize}

\item \textbf{Turn 1:} (...) just love these memory of having that moment of myself and lots to remember the past few years on life of music  .
\begin{itemize}
\item \textbf{MoE Primitive 1:}  it has become an actual meme , and then on a cracked up .  \textbf{$Q$-value:} 0.21
\item \textbf{MoE Primitive 2:}  i just love music , what genre would it be ? \textbf{$Q$-value:} 0.42
\item \textbf{MoE Question 1:} how much do you enjoy now ? \textbf{$Q$-value:} -0.15
\item \textbf{MoE Question 2:} how do you discover new music ? \textbf{$Q$-value:} 0.08
\item \textbf{MoE Exploration 1:} but i was thinking of going to be a one , and maybe <unk> my favorite shows . \textbf{$Q$-value:} 0.35
\item \textbf{MoE Exploration 2:} and it 's a good time to focus on something else , like a good <unk> ' , <unk> out of the good things . \textbf{$Q$-value:} -0.43
\item \textbf{MoE Positive Sentiment 1:} thanks for the advice , i really hope things turn to someone who does n't have a life and does n't deserve it and as much to be \textbf{$Q$-value:} -0.51
\item \textbf{MoE Positive Sentiment 2:} thank you for the inspiration ! \textbf{$Q$-value:} 0.13
\item \textbf{MoE Negative Sentiment 1:} but i hate when i have to do it all . \textbf{$Q$-value:} 0.18
\item \textbf{MoE Negative Sentiment 2:} i am not one of those and i do n't know about to be honest . \textbf{$Q$-value:} -0.47
\item \textbf{MoE Sentiment Coherence 1:} for the future , i love it and remember that song <unk> would be a very nice place to achieve something like <unk> or goals . \textbf{$Q$-value:} -0.57
\item \textbf{MoE Sentiment Coherence 2:} for me , i love the feeling of wanting to have a good christmas tree since i was a kid \textbf{$Q$-value:} -0.77
\item \textbf{MoE Joy Emotion 1:} haha , i am enjoying a lot of it , so i personally even know nothing about it .  \textbf{$Q$-value:} 0.09
\item \textbf{MoE Joy Emotion 2:} haha , this is one of my favorite things to listen in the genre . \textbf{$Q$-value:} 0.71
\item \textbf{MoE Optimism Emotion 1:} good for you , the best of luck in the future ! \textbf{$Q$-value:} 0.13
\item \textbf{MoE Optimism Emotion 2:} for me , the only thing i really like , is the music i like , and the way it gives me time to myself and be  \textbf{$Q$-value:} 0.48
\item \textbf{MoE Anger Emotion 1:} and what is an example of things you could do ? \textbf{$Q$-value:} -0.11
\item \textbf{MoE Anger Emotion 2:} and i do n't like music , i 've felt so far . \textbf{$Q$-value:} -0.02
\item \textbf{MoE Sadness Emotion 1:} i do n't know what to do with my knowledge but i have no idea what it is . \textbf{$Q$-value:} -0.03 
\item \textbf{MoE Sadness Emotion 2:} i have no clue why i do that these days . \textbf{$Q$-value:} -0.31\\
\item \textbf{Final Bot Response: } haha , this is one of my favorite things to listen in the genre . \textbf{$Q$-value:} 0.71 \\
\item \textbf{GPT-based Next-user Response: } I wonder how how it 'd go . but i do enjoy them . they feel live and have the meaning of life .\\

\end{itemize}
\end{enumerate}

\subsubsection{Reddit Casual, MoE-4 CQL}
\begin{enumerate}
\item \textbf{Turn 0:} (...) when i actually look back at it , it 's not bad .
\begin{itemize}
\item \textbf{MoE Primitive 1:} well i 'm glad you 're feeling better . \textbf{$Q$-value:} 0.38
\item \textbf{MoE Primitive 2:} did you get a song that you 're listening to ?  \textbf{$Q$-value:} -0.06
\item \textbf{MoE Question 1:} what 's it worth ? \textbf{$Q$-value:} 0.35
\item \textbf{MoE Question 2:} how did you get out of it ? \textbf{$Q$-value:} 0.12
\item \textbf{MoE Exploration 1:} lately i 've been a lot of good things to do . \textbf{$Q$-value:} -0.55
\item \textbf{MoE Exploration 2:} i love this song , and i will be happy with the music . \textbf{$Q$-value:} -0.26
\item \textbf{MoE Positive Sentiment 1:} i 'm glad you 're able to find a great life ! \textbf{$Q$-value:} 0.40
\item \textbf{MoE Positive Sentiment 2:} i hope you have a great day ! \textbf{$Q$-value:} 0.39
\item \textbf{MoE Negative Sentiment 1:} i do n't know if it 's worth a listen to it . \textbf{$Q$-value:} 0.22
\item \textbf{MoE Negative Sentiment 2:} i do n't know if i 'm going to be a bit stressful . \textbf{$Q$-value:} 0.17
\item \textbf{MoE Sentiment Coherence 1:} i 'll be able to do something with my own , but i 'll be able to find a new genre . \textbf{$Q$-value:} -0.11
\item \textbf{MoE Sentiment Coherence 2:} i might try to find some new songs that i 'm not sure if i 'm a bit interested in the music . \textbf{$Q$-value:} 0.20
\item \textbf{MoE Joy Emotion 1:} i feel so much better now \textbf{$Q$-value:} -0.45
\item \textbf{MoE Joy Emotion 2:} i want to do that myself . \textbf{$Q$-value:} -0.22
\item \textbf{MoE Optimism Emotion 1:} be proud of yourself and happy for the first time i 've been able to make a great life . \textbf{$Q$-value:} 0.29
\item \textbf{MoE Optimism Emotion 2:} be proud of yourself and be happy for the first time i 've been to a great life . \textbf{$Q$-value:} 0.09
\item \textbf{MoE Anger Emotion 1:} it 's so much easier to say than done ! \textbf{$Q$-value:} -0.50
\item \textbf{MoE Anger Emotion 2:}  i need to start doing something productive , but i 'm not doing anything . \textbf{$Q$-value:} -0.11
\item \textbf{MoE Sadness Emotion 1:} i just started a year ago and i 'm not sure what to do . \textbf{$Q$-value:} 0.16
\item \textbf{MoE Sadness Emotion 2:} i 've been trying to find some new songs that are n't a thing . \textbf{$Q$-value:} 0.12\\
\item \textbf{Final Bot Response: }i 'm glad you 're able to find a great life ! \textbf{$Q$-value:} 0.40 \\
\item \textbf{GPT-based Next-user Response: } yes , life feels young and happy again . \\
\end{itemize}

\item \textbf{Turn 1:} (...) yes , life feels young and happy again . 
\begin{itemize}
\item \textbf{MoE Primitive 1:} i feel the same way .  \textbf{$Q$-value:} 0.061
\item \textbf{MoE Primitive 2:}  i 've been through a lot of pain , but i do n't know how to do it .  \textbf{$Q$-value:} -0.045
\item \textbf{MoE Question 1:} how do you feel about it ? \textbf{$Q$-value:} -0.35
\item \textbf{MoE Question 2:} i can understand that . how do you cope with this ? \textbf{$Q$-value:} 0.03
\item \textbf{MoE Exploration 1:} but that 's how i feel about it and i 'm not going to be able to enjoy it . \textbf{$Q$-value:} 0.15
\item \textbf{MoE Exploration 2:} because it 's not a good feeling to be able to do something i \textbf{$Q$-value:} -0.56
\item \textbf{MoE Positive Sentiment 1:} be proud of yourself and be the best part of life , i 'm happy that you 're happy . \textbf{$Q$-value:} 0.85
\item \textbf{MoE Positive Sentiment 2:} i hope i can find a great life too ! \textbf{$Q$-value:} 0.65
\item \textbf{MoE Negative Sentiment 1:} i do n't know if it 's going to be the problem . \textbf{$Q$-value:} -0.22
\item \textbf{MoE Negative Sentiment 2:} yeah i feel like i have a hard time . \textbf{$Q$-value:} 0.18
\item \textbf{MoE Sentiment Coherence 1:} i might try to get into the good stuff i 'm not going to do . \textbf{$Q$-value:} 0.05
\item \textbf{MoE Sentiment Coherence 2:} i 'll be able to find the better , but i 'm not sure if i can . \textbf{$Q$-value:} -0.19
\item \textbf{MoE Joy Emotion 1:} i just enjoy playing music , and i 'm happy for myself . \textbf{$Q$-value:} 0.55
\item \textbf{MoE Joy Emotion 2:} just in the mood , i 'm going to enjoy a bit of it . \textbf{$Q$-value:} -0.33
\item \textbf{MoE Optimism Emotion 1:} just in the mood , i 'm going to enjoy a bit of it . \textbf{$Q$-value:} -0.61
\item \textbf{MoE Optimism Emotion 2:} i just enjoy playing music , and i 'm happy for myself . \textbf{$Q$-value:} 0.28
\item \textbf{MoE Anger Emotion 1:} i just started a new year ago and i 'm not a huge fan of the music . \textbf{$Q$-value:} -0.43
\item \textbf{MoE Anger Emotion 2:}  why do you have to do this ? \textbf{$Q$-value:} -0.59
\item \textbf{MoE Sadness Emotion 1:} got any hobbies that i feel like i 'm alone . \textbf{$Q$-value:} -0.13
\item \textbf{MoE Sadness Emotion 2:} do you have any advice or am i just a little curious about this ? \textbf{$Q$-value:} -0.08\\
\item \textbf{Final Bot Response: } be proud of yourself and be the best part of life , i 'm happy that you 're happy . \\
\item \textbf{GPT-based Next-user Response: } thank you i 'd proud of you 've been enjoy life too !\\

\end{itemize}

\end{enumerate}

\subsubsection{Reddit Casual, MoE-4 DQN}
\begin{enumerate}
\item \textbf{Turn 0:} (...) when i actually look back at it , it 's not bad .
\begin{itemize}
\item \textbf{MoE Primitive 1:}  oh god those words mean a lot in general too . \textbf{$Q$-value:} 0.03
\item \textbf{MoE Primitive 2:}  when i m in a really bad mood , i ll have a look about myself . \textbf{$Q$-value:} -0.05
\item \textbf{MoE Question 1:} how do you know what 's good ? \textbf{$Q$-value:} -0.25
\item \textbf{MoE Question 2:} how in the world do you have all the time ? \textbf{$Q$-value:} -0.35
\item \textbf{MoE Exploration 1:}  and that 's what i think of when i want to have a good idea . \textbf{$Q$-value:} 0.52
\item \textbf{MoE Exploration 2:} at least i have a couple of hours of listening to the same tunes . \textbf{$Q$-value:} 0.28
\item \textbf{MoE Positive Sentiment 1:} thanks for the good things , i 've found in 2018 , still , and if you like it 's a little bit of inspiration , my own memories . \textbf{$Q$-value:} 0.16
\item \textbf{MoE Positive Sentiment 2:} thank you for your recommendations , i definitely do n't have a fantastic artist , it 's absolutely amazing to me , but i really want to learn more \textbf{$Q$-value:} -0.4
\item \textbf{MoE Negative Sentiment 1:} but on my mind , i ca n't even focus on the music . \textbf{$Q$-value:} -0.06
\item \textbf{MoE Negative Sentiment 2:} but no i just do n't know ... \textbf{$Q$-value:} 0.55
\item \textbf{MoE Sentiment Coherence 1:} thanks for the recommendation i really enjoyed it ! \textbf{$Q$-value:} -0.04
\item \textbf{MoE Sentiment Coherence 2:} that 's what i 've heard about in the first place . \textbf{$Q$-value:} -0.13
\item \textbf{MoE Joy Emotion 1:} not bad , i have all my <unk> on the side and <unk> . \textbf{$Q$-value:} -0.18
\item \textbf{MoE Joy Emotion 2:} good for you , for me , my best friend is <unk> \textbf{$Q$-value:} -0.27
\item \textbf{MoE Optimism Emotion 1:} good on you for the time , but it 's actually a bit of a hiatus from the genre your get to your development aspect , regardless of songs \textbf{$Q$-value:} -0.67
\item \textbf{MoE Optimism Emotion 2:} you know what i m missing out on . \textbf{$Q$-value:} 0.24
\item \textbf{MoE Anger Emotion 1:} i do n't like the term toxic masculinity , but i think this idea of pride is accurately described . \textbf{$Q$-value:} -0.36
\item \textbf{MoE Anger Emotion 2:} why do you think the weed smell is a good idea ? \textbf{$Q$-value:} -0.64
\item \textbf{MoE Sadness Emotion 1:} i do n't have much of a high expectation in my life . \textbf{$Q$-value:} 0.15
\item \textbf{MoE Sadness Emotion 2:} i 'm no expert on turtles and different types of turtle species , i just saw an interesting documentary on sea turtles . \textbf{$Q$-value:} -0.04\\
\item \textbf{Final Bot Response: } but no i just do n't know ... \textbf{$Q$-value:} 0.55\\
\item \textbf{GPT-based Next-user Response: } i agree , just make sure what you want to do  and figure out how to proud

\end{itemize}

\item \textbf{Turn 1:} (...)  i agree , just make sure what you want to do  and figure out how to proud
\begin{itemize}
\item \textbf{MoE Primitive 1:} it has become an actual effect on my mood .  \textbf{$Q$-value:} 0.65
\item \textbf{MoE Primitive 2:} but i 've discovered that by not giving any fucks about what i want .  \textbf{$Q$-value:} 0.13
\item \textbf{MoE Question 1:} how the fuck do you think about it ? \textbf{$Q$-value:} -0.06
\item \textbf{MoE Question 2:} how are you enjoying life , what you used to do ? \textbf{$Q$-value:} 0.18
\item \textbf{MoE Exploration 1:} that s why i have a favourite meme playlist right . \textbf{$Q$-value:} -0.24
\item \textbf{MoE Exploration 2:} for me , i do n't know what a genre is hit these days . \textbf{$Q$-value:} -0.34
\item \textbf{MoE Positive Sentiment 1:} for me , i love the good drawing and trying to keep it up , but it s not a big deal , i like a lot of it \textbf{$Q$-value:} 0.23
\item \textbf{MoE Positive Sentiment 2:} good for you , for me , my best friend is music \textbf{$Q$-value:} 0.12
\item \textbf{MoE Negative Sentiment 1:} i hate to admit it , i just look through the whole world . \textbf{$Q$-value:} 0.08
\item \textbf{MoE Negative Sentiment 2:} i hate to admit it , i do n't know what i like to listen \textbf{$Q$-value:} 0.28
\item \textbf{MoE Sentiment Coherence 1:} taste is all subjective and i have no matter to taste or texture other things on the genre . \textbf{$Q$-value:} -0.55
\item \textbf{MoE Sentiment Coherence 2:}  although i 'm on youtube , i enjoy old music .  \textbf{$Q$-value:} 0.17
\item \textbf{MoE Joy Emotion 1:} good for you , there s really no way to do it , but i m in the mood for it , in my day \textbf{$Q$-value:} -0.23
\item \textbf{MoE Joy Emotion 2:} good on you for the time , i think it 's a little complicated haha \textbf{$Q$-value:} 0.45
\item \textbf{MoE Optimism Emotion 1:} good one , i have an amazing time , but , the little it can be so challenging , especially to discover the passion . \textbf{$Q$-value:} 0.75
\item \textbf{MoE Optimism Emotion 2:} this is the best recommendation ever , and at a good old dance moves are n't any other , or at the same time  \textbf{$Q$-value:} 0.28
\item \textbf{MoE Anger Emotion 1:} no taste in music , but the more <unk> ' is my favorite song ever on things . \textbf{$Q$-value:} -0.48
\item \textbf{MoE Anger Emotion 2:} also , i have n't watched an entire year of it . \textbf{$Q$-value:} -0.08
\item \textbf{MoE Sadness Emotion 1:} why do i have to not use it ? \textbf{$Q$-value:} -0.0
\item \textbf{MoE Sadness Emotion 2:} i do n't know why i stopped putting on so much . \textbf{$Q$-value:} 0.15\\
\item \textbf{Final Bot Response: } good one , i have an amazing time , but , the little it can be so challenging , especially to discover the passion . \textbf{$Q$-value:} 0.75\\
\item \textbf{GPT-based Next-user Response: } i already have a really vivid of memories of those conversations , thats what i have heard from .\\

\end{itemize}
\end{enumerate}

\subsubsection{Cornell Movie Corpus, MoE-2 MBRL}
\begin{enumerate}
\item \textbf{Turn 0:} (...) excuse me
\begin{itemize}
\item \textbf{MoE Primitive 1:} you , huh ? is that you ?  \textbf{$Q$-value:} 0.17
\item \textbf{MoE Primitive 2:} i 'll promise you . but i 'd not like one of them , do n't worry .   \textbf{$Q$-value:} 0.28
\item \textbf{MoE Question 1:} what ? what ? because he was worried about ? \textbf{$Q$-value:} -0.34
\item \textbf{MoE Question 2:} what ? oh , i thought maybe you did too ? \textbf{$Q$-value:} -0.28
\item \textbf{MoE Exploration 1:} he asked you for a few bucks ? i want him along , anyway  \textbf{$Q$-value:} 0.05
\item \textbf{MoE Exploration 2:}  i do n't know . i 've been hearing from you since i left . \textbf{$Q$-value:} -0.01
\item \textbf{MoE Positive Sentiment 1:} he says you 're very nice . he has a gun . \textbf{$Q$-value:} -0.46
\item \textbf{MoE Positive Sentiment 2:}  that 's right . he said that his name was familiar and the formal negotiation is handled ' \textbf{$Q$-value:} 0.18
\item \textbf{MoE Negative Sentiment 1:} i do n't know . i packed up my shit and ran away  to here . \textbf{$Q$-value:} -0.22
\item \textbf{MoE Negative Sentiment 2:} no , i was n't gon na see you tonight . \textbf{$Q$-value:} -0.59
\item \textbf{MoE Sentiment Coherence 1:} i do n't know what happened . i 've never been there before . \textbf{$Q$-value:} 0.05
\item \textbf{MoE Sentiment Coherence 2:} i do n't know . i was with him and you 're all i know is that these years i thought you were going to do it \textbf{$Q$-value:} -0.35
\item \textbf{MoE Joy Emotion 1:}  i 'm glad you got your memory back , father ? \textbf{$Q$-value:} 0.33
\item \textbf{MoE Joy Emotion 2:} haha , it is all good . \textbf{$Q$-value:} -0.17
\item \textbf{MoE Optimism Emotion 1:}  i 'm so sure you do . and he 's got lots of friends left \textbf{$Q$-value:} -0.07
\item \textbf{MoE Optimism Emotion 2:} i was n't saying anything . i was singing . \textbf{$Q$-value:} 0.16
\item \textbf{MoE Anger Emotion 1:} What ? you do n't pull that ? \textbf{$Q$-value:} -0.77
\item \textbf{MoE Anger Emotion 2:} you did n't work for him , did n't you ? \textbf{$Q$-value:} -0.56 
\item \textbf{MoE Sadness Emotion 1:} he does n't like me ? \textbf{$Q$-value:} -0.15
\item \textbf{MoE Sadness Emotion 2:}  i 'm sorry , i was n't feeling that way . \textbf{$Q$-value:} 0.21\\
\item \textbf{Final Bot Response: } i 'm glad you got your memory back , father ? \textbf{$Q$-value:} 0.33 \\
\item \textbf{GPT-based Next-user Response: } sure yea . you see i should worry nothing and do exactly as said\\
\end{itemize}

\item \textbf{Turn 1:} (...) sure yea . you see i should worry nothing and do exactly as said
\begin{itemize}
\item \textbf{MoE Primitive 1:} i went down to meet the president .  \textbf{$Q$-value:} -0.14
\item \textbf{MoE Primitive 2:} he 's the one i told him you .   \textbf{$Q$-value:} 0.08
\item \textbf{MoE Question 1:} what ? what did you think he wanted to do , then ? \textbf{$Q$-value:} -0.35
\item \textbf{MoE Question 2:} what did you think ? when did you talk about it ? \textbf{$Q$-value:} -0.05
\item \textbf{MoE Exploration 1:} i 'm a little girl . if you do n't get me a girlfriend , maybe five bucks . \textbf{$Q$-value:} -0.20
\item \textbf{MoE Exploration 2:} it was like tommy was possessed or something . there was a hate in his eyes that i could n't really call human . \textbf{$Q$-value:} -0.41
\item \textbf{MoE Positive Sentiment 1:} it 's so nice to see you again . \textbf{$Q$-value:} 0.45
\item \textbf{MoE Positive Sentiment 2:} it was so cool . you know who 's being a jerk . \textbf{$Q$-value:} 0.05
\item \textbf{MoE Negative Sentiment 1:} i do n't want to be a fool . you do n't want to look ? \textbf{$Q$-value:} -0.04
\item \textbf{MoE Negative Sentiment 2:} i do n't care what you say ! i do n't want to fight you . kinda <unk> . \textbf{$Q$-value:} -0.46
\item \textbf{MoE Sentiment Coherence 1:} no . i think he 's in the house . \textbf{$Q$-value:} -0.43
\item \textbf{MoE Sentiment Coherence 2:} i do n't know what else you 'd call it . \textbf{$Q$-value:} -0.35
\item \textbf{MoE Joy Emotion 1:} it 's not that bad , it just the way you always wanted it to be , or did you ? \textbf{$Q$-value:} 0.33
\item \textbf{MoE Joy Emotion 2:} there you go . i like you . \textbf{$Q$-value:} 0.34
\item \textbf{MoE Optimism Emotion 1:}  i 'm sure he 's very flattering , birdie , and i 'm sure there 's nothing wrong with that . \textbf{$Q$-value:} -0.25
\item \textbf{MoE Optimism Emotion 2:} he said it 's up to us . \textbf{$Q$-value:} 0.02
\item \textbf{MoE Anger Emotion 1:} you do n't get paid till you calm down . you 're sure you want to stay ? \textbf{$Q$-value:} -0.26
\item \textbf{MoE Anger Emotion 2:} you say that because you have no idea ? \textbf{$Q$-value:} -0.31
\item \textbf{MoE Sadness Emotion 1:}  no , i 'm surprised what three years ago . i do n't know . \textbf{$Q$-value:} -0.05
\item \textbf{MoE Sadness Emotion 2:} i 'm sorry , i was n't feeling like this , all . \textbf{$Q$-value:} -0.10\\
\item \textbf{Final Bot Response: } it 's so nice to see you again . \textbf{$Q$-value:} 0.45 \\
\item \textbf{GPT-based Next-user Response: } we think about you too , love you ,\\

\end{itemize}

\end{enumerate}

\subsubsection{Cornell Movie Corpus, MoE-2 CQL}
\begin{enumerate}
\item \textbf{Turn 0:} (...) excuse me
\begin{itemize}
\item \textbf{MoE Primitive 1:}  you , huh ? is that you ? \textbf{$Q$-value:} -0.42
\item \textbf{MoE Primitive 2:} jesus , not sure . how are you ?  \textbf{$Q$-value:} -0.24
\item \textbf{MoE Question 1:} what ? what ? you think he wanted to kill the baby ? \textbf{$Q$-value:} -0.091
\item \textbf{MoE Question 2:} what 's the matter ? \textbf{$Q$-value:} -0.080
\item \textbf{MoE Exploration 1:} i 'll destroy you , i swear it . \textbf{$Q$-value:} -0.51
\item \textbf{MoE Exploration 2:} he asked you for a few bucks ? do you want him along ? \textbf{$Q$-value:} -0.13
\item \textbf{MoE Positive Sentiment 1:} sure , i 'd love to see you again . \textbf{$Q$-value:} -0.07
\item \textbf{MoE Positive Sentiment 2:} it 's so beautiful . he was crazy about you . \textbf{$Q$-value:} 0.37
\item \textbf{MoE Negative Sentiment 1:} i do n't want to hear it . i know you 're talking to him . \textbf{$Q$-value:} -0.11
\item \textbf{MoE Negative Sentiment 2:} i did n't want to get involved . did you ever think to work with the police station chiefs . \textbf{$Q$-value:} -0.24
\item \textbf{MoE Sentiment Coherence 1:} i do n't know what else you 'd call it . \textbf{$Q$-value:} -0.26
\item \textbf{MoE Sentiment Coherence 2:} i was tired ,  what is that for ? \textbf{$Q$-value:} -0.26
\item \textbf{MoE Joy Emotion 1:} he 's likely to be in love with you . \textbf{$Q$-value:} -0.24
\item \textbf{MoE Joy Emotion 2:} i 'll be right there . i 'll have the money , you got that go . \textbf{$Q$-value:} 0.25
\item \textbf{MoE Optimism Emotion 1:}  i 'm sure you 'll remember that , in the first date ? \textbf{$Q$-value:} 0.25
\item \textbf{MoE Optimism Emotion 2:} i 'm surprised you do n't know . when i was a kid , i live off the street . \textbf{$Q$-value:} 0.12
\item \textbf{MoE Anger Emotion 1:} you do n't sound like it ! \textbf{$Q$-value:} -0.61
\item \textbf{MoE Anger Emotion 2:} you were n't attracted to him ? \textbf{$Q$-value:} -0.11
\item \textbf{MoE Sadness Emotion 1:} i 'm sorry , i do n't know how to make it out . \textbf{$Q$-value:} -0.17
\item \textbf{MoE Sadness Emotion 2:} i 'm sorry , i do n't know what to say . \textbf{$Q$-value:} 0.05\\
\item \textbf{Final Bot Response: } it 's so beautiful . he was crazy about you . \textbf{$Q$-value:} 0.37 \\
\item \textbf{GPT-based Next-user Response: } i 'm glad to know . but tonight mr. wade may not come here . \\
\end{itemize}

\item \textbf{Turn 1:} (...) i 'm glad to know . but tonight mr. mel may not come here .
\begin{itemize}
\item \textbf{MoE Primitive 1:} certainly , when i get to go up to your room .  \textbf{$Q$-value:} -0.33
\item \textbf{MoE Primitive 2:} no . do n't even be serious . but i 'm telling me the truth .  \textbf{$Q$-value:} -0.21
\item \textbf{MoE Question 1:}  what do you think he got to lose ? \textbf{$Q$-value:} -0.06
\item \textbf{MoE Question 2:} then what 's he talking about ? i 'm sorry . \textbf{$Q$-value:} -0.10
\item \textbf{MoE Exploration 1:} i just want to hear the details ... \textbf{$Q$-value:} 0.39
\item \textbf{MoE Exploration 2:} i do n't want to hear you anymore . you made love ? \textbf{$Q$-value:} -0.09
\item \textbf{MoE Positive Sentiment 1:} yes . i can live here with you \textbf{$Q$-value:} 0.25
\item \textbf{MoE Positive Sentiment 2:} right , youth . that 's why i need you to lead the troops . \textbf{$Q$-value:} -0.45
\item \textbf{MoE Negative Sentiment 1:}  i do n't know . he fell for you . \textbf{$Q$-value:} -0.35
\item \textbf{MoE Negative Sentiment 2:} i do n't blame him . what happened ? \textbf{$Q$-value:} 0.01
\item \textbf{MoE Sentiment Coherence 1:} but you did n't think about it . what 's the one said . \textbf{$Q$-value:} -0.08
\item \textbf{MoE Sentiment Coherence 2:} i do n't know . i thought i might . but i 'm sure he 's quite human . \textbf{$Q$-value:} -0.06
\item \textbf{MoE Joy Emotion 1:} it 's just you know , i mean . it 's been a pleasure . \textbf{$Q$-value:}
\item \textbf{MoE Joy Emotion 2:} i 'm sure he 's the right person . \textbf{$Q$-value:}
\item \textbf{MoE Optimism Emotion 1:} yeah , i know . it 's been so long . \textbf{$Q$-value:} -0.18
\item \textbf{MoE Optimism Emotion 2:} i 'm sure you have n't done anything . \textbf{$Q$-value:} -0.22
\item \textbf{MoE Anger Emotion 1:} what ? i do n't know what else ? \textbf{$Q$-value:} 0.13
\item \textbf{MoE Anger Emotion 2:} you do n't have to prove it ? \textbf{$Q$-value:}
\item \textbf{MoE Sadness Emotion 1:} i was n't aware of that . but this is the beauty of it . \textbf{$Q$-value:} 0.55
\item \textbf{MoE Sadness Emotion 2:} clark . i would n't want to come in tonight . \textbf{$Q$-value:} -0.24\\
\item \textbf{Final Bot Response: } i was n't aware of that . but this is the beauty of it . \textbf{$Q$-value:} 0.55 \\
\item \textbf{GPT-based Next-user Response: } i 'm surprised you do n't know . he said some good thing about you too \\
\end{itemize}

\end{enumerate}

\subsubsection{Cornell Movie Corpus, MoE-2 DQN}
\begin{enumerate}
\item \textbf{Turn 0:} (...) excuse me
\begin{itemize}
\item \textbf{MoE Primitive 1:} i 'll promise you . but i do n't have any new evidence .  \textbf{$Q$-value:} -0.19
\item \textbf{MoE Primitive 2:} no , it 's not the same thing . you do n't have the same . \textbf{$Q$-value:} -0.26
\item \textbf{MoE Question 1:} what ? what do you think happened here ? \textbf{$Q$-value:} 0.45
\item \textbf{MoE Question 2:} what did you think ? i 'm going to tell you what . it 's over , boy . \textbf{$Q$-value:} 0.11
\item \textbf{MoE Exploration 1:} it 's a lot of money for you to do . you know , i mean , you know , i 'm not going to do that \textbf{$Q$-value:} 0.04
\item \textbf{MoE Exploration 2:} it 's the only way i can do it . i 'm not going to be a writer . \textbf{$Q$-value:} 0.05
\item \textbf{MoE Positive Sentiment 1:} yes . it 's very nice . \textbf{$Q$-value:} -0.20
\item \textbf{MoE Positive Sentiment 2:} yes , it 's true . \textbf{$Q$-value:} -0.32
\item \textbf{MoE Negative Sentiment 1:} i do n't want to be here . i 'm not in love with you . \textbf{$Q$-value:} -0.25
\item \textbf{MoE Negative Sentiment 2:} do you want to be sure ? i want to know what you want . \textbf{$Q$-value:} -0.16
\item \textbf{MoE Sentiment Coherence 1:} i do n't know what you 're talking about . \textbf{$Q$-value:} -0.15
\item \textbf{MoE Sentiment Coherence 2:} you do n't have to be a cop . you 're not going to be there . \textbf{$Q$-value:} 0.03
\item \textbf{MoE Joy Emotion 1:} i 'm so sure you could have been anything ? \textbf{$Q$-value:} -0.28
\item \textbf{MoE Joy Emotion 2:} it was you , was n't it ? i 've seen your show . \textbf{$Q$-value:} 0.11
\item \textbf{MoE Optimism Emotion 1:} it 's not your fault . \textbf{$Q$-value:} 0.15
\item \textbf{MoE Optimism Emotion 2:} that 's all right , you 're gon na be fine . \textbf{$Q$-value:} -0.51
\item \textbf{MoE Anger Emotion 1:} you do n't pull that kick out your ass , you 'll try ? \textbf{$Q$-value:} -0.26
\item \textbf{MoE Anger Emotion 2:}  you do n't have to do this to me ! \textbf{$Q$-value:} -0.53
\item \textbf{MoE Sadness Emotion 1:} i 'm sorry , i did n't know it . all i know is he did n't get to be here first . \textbf{$Q$-value:} 0.23
\item \textbf{MoE Sadness Emotion 2:} i 'm sorry to trouble you , but i have to . \textbf{$Q$-value:} 0.15\\
\item \textbf{Final Bot Response: } what ? what do you think happened here ? \textbf{$Q$-value:} 0.45 \\
\item \textbf{GPT-based Next-user Response: } jesus , not sure . huh ? is that you ?\\
\end{itemize}

\item \textbf{Turn 1:} (...) jesus , not sure . huh ? is that you ?
\begin{itemize}
\item \textbf{MoE Primitive 1:} we ran up against a platoon of soldiers . there were actually walking dead .  \textbf{$Q$-value:} -0.05
\item \textbf{MoE Primitive 2:} no . it 's him , but he did n't make the choice .  \textbf{$Q$-value:} 0.87
\item \textbf{MoE Question 1:} what ? what would you think ? i 'm going to be here . i 've never seen this guy . \textbf{$Q$-value:} 0.75
\item \textbf{MoE Question 2:} what did you think ? when did you talk about it ? \textbf{$Q$-value:} 0.18
\item \textbf{MoE Exploration 1:} that 's all i ever wanted . a man with the world without a different girl from one , desert one 's a dead , and later . \textbf{$Q$-value:} -0.51
\item \textbf{MoE Exploration 2:} i dunno ... police . \textbf{$Q$-value:} -0.02
\item \textbf{MoE Positive Sentiment 1:} it was like , was n't it ? you 're far too good . \textbf{$Q$-value:} -0.44
\item \textbf{MoE Positive Sentiment 2:} it 's so nice to see you again . \textbf{$Q$-value:} -0.56
\item \textbf{MoE Negative Sentiment 1:}  oh , i do n't care . but it 's got to be different . \textbf{$Q$-value:} -0.11
\item \textbf{MoE Negative Sentiment 2:} i do n't want to hurt you . he 's not trying to get involved . \textbf{$Q$-value:} -0.30
\item \textbf{MoE Sentiment Coherence 1:} no . i was just going to say , you know . \textbf{$Q$-value:} -0.45
\item \textbf{MoE Sentiment Coherence 2:} i do n't know what else you 'd call it . \textbf{$Q$-value:} -0.53
\item \textbf{MoE Joy Emotion 1:} i can be sure of that when you want your opinion . \textbf{$Q$-value:} -0.07
\item \textbf{MoE Joy Emotion 2:} it was so obvious . i mean , all of nicky 's half assed mechanics , they were real signal happy . \textbf{$Q$-value:} 0.28
\item \textbf{MoE Optimism Emotion 1:} i 'm sure he 's very flattering , birdie , and i 'm sure there 's nothing wrong with that . \textbf{$Q$-value:} -0.15
\item \textbf{MoE Optimism Emotion 2:} there you go . that 's the truth . \textbf{$Q$-value:} 0.08
\item \textbf{MoE Anger Emotion 1:} you do n't like him , do you ? \textbf{$Q$-value:} -0.07
\item \textbf{MoE Anger Emotion 2:} he 's sleeping in the back , does n't he ? \textbf{$Q$-value:} -0.31
\item \textbf{MoE Sadness Emotion 1:}  no i was right behind you , but you 're too dumb to turn around . \textbf{$Q$-value:} 0.07
\item \textbf{MoE Sadness Emotion 2:}  i do n't know , i 've seen him . that 's the name of a man i thought . \textbf{$Q$-value:} -0.25\\
\item \textbf{Final Bot Response: } no . it 's him , but he did n't make the choice . \textbf{$Q$-value:} 0.87 \\
\item \textbf{GPT-based Next-user Response: }  that left me a surprise . did he say something ?\\

\end{itemize}

\end{enumerate}

}

\end{document}